\documentclass[10pt,twocolumn,letterpaper]{article}

\usepackage[pagenumbers]{cvpr} 

\definecolor{cvprblue}{rgb}{0.21,0.49,0.74}
\usepackage[pagebackref,breaklinks,colorlinks,allcolors=cvprblue]{hyperref}
\usepackage{booktabs}
\usepackage{multirow}

\def\confName{CVPR}
\def\confYear{2027}

\title{BlockAlign: Preference Optimization for Trajectory-Distilled Video Generation}

\author{Jingyuan Zhu \quad Biaolong Chen \quad Yangyang Qin \quad Shengyang Shen\\
Yaohua Wang \quad Weiming Bai \quad Miao Lu \quad Aixi Zhang\\[0.4em]
Alibaba Group\\
{\tt\small \{zhujingyuan.zjy, biaolong.cbl\}@alibaba-inc.com, \{qyy435227, shenshengyang.ssy\}@taobao.com}\\
{\tt\small \{xiachen.wyh, baiweiming.bwm, lumiao.lm, aixi.zhax\}@alibaba-inc.com}
}

\begin{document}
\maketitle
\begin{abstract}
Directly aligning trajectory-distilled video generators slashes rollout costs and optimizes the deployed few-step model without teacher fine-tuning or re-distillation. However, conventional flow-matching preference objectives assume single-head local velocity supervision, mismatching the finite-interval, block-average transition interface of trajectory-distilled generators such as Parallel Decoding Distillation (PDD) and Parametric Trajectory Distillation (PTD). Naively supervising individual unfused sub-interval velocity heads across all sub-steps rather than the fused block-average velocity at block starts (DirectAlign) induces a severe training--inference mismatch that degrades preference below the unaligned baseline. We introduce BlockAlign, an interface-consistent preference optimization framework that constructs states at valid block boundaries, queries the fused block-level velocity predictor, and supervises span-normalized transition residuals---seamlessly supporting DPO-, NFT-, and rollout-style objectives while preserving the native few-step sampler. We further build a unified post-training pipeline integrating offline pairs, reward-free semi-online continuations, and online rollouts with hierarchical preference selection and objective-aware block scheduling. Experiments on multi-reference video generation across MiniMax-H3 Ref2VA and an in-house model RefVid-V4 show that BlockAlign consistently outperforms unaligned distilled models and DirectAlign in structural plausibility, reference consistency, and visual quality. Notably, our approach unlocks PDD's $5\times$ few-step rollout speedup for $3\times$ faster end-to-end online training, achieving statistically significant human preference gains over PDD baselines ($54.0\%$ vs.\ $33.3\%$ wins/losses, or $61.8\%$ non-tie win rate on RefVid-V4; $51.3\%$ vs.\ $34.7\%$, or $59.7\%$ non-tie win rate on MiniMax-H3) and up to an $82.7\%$ win rate against DirectAlign.
\end{abstract}
    
\section{Introduction}
\label{sec:intro}

Reference-conditioned video generation~\cite{liu2025phantom,fei2025skyreels,kaleido2025,jiang2024videobooth,wang2024customvideo} requires models to follow text instructions, preserve reference identities, and synthesize temporally coherent motion with plausible structures. While large-scale pre-training provides strong visual priors, likelihood objectives imperfectly capture human preferences regarding reference fidelity, composition, anatomical correctness, and perceptual quality. Preference-based post-training addresses these gaps via pairwise comparisons, learned rewards, or differentiable feedback~\cite{diffusiondpo,diffusionnft,diffusionart,ddpo,draft,alignprop,videoscore,videodpo,instructvideo}, yet remains computationally prohibitive when online rollouts repeatedly traverse long denoising trajectories.

Directly post-training few-step distilled generators makes candidate rollouts far more affordable and optimizes the deployed model without costly teacher fine-tuning or re-distillation, which often weaken alignment gains during transfer~\cite{dopsd}. While existing few-step post-training efforts largely focus on distribution-matching or adversarial models (e.g., DMD~\cite{dmd,dmd2} and ADD~\cite{add}), trajectory distillation paradigms such as PDD~\cite{pdd} and PTD~\cite{ptd} have recently emerged as highly competitive alternatives for fast video generation by learning finite-interval transport, making their direct preference alignment critical yet unexplored.

However, achieving this is non-trivial: standard flow-matching preference objectives~\cite{flowmatching} cannot be directly applied to trajectory-distilled generators due to a fundamental shift in prediction interface and semantics---flow matching queries a single head for instantaneous velocities at arbitrary timestamps, whereas PDD and PTD parameterize interval-average velocities over finite denoising spans. Specifically, PDD and PTD share an identical finite-interval transition interface at deployment: PDD obtains the interval-average velocity by fusing its sub-interval velocity heads at each block boundary, whereas PTD directly retains its average-velocity head and discards its auxiliary distillation curvature head~\cite{ptd}. In this work, we directly post-train fused-head PDD generators—whose execution is strictly tied to valid block starts and coupled sub-interval heads—and further validate that our approach generalizes seamlessly to PTD generators in Appendix~\ref{sec:ptd_experiments}. Naively applying preference updates to individual unfused sub-interval velocity heads across sub-steps---termed \textbf{DirectAlign}---ignores block-entry conditioning and multi-head fusion, causing a severe post-training--inference mismatch that degrades preference below unaligned baselines.

To resolve this interface mismatch, we propose \textbf{BlockAlign}, which, to our knowledge, is the first framework to demonstrate effective direct preference post-training of few-step, trajectory-distilled video generators with finite-interval average-velocity predictions (including PDD and PTD) without teacher fine-tuning or re-distillation.
BlockAlign constructs training states at valid block boundaries, queries the fused block-level velocity operator, and supervises block transitions via span-normalized endpoint residuals. This unified interface seamlessly adapts DPO-style pairwise optimization~\cite{rafailov2023direct,diffusiondpo}, NFT-style reward-weighted regression~\cite{diffusionnft}, and rollout-based group feedback~\cite{diffusionart} to trajectory-distilled finite-block generators without altering their native few-step Euler sampler.

Building upon BlockAlign, we develop a unified post-training framework combining offline human pairs, reward-free semi-online partial continuations for texture and composition, and fully online rollouts to resolve compound failure modes in multi-reference video generation. Within this pipeline, hierarchical preference selection enforces anatomical and physical validity as a primary gate before ranking conditioning and visual quality, while an objective-aware block scheduler directs structural and reference updates to earlier blocks and texture/text refinement to later blocks.

Our contributions are summarized as follows:
\begin{itemize}
    \item We identify the semantic and gradient mismatch between unfused sub-head flow-matching preference updates and the fused block-average transition semantics of PDD video generators.
    \item We propose \textbf{BlockAlign}, a boundary-aligned, interface-consistent optimization framework that supervises fused-head PDD predictions at valid block boundaries via span-normalized transition residuals, supporting DPO, NFT, and rollout-based objectives without re-distillation.
    \item We design a multi-regime alignment pipeline integrating offline, reward-free semi-online, and online supervision with hierarchical preference selection and objective-aware block scheduling.
    \item Across RefVid-V4 and MiniMax-H3 Ref2VA~\cite{minimax2026h3}, BlockAlign unlocks PDD's $5\times$ few-step rollout speedup for $3\times$ faster end-to-end online training, achieving statistically significant human preference gains over PDD baselines ($54.0\%$ vs.\ $33.3\%$, or $61.8\%$ non-tie win rate on RefVid-V4; $51.3\%$ vs.\ $34.7\%$, or $59.7\%$ non-tie win rate on MiniMax-H3) and up to an $82.7\%$ win rate against DirectAlign without sacrificing diversity.
\end{itemize}
\section{Related Work}
\label{sec:related_work}

\paragraph{Preference alignment for diffusion-based video generation models.}
Diffusion alignment has been studied via pairwise preference optimization~\cite{diffusiondpo,d3po,spo}, policy-gradient reinforcement learning~\cite{ddpo,flowgrpo,dancegrpo}, differentiable reward backpropagation~\cite{xu2023imagereward,draft,alignprop,vader}, and reward-weighted regression~\cite{awm,diffusionnft,diffusionart}. In video generation, these paradigms have been extended with human and multidimensional feedback to improve temporal coherence, motion dynamics, and reference fidelity on standard multi-step video diffusion generators~\cite{videoscore,videodpo,instructvideo}. While Diffusion-APO~\cite{diffusionapo} applies DPO to a classifier-free guidance (CFG)-distilled video generator~\cite{ho2022classifier}, it only eliminates the dual-branch guidance overhead without compressing the multi-step denoising trajectory. BlockAlign is independent of the underlying preference rule, formulating an inference-consistent interface tailored to the finite-span, blockwise transitions of trajectory-distilled video generators.

\paragraph{Few-step distillation and direct post-training.}
Few-step generation has been achieved via progressive and consistency distillation~\cite{progressivedistillation,consistencymodels,lcm,ctm,tract}, flow straightening~\cite{instaflow}, adversarial or distribution matching~\cite{add,dmd,dmd2}, and learned finite-interval trajectory transport~\cite{piflow,ptd,alignyourflow,pdd}. Notably, PTD~\cite{ptd} and PDD~\cite{pdd} replace instantaneous velocity predictions with interval-average velocities over finite denoising spans, sharing the same deployed block-transition interface via PDD's sub-interval head fusion and PTD's average-velocity head (after discarding its auxiliary curvature head). Recent works directly post-train few-step models via on-policy self-distillation~\cite{dopsd,diffusionopd,opsdv,diffusionopsd}, pairwise sample optimization (PSO)~\cite{pso}, or few-step reward optimization~\cite{advfidelity,leapalign}. However, existing methods either retain costly teacher distillation or adversarial discriminators during post-training~\cite{dopsd,diffusionopd,opsdv,diffusionopsd,advfidelity}, rely on inter-step stochastic re-noising variance that excludes deterministic Euler transitions~\cite{pso}, or assume standard single-head instantaneous velocity models queried at arbitrary timestamps~\cite{leapalign}. None of them address the blockwise interval-average velocity interface of trajectory-distilled generators like PDD and PTD, which---whether sampled via deterministic Euler steps or stochastic re-noising---operate strictly at discrete block boundaries, with PDD further fusing multiple coupled sub-interval heads per block. BlockAlign bridges this gap without teacher distillation or sampler-variance constraints by querying valid block starts and directly supervising the fused block-average velocity operator across all blocks alongside objective-aware block scheduling.
\section{BlockAlign}
\label{sec:method}

We first contrast standard flow matching with fused-head PDD transitions (Sec.~\ref{sec:fm_to_pdd}), analyze why unfused sub-head updates mismatch PDD (Sec.~\ref{sec:mismatch}), and formulate the BlockAlign optimization interface (Sec.~\ref{sec:interface}).

\subsection{From Flow Matching to PDD}
\label{sec:fm_to_pdd}

Let $x_1$ denote a target video latent, $x_0 \sim \mathcal{N}(0, I)$ a Gaussian noise latent, and $c$ the conditioning signal. Under linear flow matching~\cite{flowmatching}, an intermediate state $x_t$ at $t \in [0, 1]$ and the target velocity $v^\star$ are defined as
\begin{equation}
    x_t = (1 - t)x_0 + t x_1, \quad v^\star = x_1 - x_0.
    \label{eq:fm_interp}
\end{equation}
A standard flow-matching model $v_\theta(x_t, t, c)$ predicts the instantaneous velocity at $t$ and advances $x_t$ over step size $\Delta t > 0$ via $\hat{x}_{t+\Delta t} = x_t + \Delta t \, v_\theta(x_t, t, c)$.

In contrast, PDD~\cite{pdd} parameterizes finite-interval block transitions. A $K$-step schedule $0 = t_0 < t_1 < \dots < t_K = 1$ is partitioned into $B$ contiguous blocks, where block $b \in \{1, \dots, B\}$ starts at $s_b = t_{i_b}$, ends at $e_b = t_{j_b}$ ($j_b = i_b + m_b$), and spans $h_b = e_b - s_b$ across $m_b$ elementary intervals. At block-start state $z_b = x_{s_b}$, PDD fuses its $m_b$ active sub-interval velocity heads $\{u_\theta^{(r)}(z_b, s_b, c)\}_{r=0}^{m_b-1}$ with schedule-fixed weights $\alpha_{b,r} = (t_{i_b+r+1} - t_{i_b+r}) / h_b$ into a single block-level average velocity predictor $\bar{v}_{\theta,b}(z_b, c) = \sum_{r=0}^{m_b-1} \alpha_{b,r} u_\theta^{(r)}(z_b, s_b, c)$, advancing $z_b$ to the block-end state $\hat{z}_{b+1}$ via the transition operator $F_{\theta,b}$:
\begin{equation}
    \hat{z}_{b+1} = F_{\theta,b}(z_b, c) = z_b + h_b \bar{v}_{\theta,b}(z_b, c).
    \label{eq:block_transition}
\end{equation}
Because $\{\alpha_{b,r}\}$ are fixed by the schedule, PDD's sub-interval heads merge directly into $\bar{v}_{\theta,b}$, matching the deployed interval-average transition interface of PTD~\cite{ptd} (which directly retains its average-velocity head after discarding its auxiliary curvature head). We post-train this fused-head PDD generator strictly at valid block starts $s_b$ over spans $h_b$, and demonstrate seamless generalization to PTD generators with step-specific adapters in Appendix~\ref{sec:ptd_experiments}.

\subsection{Why Naive Post-Training Mismatches PDD}
\label{sec:mismatch}

Conventional flow-matching preference objectives~\cite{diffusiondpo,diffusionnft} assume single-head instantaneous velocity semantics. Comparing the one-step prediction $\hat{x}_{t+\Delta t}$ against the reference state $x_{t+\Delta t} = x_t + \Delta t \, v^\star$ yields the normalized residual $(\hat{x}_{t+\Delta t} - x_{t+\Delta t}) / \Delta t = v_\theta(x_t, t, c) - v^\star$ and flow-matching (FM) squared $\ell_2$ error $\mathcal{E}^{\text{FM}}_\theta$:
\begin{equation}
    \mathcal{E}^{\text{FM}}_\theta(x_1, x_0, t, c) = \|v_\theta(x_t, t, c) - v^\star\|_2^2.
    \label{eq:fm_error}
\end{equation}
Here $\Delta t$ cancels out, making supervision at arbitrary $t$ consistent with standard single-head multi-step inference. However, directly applying Eq.~\eqref{eq:fm_error} to PDD across all $K$ sub-steps $t_{i_b+r}$ by supervising each unfused sub-head $u_\theta^{(r)}$ individually against $v^\star$---termed \textbf{DirectAlign}---mismatches PDD inference in two orthogonal ways:
\begin{enumerate}
    \item \textbf{Sub-step state conditioning vs.\ block-entry look-ahead:} In PDD distillation and inference, every sub-head $u_\theta^{(r)}$ ($r \ge 0$) is conditioned strictly on the block-start state $(z_b, s_b)$ to look ahead to $[t_{i_b+r}, t_{i_b+r+1}]$, whereas re-noising at intermediate sub-steps $(x_{t_{i_b+r}}, t_{i_b+r})$ feeds out-of-interface inputs never queried at deployment.
    \item \textbf{Instantaneous sub-head vs.\ fused block-average velocity semantics:} Even at $(z_b, s_b)$, PDD transitions via the fused interval-average velocity $\bar{v}_{\theta,b}(z_b, c) = \sum_{r=0}^{m_b-1}\alpha_{b,r}u_\theta^{(r)}(z_b, s_b, c)$. With sub-heads frozen, penalizing each unfused sub-head $u_\theta^{(r)}$ separately forces local instantaneous tangent velocities toward constant $v^\star$, erasing intra-block curvature and ignoring cross-head cancellation upon fusion (see Appendix~\ref{sec:limitations} for supplementary analysis).
\end{enumerate}

\subsection{BlockAlign Optimization Interface}
\label{sec:interface}

To resolve this interface mismatch, \textbf{BlockAlign} formulates preference supervision directly on native PDD blocks. Given a video latent $x_1$, noise $x_0 \sim \mathcal{N}(0, I)$, and block index $b \in \{1, \dots, B\}$, we construct the forward re-noised block-start state $z_b$ and linear reference endpoint $z_{b+1}^\star$:
\begin{align}
    z_b &= (1 - s_b)x_0 + s_b x_1, \label{eq:block_start_state} \\
    z_{b+1}^\star &= (1 - e_b)x_0 + e_b x_1 = z_b + h_b v^\star. \label{eq:block_end_state}
\end{align}
Note that Eqs.~\eqref{eq:block_start_state}--\eqref{eq:block_end_state} provide boundary-aligned, interface-consistent re-noising supervision along the standard linear path (which remains an approximation to the curved reverse-rollout state distribution $z_b^{\text{rollout}}$; we further study cached rollout-state supervision in Appendix~\ref{app:block_art}). Querying the fused-head PDD predictor at $(z_b, s_b, c)$ yields $\hat{z}_{b+1} = F_{\theta,b}(z_b, c)$ via Eq.~\eqref{eq:block_transition}. Normalizing the endpoint discrepancy by $h_b$ gives the interface-consistent block error $\mathcal{E}_\theta$:
\begin{align}
    \mathcal{E}_\theta(x_1, x_0, b, c)
    &= \left\| \frac{F_{\theta,b}(z_b, c) - z_{b+1}^\star}{h_b} \right\|_2^2 \notag \\
    &= \left\| \bar{v}_{\theta,b}(z_b, c) - v^\star \right\|_2^2.
    \label{eq:block_error}
\end{align}
Dividing by $h_b$ removes trivial step-size scaling across blocks of unequal duration and reduces the span-normalized endpoint error to the squared velocity residual of the fused predictor $\bar{v}_{\theta,b}$. Whereas DirectAlign (Eq.~\eqref{eq:fm_error}) supervises unfused sub-interval velocity heads $u_\theta^{(r)}$ individually across sub-steps against $v^\star = x_1 - x_0$, BlockAlign queries native block-entry states $(z_b, s_b)$ and supervises the fused block-average velocity $\bar{v}_{\theta,b}(z_b, c) = \sum_{r=0}^{m_b-1} \alpha_{b,r} u_\theta^{(r)}(z_b, s_b, c)$, backpropagating gradients into the shared backbone through the exact $\alpha_{b,r}$-weighted combination of frozen sub-head projections executed at inference. This boundary-aligned residual directly interfaces with diverse post-training surrogates as below.

\paragraph{BlockAlign-DPO.}
Given condition $c$ and a preference pair $(x_1^+, x_1^-)$, we sample a block index $b \sim p_{\text{obj}}(b)$ (Sec.~\ref{sec:block_scheduling}) and shared noise $x_0 \sim \mathcal{N}(0, I)$ to construct $(z_b^+, z_{b+1}^{\star,+})$ and $(z_b^-, z_{b+1}^{\star,-})$ via Eqs.~\eqref{eq:block_start_state}--\eqref{eq:block_end_state}. With frozen reference generator $\theta_{\text{ref}}$ and error difference $\Delta \mathcal{E}_b(x_1) = \mathcal{E}_\theta(x_1, x_0, b, c) - \mathcal{E}_{\theta_{\text{ref}}}(x_1, x_0, b, c)$, we formulate the BlockAlign-DPO surrogate objective~\cite{rafailov2023direct,diffusiondpo} for finite-interval block transitions as
\begin{align}
    \mathcal{L}_{\text{BA-DPO}}(\theta) = -\mathbb{E}_{c, (x_1^+, x_1^-), b, x_0} \Big[
    &\log \sigma \big( -\beta_{\text{DPO}} \big( \Delta \mathcal{E}_b(x_1^+) \notag \\
    &- \Delta \mathcal{E}_b(x_1^-) \big) \big) \Big],
    \label{eq:ba_dpo}
\end{align}
where $\sigma(\cdot)$ is the sigmoid function and $\beta_{\text{DPO}} > 0$ controls the preference margin. At the initial block ($b=1, s_1=0$), shared noise implies $z_1^+ = z_1^- = x_0$, so the quadratic norm terms in $\Delta \mathcal{E}_1$ cancel and yield a linear margin $\Delta \mathcal{E}_1(x_1^+) - \Delta \mathcal{E}_1(x_1^-) = -2\langle \bar{v}_{\theta,1}(x_0, c) - \bar{v}_{\text{ref},1}(x_0, c), x_1^+ - x_1^- \rangle$; at subsequent blocks ($b \ge 2$), $z_b^+$ and $z_b^-$ generally differ, so the quadratic terms do not cancel.

\paragraph{BlockAlign-NFT.}
Following DiffusionNFT~\cite{diffusionnft}, let $\theta_{\text{old}}$ denote the behavior PDD generator, $x_1$ a candidate with optimality reward $r \in [0, 1]$, and $z_b = (1 - s_b)x_0 + s_b x_1$. We form positive and negative implicit velocity branches $\bar{v}_{\theta,b}^+$ and $\bar{v}_{\theta,b}^-$ in the fused block velocity space with guidance scale $\beta_{\text{NFT}} > 0$:
\begin{align}
    \bar{v}_{\theta,b}^+(z_b, c) &= (1 - \beta_{\text{NFT}})\bar{v}_{\theta_{\text{old}},b}(z_b, c) + \beta_{\text{NFT}}\bar{v}_{\theta,b}(z_b, c), \label{eq:ba_nft_pos} \\
    \bar{v}_{\theta,b}^-(z_b, c) &= (1 + \beta_{\text{NFT}})\bar{v}_{\theta_{\text{old}},b}(z_b, c) - \beta_{\text{NFT}}\bar{v}_{\theta,b}(z_b, c). \label{eq:ba_nft_neg}
\end{align}
The block-level NFT surrogate $\mathcal{L}_{\text{BA-NFT}}$ is given by
\begin{align}
    \mathcal{L}_{\text{BA-NFT}}(\theta) = \mathbb{E}_{c, x_1, r, b, x_0} 
    &\Big[ w_b \Big(
    r \left\| \bar{v}_{\theta,b}^+(z_b, c) - v^\star \right\|_2^2 \notag \\
    &+ (1 - r) \left\| \bar{v}_{\theta,b}^-(z_b, c) - v^\star \right\|_2^2 \Big) \Big],
    \label{eq:ba_nft}
\end{align}
where $w_b$ is an optional block-dependent weighting factor.

\paragraph{BlockAlign-ART.}
Beyond pairwise and single-sample regression, BlockAlign naturally extends to group-relative rollout optimization following DiffusionART~\cite{diffusionart}. Given a group of $G$ online rollouts $\{x_1^{(g)}\}_{g=1}^G$ generated by $\theta_{\text{old}}$, we compute a reward-guided endpoint direction and minimize the span-normalized block endpoint residual from cached rollout states $z_b^{(g)}$ at valid boundaries $s_b$ (see Appendix~\ref{app:block_art} for the complete formulation and evaluation).

\section{Alignment Training Framework}
\label{sec:framework}

Building upon the BlockAlign interface, we develop a unified post-training framework that integrates multi-regime preference supervision (Sec.~\ref{sec:supervision_modes}), an objective-aware block scheduler (Sec.~\ref{sec:block_scheduling}), and a modular training procedure (Sec.~\ref{sec:training_procedure}) to address compound failure modes in multi-reference video generation.

\subsection{Multi-Regime Preference Supervision}
\label{sec:supervision_modes}

Multi-reference video generation faces diverse failure modes—instruction misinterpretation, identity drift, anatomical/physical distortion, poor composition, and degraded textures or text—that no single regime can resolve. We therefore combine three complementary alignment modes within BlockAlign:

\paragraph{Offline alignment.}
Pre-generated model rollouts are annotated by human raters to form fixed preference pairs $\mathcal{P}_{\text{off}} = (x_1^{\text{win}}, x_1^{\text{lose}})$ under condition $c$, providing stable foundational supervision for instruction following, multi-reference consistency, and structural plausibility.

\paragraph{Semi-online alignment for texture and composition.}
To empirically decouple high-frequency appearance refinement from global layout optimization, we design two reward-free semi-online modes via block-boundary partial noising and denoising inspired by SDEdit~\cite{sdedit}. Given a curated reference latent $x_1^{\text{ref}}$, we sample a boundary block $b_0 \in \{1, \dots, B\}$ and noise $\epsilon \sim \mathcal{N}(0, I)$ to construct $z_{b_0}^{\text{ref}}$ at $s_{b_0}$, then denoise from block $b_0$ to $B$ using the current PDD generator $F_{\theta,b}$ to obtain a partial continuation $\tilde{x}_1$:
\begin{align}
    z_{b_0}^{\text{ref}} &= (1 - s_{b_0})\epsilon + s_{b_0} x_1^{\text{ref}}, \label{eq:semi_anchor_noise} \\
    \tilde{x}_1 &= \left( F_{\theta,B}(\cdot, c) \circ \dots \circ F_{\theta,b_0}(\cdot, c) \right) \left( z_{b_0}^{\text{ref}} \right). \label{eq:semi_anchor_rollout}
\end{align}
Because coarse spatiotemporal structure is largely established by boundary $s_{b_0}$, $\tilde{x}_1$ empirically retains the coarse layout of $x_1^{\text{ref}}$ while exhibiting the current generator's surface rendering. Using $\tilde{x}_1$ as a pivot, we form two heuristic pairs without human or reward-model scoring:
\begin{itemize}
    \item \textbf{Texture-oriented ($\mathcal{P}_{\text{tex}} = (x_1^{\text{ref}}, \tilde{x}_1)$):} Pairing preferred reference $x_1^{\text{ref}}$ against continuation $\tilde{x}_1$—which share coarse layout but differ in details—biases updates toward finer textures, sharpness, and color fidelity.
    \item \textbf{Composition-oriented ($\mathcal{P}_{\text{comp}} = (\tilde{x}_1, \hat{x}_1^{\text{full}})$):} Pairing preferred continuation $\tilde{x}_1$ against a full-trajectory rollout $\hat{x}_1^{\text{full}}$ decoded from $\epsilon \sim \mathcal{N}(0, I)$ matches late-stage generator rendering statistics while transferring the stronger compositional priors of $x_1^{\text{ref}}$ with reduced texture confounding.
\end{itemize}
While not guaranteeing strict sample-wise disentanglement or noise-free labels, $\mathcal{P}_{\text{tex}}$ and $\mathcal{P}_{\text{comp}}$ provide complementary statistical biases in expectation without reward hacking, supporting both $\mathcal{L}_{\text{BA-DPO}}$ and binary $\mathcal{L}_{\text{BA-NFT}}$ ($r \in \{0, 1\}$).

\paragraph{Online alignment and hierarchical selection.}
To suppress policy-specific failure modes across the full trajectory, the behavior generator $\theta_{\text{old}}$ rolls out $G$ candidates $\{x_1^{(g)}\}_{g=1}^G$ from $\epsilon \sim \mathcal{N}(0, I)$ across all $B$ blocks per condition $c$, which a multidimensional judge scores to form pairs $\mathcal{P}_{\text{on}} = (x_1^{(g^+)}, x_1^{(g^-)})$ for $\mathcal{L}_{\text{BA-DPO}}$ or rewards $r^{(g)} \in [0, 1]$ for $\mathcal{L}_{\text{BA-NFT}}$. Since naive score averaging lets strong composition or reference similarity mask severe anatomical or physical defects (e.g., extra limbs or interpenetration), we enforce a \emph{hierarchical preference selection} rule: structural and physical plausibility serves as a primary gate that immediately rejects defective candidates, and only structurally valid videos are ranked by instruction following, reference consistency, composition, and visual quality.

\subsection{Objective-Aware Block Scheduler}
\label{sec:block_scheduling}

Different PDD blocks govern visual attributes at distinct granularities. Partitioning our 10-block ($B=10$) schedule by block-start noise $\sigma_b = 1 - s_b$ with thresholds at $\sigma_5 = 0.9231$ and $\sigma_8 = 0.6316$ (detailed schedules in Table~\ref{tab:block_schedule}, Appendix~\ref{sec:block_scheduling_ablations}), early blocks ($b \in \{1, \dots, 4\}$, $\sigma_b > 0.9231$) control global motion and coarse layout; middle blocks ($b \in \{5, 6, 7\}$, $0.6316 < \sigma_b \le 0.9231$) refine subject interactions, multi-reference identity binding, and composition; and late blocks ($b \in \{8, 9, 10\}$, $\sigma_b \le 0.6316$) govern high-frequency textures, color saturation, and text rendering. Uniformly sampling $b$ wastes capacity by applying structural updates to late texture blocks or fine-detail updates to early high-noise blocks. We therefore define an objective-aware categorical scheduler $p_{\text{obj}}(b)$ over complete PDD blocks $\{1, \dots, B\}$ with importance weights $\omega_{\text{obj}, b}$:
\begin{equation}
    p_{\text{obj}}(b) = \frac{\omega_{\text{obj}, b}}{\sum_{b'=1}^B \omega_{\text{obj}, b'}}, \quad b \in \{1, \dots, B\},
    \label{eq:block_scheduling}
\end{equation}
using a $2{:}1$ weight ratio for prioritized blocks:
\begin{itemize}
    \item \textbf{Early-to-mid emphasis} ($b \in \{1, \dots, 7\}$, $\sigma_b > 0.6316$): Prioritizes early and middle blocks for physical/anatomical plausibility, instruction following, and reference consistency ($\mathcal{P}_{\text{off}}, \mathcal{P}_{\text{on}}$).
    \item \textbf{Mid-to-late emphasis} ($b \in \{5, \dots, 10\}$, $\sigma_b \le 0.9231$): Prioritizes middle and late blocks for compositional layout, surface texture, color saturation, and text rendering ($\mathcal{P}_{\text{comp}}, \mathcal{P}_{\text{tex}}$).
\end{itemize}
Under any profile, $p_{\text{obj}}(b)$ always samples complete PDD blocks and supervises the span-normalized block error $\mathcal{E}_\theta(x_1, x_0, b, c)$ in Eq.~\eqref{eq:block_error}, preserving strict block-interface consistency.

\subsection{Overall Training Procedure}
\label{sec:training_procedure}
In each iteration, we sample a condition $c$ and objective $\text{obj}$, obtain supervision from the corresponding regime ($\mathcal{P}_{\text{off}}$, $\mathcal{P}_{\text{tex}}$, $\mathcal{P}_{\text{comp}}$, or hierarchically gated online rollouts $\mathcal{P}_{\text{on}}$), draw $b \sim p_{\text{obj}}(b)$ and $x_0 \sim \mathcal{N}(0, I)$ to form $z_b$ at $s_b$ (or reuse cached rollout states in BlockAlign-ART), and update the PDD generator via $\mathcal{L}_{\text{BA-DPO}}$ (Eq.~\eqref{eq:ba_dpo}) or $\mathcal{L}_{\text{BA-NFT}}$ (Eq.~\eqref{eq:ba_nft}). The regimes and block scheduler can be combined in multi-stage pipelines (offline/semi-online followed by online alignment) or tailored to individual backbones without altering the native PDD sampler.

\section{Experiments}
\label{sec:experiments}
We evaluate BlockAlign in e-commerce multi-reference video generation, which demands high material/texture fidelity, compositional layout, multi-reference consistency, and plausible human--product interactions while retaining few-step PDD inference efficiency. We test two backbones with distinct refinement needs: on our in-house 30B multimodal DiT (RefVid-V4), we combine offline pairs with texture- and composition-oriented semi-online supervision under BlockAlign-NFT to jointly enhance rendering, layout, and structure; on the 33B MiniMax-H3 Ref2VA~\cite{minimax2026h3} (henceforth ``MiniMax-H3''), which has stronger baseline appearance, we apply online BlockAlign-DPO to eliminate anatomical and structural defects. These setups validate complementary regimes rather than cross-backbone strategy comparisons (see Appendix~\ref{sec:limitations} for detailed discussion).

\subsection{Experimental Setup}
\label{sec:experimental_setup}

\textbf{Models and implementation.} Both models initialize from 10-step ($B=10$) PDD~\cite{pdd} checkpoints and retain their native inference schedules (detailed in Table~\ref{tab:block_schedule}, Appendix~\ref{sec:block_scheduling_ablations}) after post-training, generating $768\times 1280$ videos of 5--15\,s at 24\,fps. We apply LoRA~\cite{lora} ($r=128, \alpha=128$) to attention and MLP layers on 96 NVIDIA H200 GPUs using Muon~\cite{liu2025muon} (learning rate $10^{-5}$, global batch size 12 across 12 DP groups of SP=8) for 1,500 steps, with $\beta_{\text{NFT}}=1.0$ ($w_b=1$) and $\beta_{\text{DPO}}=50$; $\theta_{\text{old}}$ syncs with $\theta$ every step, while $\theta_{\text{ref}}$ stays frozen at initial PDD weights. We also evaluate a 10-step PTD~\cite{ptd} generator in Appendix~\ref{sec:ptd_experiments}.

\begin{figure*}[t]
\centering
\includegraphics[width=\textwidth]{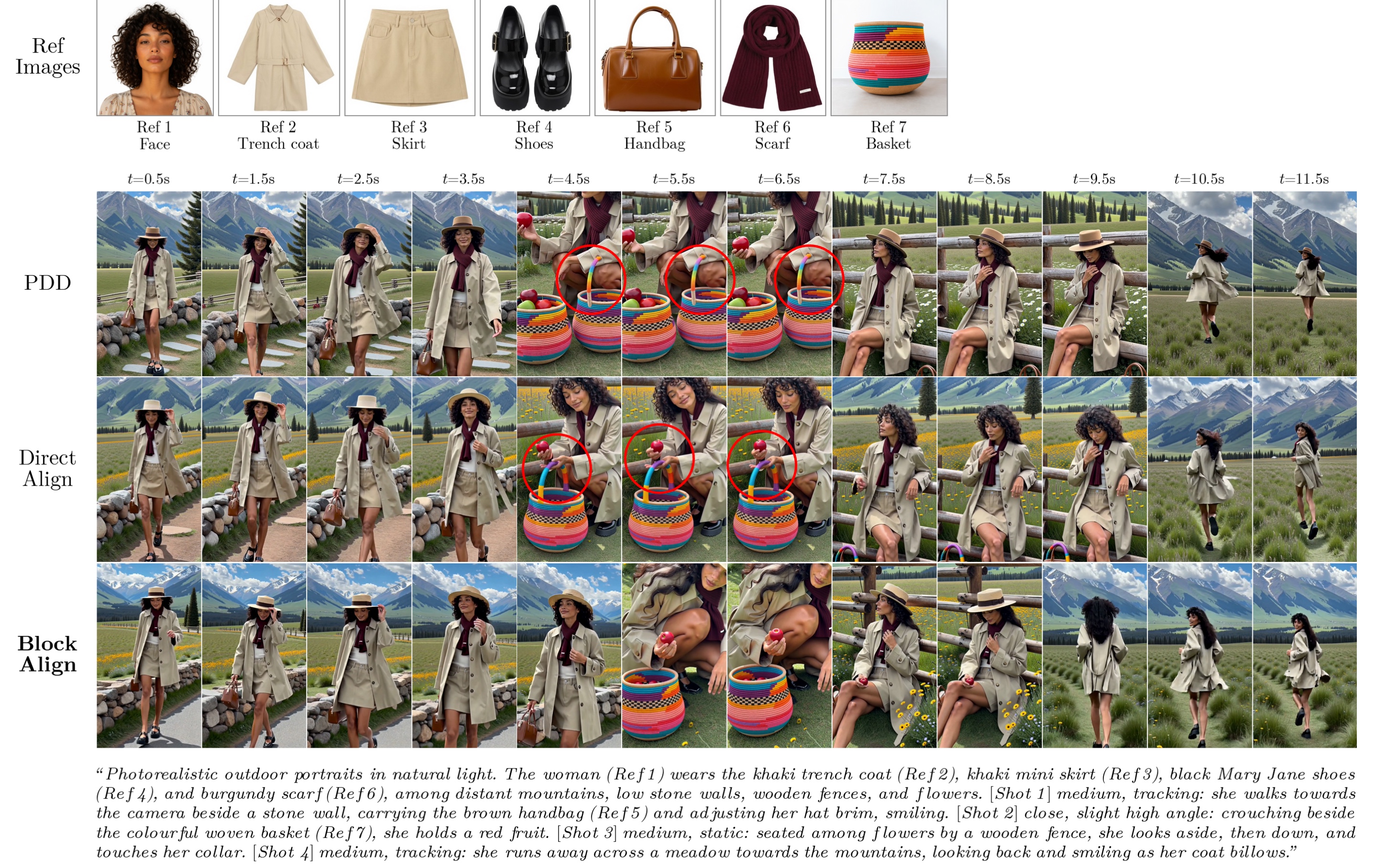}
\caption{Qualitative comparison on RefVid-V4. All methods use identical inputs, random seeds, inference settings, and matched timestamps. Red circles highlight representative baseline errors in the geometry and spatial relationships among the hands, apples, and basket handles. BlockAlign produces more plausible interactions while preserving the referenced identity, clothing, accessories, and basket. It also yields clearer, more saturated frames with fewer patch-like texture artifacts. Best viewed zoomed in.}
\label{fig:qualitative_hh_case1}
\end{figure*}

\begin{figure*}[t]
\centering
\includegraphics[width=\textwidth]{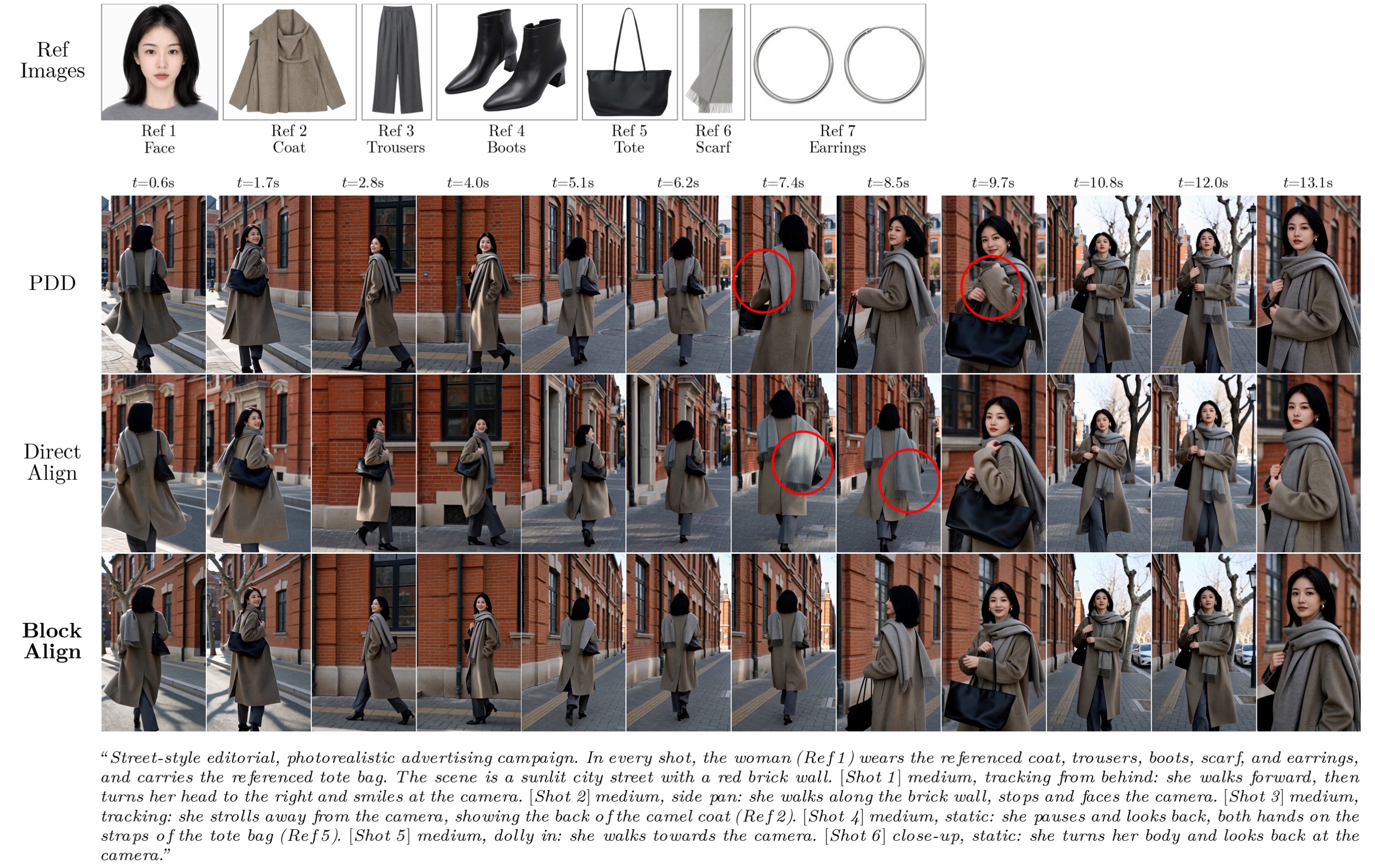}
\caption{Qualitative comparison on MiniMax-H3. Red circles highlight representative baseline errors in arm, sleeve, scarf, and garment structure. BlockAlign produces more coherent anatomy and garment geometry while preserving the referenced face, coat, trousers, boots, tote bag, scarf, and earrings across matched timestamps. Best viewed zoomed in.}
\label{fig:qualitative_h3_case2}
\end{figure*}

\textbf{Training data and feedback.} Both BlockAlign and DirectAlign use the regimes and scheduler of Secs.~\ref{sec:supervision_modes}--\ref{sec:block_scheduling}. RefVid-V4 uses 5,000 human-annotated pairs $\mathcal{P}_{\text{off}}$ and 10,000 curated reference videos across 20,000 conditions, sampling $\mathcal{P}_{\text{off}}$, $\mathcal{P}_{\text{tex}}$, and $\mathcal{P}_{\text{comp}}$ at 2:1:1 with boundary $b_0=5$ ($\sigma_{b_0}=0.9231$) and binary BlockAlign-NFT rewards $r \in \{1,0\}$. MiniMax-H3 rolls out $G=4$ online candidates per condition and queries \texttt{gemini-4-argon}~\cite{gemini4argon} under hierarchical selection to form $\mathcal{P}_{\text{on}}$ for BlockAlign-DPO (skipping updates if all $G=4$ candidates fail the structural gate or tie).

\textbf{Evaluation protocol.} We evaluate on 150 held-out e-commerce test conditions (disjoint from training), each pairing 3--9 reference images (characters, garments, accessories, props, and backgrounds) with a 5--15\,s script across 3--10 shots involving complex human actions, human--object--environment interactions, and multi-background transitions. Both \texttt{gemini-4-argon}~\cite{gemini4argon} and \texttt{gpt-6-astra}~\cite{openai2026gpt6astra} score uniformly sampled frames (scaled to duration within context limits) given all references, prompts, and rubrics. Retaining and freezing all original sub-heads $\{u_\theta^{(r)}\}_{r=0}^{m_b-1}$ (updating only shared Attention/MLP LoRA adapters from identical PDD weights), DirectAlign and BlockAlign share identical regimes and block priorities $p_{\text{obj}}(b)$ (uniform vs.\ $p_{\text{obj}}(b)$ ablated in Appendix~\ref{sec:block_scheduling_ablations}) under identical single-pass backbone evaluations, differing solely in the supervised output head interface: DirectAlign supervises unfused sub-interval heads $u_\theta^{(r)}$ (Eq.~\eqref{eq:fm_error}), whereas BlockAlign supervises the fused block-average velocity $\bar{v}_{\theta,b}(z_b, c) = \sum_{r=0}^{m_b-1} \alpha_{b,r} u_\theta^{(r)}(z_b, s_b, c)$ at $(z_b, s_b)$ (Eq.~\eqref{eq:block_error}). At inference, both use the fused output $\bar{v}_{\theta,b}$ under identical 10-step PDD schedules and seeds.

\subsection{Qualitative Results}
\label{sec:qualitative_results}

Figures~\ref{fig:qualitative_hh_case1} and~\ref{fig:qualitative_h3_case2} compare PDD, DirectAlign, and BlockAlign under identical inputs, random seeds, inference settings and matched timestamps. Red circles mark representative errors. More cases are provided in Appendix~\ref{sec:additional_qualitative}.

\paragraph{RefVid-V4.}
In Figure~\ref{fig:qualitative_hh_case1} (multi-shot hand--apple--basket interactions), PDD produces implausible hand/handle geometry and DirectAlign distorts finger--object relations, whereas BlockAlign yields physically plausible interactions, preserves all references, and delivers sharper, more saturated textures.

\paragraph{MiniMax-H3.}
In Figure~\ref{fig:qualitative_h3_case2} (viewpoint transitions with scarf/tote interactions), PDD and DirectAlign suffer from arm, sleeve, and garment distortions, whereas BlockAlign maintains coherent anatomy and garment structure while faithfully preserving the referenced identity and outfit.

\begin{table*}[t]
\centering
\small
\caption{Automated absolute assessment scores (1--5 scale; higher is better). Mean is the unweighted average over the five dimensions.}
\label{tab:auto_absolute}
\begin{tabular}{llrrrrrr}
\toprule
Backbone & Method & SC & PF & VQ & DF & ST & Mean \\
\midrule
& PDD         & 3.35 & 3.79 & 3.93 & 3.55 & \textbf{3.75} & 3.67 \\
RefVid-V4 & DirectAlign & 3.37 & 3.73 & 3.93 & 3.61 & 3.71 & 3.67 \\
& BlockAlign  & \textbf{3.41} & \textbf{3.81} & \textbf{3.94} & \textbf{3.64} & \textbf{3.75} & \textbf{3.71} \\
\midrule
& PDD         & 3.41 & 3.62 & \textbf{3.96} & 3.60 & 3.65 & 3.65 \\
MiniMax-H3 & DirectAlign & 3.40 & 3.61 & 3.95 & 3.55 & 3.59 & 3.62 \\
& BlockAlign  & \textbf{3.50} & \textbf{3.65} & \textbf{3.96} & \textbf{3.75} & \textbf{3.66} & \textbf{3.70} \\
\bottomrule
\end{tabular}
\end{table*}

\begin{table*}[t]
\centering
\small
\caption{Adjusted pairwise win rates (\%) for the first method against the second; $50\%$ denotes parity. Agr.\ denotes bidirectional agreement averaged over dimensions.}
\label{tab:auto_pairwise}
\begin{tabular}{llrrrrrrr}
\toprule
Backbone & Pair & SC & PF & VQ & DF & ST & Mean & Agr. \\
\midrule
& DirectAlign vs.\ PDD & 45.7 & 42.8 & 42.0 & 50.8 & 43.3 & 44.9 & 55.9 \\
RefVid-V4 & BlockAlign vs.\ PDD & 52.5 & 49.2 & 57.5 & 56.5 & 53.3 & 53.8 & 59.6 \\
& BlockAlign vs.\ DirectAlign & 58.7 & 60.8 & 66.7 & 58.3 & 60.3 & 61.0 & 68.0 \\
\midrule
& DirectAlign vs.\ PDD & 38.0 & 41.7 & 37.3 & 38.2 & 43.3 & 39.7 & 57.7 \\
MiniMax-H3 & BlockAlign vs.\ PDD & 52.5 & 55.2 & 52.8 & 55.3 & 53.8 & 53.9 & 57.2 \\
& BlockAlign vs.\ DirectAlign & 62.5 & 61.7 & 67.3 & 65.7 & 62.0 & 63.8 & 58.9 \\
\bottomrule
\end{tabular}
\end{table*}

\subsection{Quantitative Results}
\label{sec:quantitative_results}

\paragraph{Automated assessment.}
We evaluate all methods on the 150 test conditions across subject consistency (SC), prompt following (PF), visual quality (VQ), deformation (DF), and shot transitions (ST) using \texttt{gpt-6-astra}~\cite{openai2026gpt6astra}. We report 1--5 absolute scores (Table~\ref{tab:auto_absolute}; higher indicates better quality and fewer DF defects) and bidirectional pairwise win rates (Table~\ref{tab:auto_pairwise}; scoring wins/ties/losses as $1/0.5/0$ across both orders to mitigate positional bias, with agreement Agr.). BlockAlign achieves the highest mean score on both RefVid-V4 ($3.71$ vs.\ $3.67$ for PDD/DirectAlign) and MiniMax-H3 ($3.70$ vs.\ $3.65$ for PDD and $3.62$ for DirectAlign), led by gains in DF ($+0.09, +0.15$) and SC ($+0.06, +0.09$). In pairwise comparisons, DirectAlign falls below PDD ($44.9\%$ and $39.7\%$ mean win rates), whereas BlockAlign consistently beats PDD ($53.8\%$ and $53.9\%$) and DirectAlign ($61.0\%$ and $63.8\%$). Because VLM judges score sampled frames holistically, transient localized defects only moderately shift full-video scalar averages (+0.04--0.05) and fine-grained dimensions (e.g., VQ) can exhibit order sensitivity, motivating our bidirectional averaging and human evaluation.

Benchmark suites such as VBench~\cite{huang2024vbench,huang2025vbench++,zheng2025vbench}, EvalCrafter~\cite{liu2024evalcrafter}, and T2V-CompBench~\cite{xie2024t2vcompbench} lack multi-reference inputs; instead, we evaluate zero-shot generalization on two public video generation benchmarks in Appendix~\ref{sec:public_benchmarks}: OpenS2V-Eval~\cite{yuan2025opens2v} ($360$ test cases across four subject--scene domains) and OmniVBench~\cite{omnivbench2026} ($813$ test cases across seven omni-reference task families). BlockAlign consistently improves aesthetics, motion smoothness, prompt alignment, and human subject consistency over PDD on OpenS2V-Eval, and achieves the best video quality, technical fidelity, and aesthetics across all seven OmniVBench task families ($7/7$) while avoiding DirectAlign's consistent technical, aesthetic, and smoothness degradation.

\begin{figure}[t]
    \centering
    \includegraphics[width=\linewidth,keepaspectratio]{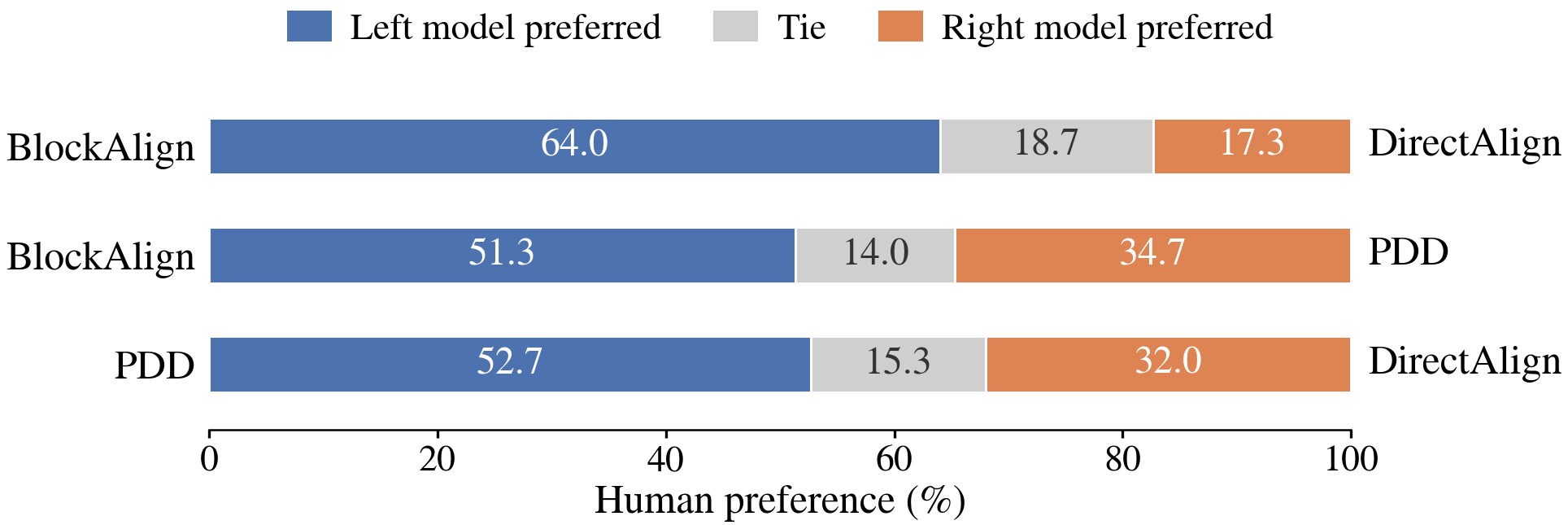}
    \vspace{4pt}
    \includegraphics[width=\linewidth,keepaspectratio]{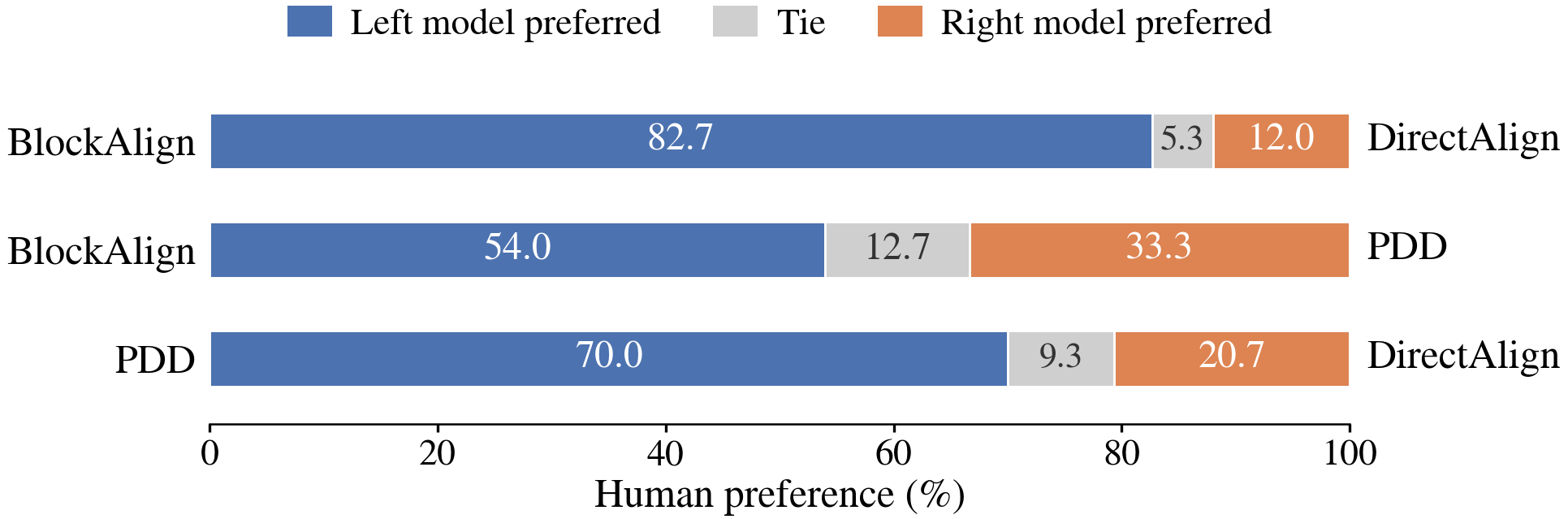}
    \caption{Human preference evaluation on MiniMax-H3 (top) and RefVid-V4 (bottom). Bars show condition-level win, tie, and loss rates over 150 comparisons per method pair, aggregated from five independent annotators per pair. On both backbones, BlockAlign is consistently preferred over both DirectAlign and the original PDD model, while PDD is preferred over DirectAlign.}
    \label{fig:h3_user_study}
\end{figure}

\paragraph{Human evaluation.}
We conduct blind pairwise human preference studies on RefVid-V4 and MiniMax-H3 over all $N=150$ test conditions across all method pairs among PDD, DirectAlign, and BlockAlign. Five annotators independently evaluate each pair on visual quality, multi-reference fidelity, prompt adherence, and structural plausibility to judge overall usability (A better, B better, or tie). To avoid inflating sample size across raters, statistics are clustered at the condition level ($N=150$): a method wins a condition only when it receives strictly more votes than the other without ties forming a majority (and ties otherwise).

As shown in Figure~\ref{fig:h3_user_study}, human raters consistently prefer BlockAlign over both baselines. On RefVid-V4, BlockAlign achieves $81/50/19$ wins/losses/ties against PDD ($54.0\%$ vs.\ $33.3\%$, $12.7\%$ ties), yielding a $60.3\%$ win-plus-half-tie score and a $61.8\%$ non-tie win rate (condition-level $95\%$ Wilson CI: $[53.3\%, 69.7\%]$, exact binomial $p=0.0085$), and beats DirectAlign with an $82.7\%$ win rate ($12.0\%$ losses), while PDD outperforms DirectAlign ($70.0\%$ vs.\ $20.7\%$). On MiniMax-H3, BlockAlign similarly outperforms PDD with $77/52/21$ wins/losses/ties ($51.3\%$ vs.\ $34.7\%$, $14.0\%$ ties), corresponding to a $58.3\%$ win-plus-half-tie score and a $59.7\%$ non-tie win rate ($95\%$ Wilson CI: $[51.1\%, 67.8\%]$, $p=0.0342$), while beating DirectAlign ($64.0\%$ vs.\ $17.3\%$). The stronger margins in human evaluation relative to scalar VLM averages reflect that human raters heavily penalize transient anatomical defects, interpenetration, and identity drift that are diluted in frame-averaged automated scores.

\paragraph{Generation diversity.}
To verify that post-training does not induce mode collapse, we evaluate multi-seed sample diversity using VideoMAE~v2~\cite{wang2023videomae} and V-JEPA~2~\cite{assran2025v} embeddings. Across both backbones, BlockAlign preserves feature dispersion and pairwise distances comparable to the base PDD models, confirming that its quality gains do not sacrifice diversity (see quantitative results in Appendix~\ref{sec:diversity}).

\subsection{Rollout and Training Efficiency}
\label{sec:efficiency}

BlockAlign enables effective preference optimization directly on fused-head PDD generators without DirectAlign's preference degradation. On 96 H200 GPUs (12 DP groups of SP=8, global batch size 12 at 1 condition per group), 10-step PDD sampling cuts per-video generation time (768p, 15\,s MiniMax-H3) from ${\sim}5$\,min (50-step teacher) to ${\sim}1$\,min ($5\times$ rollout speedup). With $G=4$ rollouts per condition, per-step latency drops from ${\sim}24$\,min to ${\sim}8$\,min, cutting 1,500-step training time from an extrapolated 600 hours to 200 measured hours ($3\times$ end-to-end training speedup; see latency breakdown in Appendix~\ref{sec:supp_rollout_efficiency}).

\section{Conclusion}
\label{sec:conclusion}

We present BlockAlign, a block-interface-consistent preference optimization framework for trajectory-distilled video generators such as PDD~\cite{pdd} and PTD~\cite{ptd}. By supervising interval-average velocity predictors at valid block boundaries via span-normalized transition residuals, BlockAlign resolves the post-training--inference mismatch of unfused sub-head flow-matching updates without altering the native few-step sampler. Combined with our multi-regime supervision pipeline, hierarchical preference selection, and objective-aware block scheduling, BlockAlign substantially improves structural plausibility, reference fidelity, and human preference across distinct trajectory-distilled backbones while translating $5\times$ faster few-step rollouts into a $3\times$ end-to-end online training speedup.

{
    \small
    \bibliographystyle{ieeenat_fullname}
    \bibliography{main}
}

\clearpage
\appendix
\setcounter{page}{1}
\maketitlesupplementary

This supplementary material provides the experimental design rationale and limitations (Appendix~\ref{sec:limitations}), the extension and evaluation of BlockAlign on Parametric Trajectory Distillation (PTD) (Appendix~\ref{sec:ptd_experiments}), the formulation and evaluation of BlockAlign-ART (Appendix~\ref{app:block_art}), additional qualitative comparisons (Appendix~\ref{sec:additional_qualitative}), public video generation benchmark evaluations on OpenS2V-Eval and OmniVBench (Appendix~\ref{sec:public_benchmarks}), generation diversity analyses (Appendix~\ref{sec:diversity}), rollout and training efficiency details (Appendix~\ref{sec:supp_rollout_efficiency}), and block scheduling specifications and ablations (Appendix~\ref{sec:block_scheduling_ablations}).

\section{Discussion and Limitations}
\label{sec:limitations}

\paragraph{Computational cost and modular ablation design.}
Post-training large-scale multi-reference video generators at $768\times 1280$ resolution for 5--15\,s durations (24\,fps) remains computationally intensive even with few-step PDD rollouts, requiring 96 NVIDIA H200 GPUs. This substantial compute footprint precludes exhaustive grid ablations over every combination of supervision regimes and hyperparameters across both backbones. Instead, our experimental design isolates the core technical contributions through targeted, complementary comparisons:
\begin{itemize}
    \item \textbf{Core optimization interface:} Across both RefVid-V4 and MiniMax-H3, the strictly controlled comparison between \textbf{DirectAlign} and \textbf{BlockAlign} (sharing identical initial checkpoints, training data, preference feedback, objectives, block priorities, LoRA parameters, and training budgets) directly serves as the primary ablation isolating our boundary-aligned, fused interval-average block supervision (Eq.~\eqref{eq:block_error} in the main paper) from naive unfused sub-head flow-matching updates across sub-steps (Eq.~\eqref{eq:fm_error}).
    \item \textbf{Modular regime validation:} Rather than redundantly ablating standard offline alignment ($\mathcal{P}_{\text{off}}$)—whose role as a foundational anchor is well established—we allocate our compute budget to validate the proposed supervision regimes on the backbones best suited to diagnose them. Specifically, MiniMax-H3 isolates fully online alignment ($\mathcal{P}_{\text{on}}$) with hierarchical preference selection to verify structural and anatomical defect suppression, whereas RefVid-V4 validates the synergy of our reward-free texture-oriented ($\mathcal{P}_{\text{tex}}$) and composition-oriented ($\mathcal{P}_{\text{comp}}$) semi-online continuations in improving surface rendering and spatial layout, complemented by our controlled ablation of the objective-aware block scheduler in Appendix~\ref{sec:block_scheduling_ablations}.
\end{itemize}

\paragraph{Detailed analysis of unfused sub-head supervision and single-subhead degeneracy.}
As analyzed in Sec.~\ref{sec:mismatch} of the main paper, \textbf{DirectAlign} applies standard flow-matching preference updates across all $K = \sum_{b=1}^B m_b = 50$ sub-steps by supervising the unfused sub-interval velocity heads ($\{u_\theta^{(r)}\}_{r=0}^{m_b-1}$ within each block $b$). Here we further detail the two orthogonal mismatches of supervising all $K$ sub-heads individually, and explain why restricting unfused supervision to only the initial sub-head at block starts is theoretically an even more degenerate special case:
\begin{enumerate}
    \item \textbf{Look-ahead conditioning mismatch and cross-head Jensen gap in DirectAlign (all $K$ sub-heads):} First, on the input side, every sub-head $u_\theta^{(r)}(z_b, s_b, c)$ ($r \in \{0, \dots, m_b-1\}$) within block $b$ is conditioned strictly on the \emph{block-entry state and timestamp} $(z_b, s_b)$ (the start of the first sub-interval $t_{i_b}$) during PDD distillation and inference to look ahead and predict the future sub-interval velocity over $[t_{i_b+r}, t_{i_b+r+1}]$. Re-noising latents at intermediate sub-steps $(x_{t_{i_b+r}}, t_{i_b+r})$ ($r \ge 1$) feeds out-of-interface inputs never queried at deployment. Second, on the output side, even if evaluated at $(z_b, s_b)$, penalizing each local instantaneous sub-head separately against the constant secant target $v^\star = x_1 - x_0$ differs from BlockAlign's fused residual by the intra-block cross-head variance (Jensen's gap):
    \begin{small}
    \begin{align}
        \sum_{r=0}^{m_b-1} & \alpha_{b,r}  \bigl\|u_\theta^{(r)}(z_b, s_b, c) - v^\star\bigr\|_2^2
        \;=\; \bigl\|\bar{v}_{\theta,b}(z_b, c) - v^\star\bigr\|_2^2 \notag \\
        &+ \sum_{r=0}^{m_b-1} \alpha_{b,r}\bigl\|u_\theta^{(r)}(z_b, s_b, c) - \bar{v}_{\theta,b}(z_b, c)\bigr\|_2^2.
        \label{eq:jensen}
    \end{align}
    \end{small}
    Because PDD sub-heads capture distinct local tangent velocities along a curved trajectory whose deviations from the interval average $\bar{v}_{\theta,b}$ naturally cancel upon linear fusion, penalizing the second term forces all $m_b$ sub-heads toward the same constant vector $v^\star$, erasing intra-block curvature specialization and distorting preference gradients under nonlinear objectives (e.g., $\log\sigma$ in DPO).
    \item \textbf{Theoretical degeneracy of initial-subhead-only supervision at block starts:} A naive alternative to avoid intermediate sub-step timestamps $t_{i_b+r}$ ($r \ge 1$) would be to query only valid block starts $(z_b, s_b)$ while supervising solely the primary initial-subinterval head $u_\theta^{(0)}(z_b, s_b, c)$ (which approximates the entry instantaneous velocity over $[t_{i_b}, t_{i_b+1}]$). However, with all sub-head linear projections frozen, updating the shared backbone solely through $u_\theta^{(0)}$ would leave the remaining $m_b-1$ look-ahead sub-heads $\{u_\theta^{(r)}(z_b, s_b, c)\}_{r=1}^{m_b-1}$---which carry $1 - \alpha_{b,0} = 80\%$ of the total block transition weight under $m_b = 5$ uniform sub-steps---completely unconstrained by preference supervision. Consequently, their outputs can shift uncontrolled as shared backbone features change, risking even more severe trajectory drift when fused into $\bar{v}_{\theta,b}(z_b, c) = \sum_{r=0}^{m_b-1}\alpha_{b,r}u_\theta^{(r)}(z_b, s_b, c)$ at inference. By contrast, BlockAlign queries valid block starts $(z_b, s_b)$ and directly supervises the fused block-average operator $\bar{v}_{\theta,b}$, resolving both mismatches while seamlessly generalizing to single-head PTD generators.
\end{enumerate}

\paragraph{Reward model bottleneck and extension to score-based or reward-gradient paradigms.}
Conceptually, our BlockAlign principle—replacing single-head instantaneous velocity supervision with boundary-aligned, finite-span block transitions—clarifies how broader score-based and reward-gradient paradigms can be adapted to trajectory-distilled generators, alongside the practical considerations involved:
\begin{itemize}
    \item \textbf{Advantage-weighted regression (e.g., AWM~\cite{awm}):} Rather than weighting instantaneous velocity errors at arbitrary timestamps, adapting AWM to PDD/PTD requires constructing latent states $z_b$ at valid block-start timestamps $s_b$ and directly weighting our span-normalized block error $E_\theta$ (Eq.~\eqref{eq:block_error} in the main paper) with group-normalized scalar advantages, making it a natural advantage-weighted variant of our BlockAlign-NFT/DPO formulation.
    \item \textbf{Stochastic policy-gradient RL (e.g., Flow-GRPO~\cite{flowgrpo} and DanceGRPO~\cite{dancegrpo}):} Adapting GRPO requires formulating Gaussian transition likelihoods at valid block boundaries around the finite-span block transition operator $F_{\theta,b}(z_b, c) = z_b + h_b \bar{v}_{\theta,b}(z_b, c)$ (Eq.~\eqref{eq:block_transition} in the main paper, where $h_b$ denotes the block span, $\bar{v}_{\theta,b}$ the interval-average velocity prediction, and $c$ the conditioning input) and injecting stochastic exploration noise at intermediate block endpoints during rollouts. However, when PDD/PTD generators are deployed with deterministic Euler transitions, requiring stochastic rollout transitions strictly for policy-gradient estimation alters the native inference trajectory and introduces sensitivity to the exploration variance schedule. In contrast, our implemented BlockAlign-ART (Appendix~\ref{app:block_art}) similarly exploits group-relative rollout advantages while remaining directly compatible with deterministic Euler PDD rollouts without requiring intermediate noise injection.
    \item \textbf{Differentiable reward-gradient backpropagation (e.g., ReFL~\cite{xu2023imagereward} and VADER~\cite{vader}):} Instead of extrapolating clean targets from arbitrary timestamps via instantaneous velocities, adapting ReFL requires initiating denoising strictly at valid block boundaries $s_b$, stepping through the blockwise transition operators $F_{\theta,b}$ to reach the terminal video latent, and backpropagating gradients from a differentiable reward model across the unrolled block boundaries. In practice, applying this to long, high-resolution videos may face challenges in memory overhead, incompatibility with black-box VLM judges, and potential sensitivity to reward over-optimization when backpropagating directly through imperfectly calibrated reward networks.
\end{itemize}
Beyond these algorithmic considerations, all three paradigms share a primary practical bottleneck: they place stringent demands on the reward model's continuous scoring calibration or gradient quality. For already strong, large-scale video generators (our 30B RefVid-V4 and 33B MiniMax-H3)—especially in complex multi-reference e-commerce scenarios involving fine-grained human--object interactions, multi-garment identity preservation, and subtle physical or anatomical constraints—we found it difficult in practice to obtain an off-the-shelf video reward model accurate enough to drive fine-grained score-based or gradient-based updates beyond the base models' quality ceiling. Consequently, we prioritize noise-resilient supervision in our main experiments (reward-free binary semi-online pairs on RefVid-V4 and gated extreme pairwise selection on MiniMax-H3) and restrict direct scalar reward guidance to our 400-step BlockAlign-ART study. Developing more discriminative, well-calibrated video reward models and systematically exploring block-aligned adaptations of AWM, GRPO, and ReFL on strong trajectory-distilled backbones remain promising directions for future work.

\paragraph{Backbone interface.}
Both our in-house 30B multimodal DiT (\textbf{RefVid-V4}) and the open-source 33B \textbf{MiniMax-H3}~\cite{minimax2026h3} operate as latent flow-matching video generators conditioned on text prompts and multiple reference images $c$, distilled via CFG distillation and PDD~\cite{pdd} into $B = 10$ fused-head transition blocks. Because BlockAlign interacts solely with the block-boundary latent states $z_b$ and fused velocity outputs $\bar{v}_{\theta,b}(z_b, c)$, it remains completely agnostic to the internal conditioning architecture of each backbone.

\section{Extension to PTD Generators}
\label{sec:ptd_experiments}

In the main paper, we formulate and evaluate BlockAlign primarily on fused-head Parallel Decoding Distillation (PDD)~\cite{pdd} generators. To further demonstrate that BlockAlign serves as a general, paradigm-wide preference optimization interface for trajectory-distilled video models, we extend our experiments to Parametric Trajectory Distillation (PTD)~\cite{ptd} on the MiniMax-H3 backbone.

\subsection{PDD vs.\ PTD: Formulation and Architecture}
\label{subsec:pdd_vs_ptd}

\paragraph{Distillation mechanism comparison.}
\label{sec:pdd_vs_ptd_formulation}
Both PDD~\cite{pdd} and PTD~\cite{ptd} are pure trajectory-based distillation methods that compress multi-step flow-matching trajectories into $B$ finite-interval transitions---referred to as \emph{hops} in PTD (corresponding to \emph{blocks} in PDD)---without relying on distribution-matching or adversarial losses. For clarity, we adopt the PTD terminology and refer to each finite transition interval $b \in \{1, \dots, B\}$ spanning $[s_b, e_b]$ with duration $h_b = e_b - s_b$ as a \emph{hop} throughout this section. Their primary distinction lies in how intra-hop curvature is parameterized and supervised during the distillation phase:
\begin{itemize}
    \item \textbf{PDD (Discrete sub-interval decomposition):} PDD discretizes the time domain into a fine grid of $K$ elementary sub-intervals and expands the final linear layer into $K$ parallel prediction heads. Within hop (block) $b \in \{1, \dots, B\}$, PDD simultaneously predicts $m_b$ discrete sub-interval velocities $\{u_\theta^{(r)}(z_b, s_b, c)\}_{r=0}^{m_b-1}$ in a single forward pass, regressing them onto teacher Runge--Kutta solver targets. At inference, because advancing across the entire interval requires only the span-weighted average velocity, the $m_b$ linear heads are mathematically fused into a single linear projection layer that outputs the interval-average velocity $\bar{v}_{\theta,b}(z_b, c) = \sum_{r=0}^{m_b-1} \alpha_{b,r} u_\theta^{(r)}(z_b, s_b, c)$.
    \item \textbf{PTD (Continuous polynomial parameterization):} Rather than discretizing sub-intervals, PTD parameterizes each finite denoising hop $[s_b, e_b]$ as a continuous polynomial curve. In a single forward evaluation at hop start $(z_b, s_b)$, the student predicts the interval-average velocity $\bar{v}_{\theta,b}(z_b, c)$ alongside $M$ Legendre polynomial curvature coefficients emitted by an auxiliary curvature head. Teacher instantaneous velocities are queried directly along this student-predicted curve to jointly supervise both heads via analytic basis derivatives (with an optional teacher-marched target on the smooth final hop~\cite{ptd}). Because the polynomial curvature basis functions are constructed to vanish strictly at both hop endpoints, the auxiliary curvature head serves solely as a training scaffold and is completely discarded after distillation.
\end{itemize}

\paragraph{Unified inference interface and hop-wise LoRA difference on MiniMax-H3.}
Despite their distinct distillation formulations, once PDD's sub-interval heads are fused per hop and PTD's auxiliary curvature head is removed, both methods share the exact same output interface and transition semantics at inference. Specifically, for any hop $b \in \{1, \dots, B\}$, both deployed generators map the backbone hidden features to a single latent-sized interval-average velocity $\bar{v}_{\theta,b}(z_b, c)$ via a linear projection head. While this interval-average velocity predictor supports both deterministic Euler transitions and stochastic re-noising sampling, under standard deterministic Euler deployment both models advance the hop-start state $z_b$ to the hop endpoint $\hat{z}_{b+1}$ via the identical finite-span update rule (Eq.~\eqref{eq:block_transition} of the main paper):
\begin{equation}
\hat{z}_{b+1} = F_{\theta,b}(z_b, c) = z_b + h_b \bar{v}_{\theta,b}(z_b, c).
\label{eq:ptd_update}
\end{equation}
In our post-training experiments on the 33B MiniMax-H3 backbone, the primary architectural distinction between the two setups lies in the parameter-sharing scheme of their LoRA~\cite{lora} adapters across denoising hops:
\begin{itemize}
\item \textbf{PDD on MiniMax-H3 (Hop-shared LoRA parameterization):}
Our PDD post-training employs a single set of shared backbone parameters and shared LoRA adapters across all $B$ denoising hops, conditioning on the hop-start timestamp $s_b$ while querying the fused hop-level velocity projection.
\item \textbf{PTD on MiniMax-H3 (Hop-specific LoRA adapters):}
Following the official PTD design for MiniMax-H3~\cite{ptd}, the PTD generator freezes the shared base weights and equips each denoising hop $b \in \{1, \dots, B\}$ with its own dedicated set of LoRA adapters on both the backbone and the velocity output projection head, allocating hop-specific low-rank capacity to adapt to the varying trajectory curvature across noise levels.
\end{itemize}

Because BlockAlign formulates preference supervision strictly at valid hop boundaries $s_b$ and supervises the span-normalized endpoint error $\mathcal{E}_\theta(x_1, x_0, b, c) = \|\bar{v}_{\theta,b}(z_b, c) - v^\star\|_2^2$ (Eq.~\eqref{eq:block_error} of the main paper), it seamlessly accommodates PTD on MiniMax-H3 without modification: during post-training, sampling hop index $b \sim p_{\text{obj}}(b)$ simply activates and updates the corresponding $b$-th hop-specific LoRA module while preserving native few-step inference consistency.

\subsection{Experimental Results}
\label{subsec:ptd_results}
\paragraph{Experimental setup.}
We evaluate BlockAlign on a 10-hop ($B=10$) PTD~\cite{ptd} generator distilled from the MiniMax-H3 backbone, where each denoising hop $b \in \{1, \dots, 10\}$ is equipped with a dedicated set of LoRA adapters. We follow the exact same online post-training strategy as used for MiniMax-H3 in Sec.~\ref{sec:experimental_setup} of the main paper: in each iteration, we generate $G=4$ online candidate rollouts per condition across all 10 PTD hops and employ \texttt{gemini-4-argon}~\cite{gemini4argon} under our hierarchical preference selection rule to identify the best and worst videos $\mathcal{P}_{\text{on}} = (x_1^{(g_+)}, x_1^{(g_-)
})$ for BlockAlign-DPO optimization (Eq.~\eqref{eq:ba_dpo}). During each update at sampled hop boundary $s_b$ ($b \sim p_{\text{obj}}(b)$), gradients are routed solely to the $b$-th hop-specific LoRA module while keeping the shared base weights and remaining hop adapters frozen. We adopt the same hardware and optimization configuration as in the main paper ($768\times 1280$ resolution, 5--15\,s duration at 24\,fps, Muon~\cite{liu2025muon} optimizer with learning rate $10^{-5}$, global batch size 12, and sequence-parallel degree 8 on 96 NVIDIA H200 GPUs) and train the model for 500 optimization steps; rather than attaching new adapters, we directly fine-tune the existing hop-specific LoRA adapters~\cite{lora} obtained from PTD distillation (across the attention, MLP, and average-velocity output projection layers) while keeping the base backbone frozen.

For a strictly controlled comparison, we evaluate two configurations on the same 150 multi-reference test conditions using identical inputs, inference settings, and random seeds: (1) the original unaligned 10-hop PTD checkpoint, and (2) PTD post-trained for 500 steps with our BlockAlign-DPO (Eq.~\eqref{eq:ba_dpo}).

\begin{figure*}[t]
    \centering
    \includegraphics[width=\textwidth]{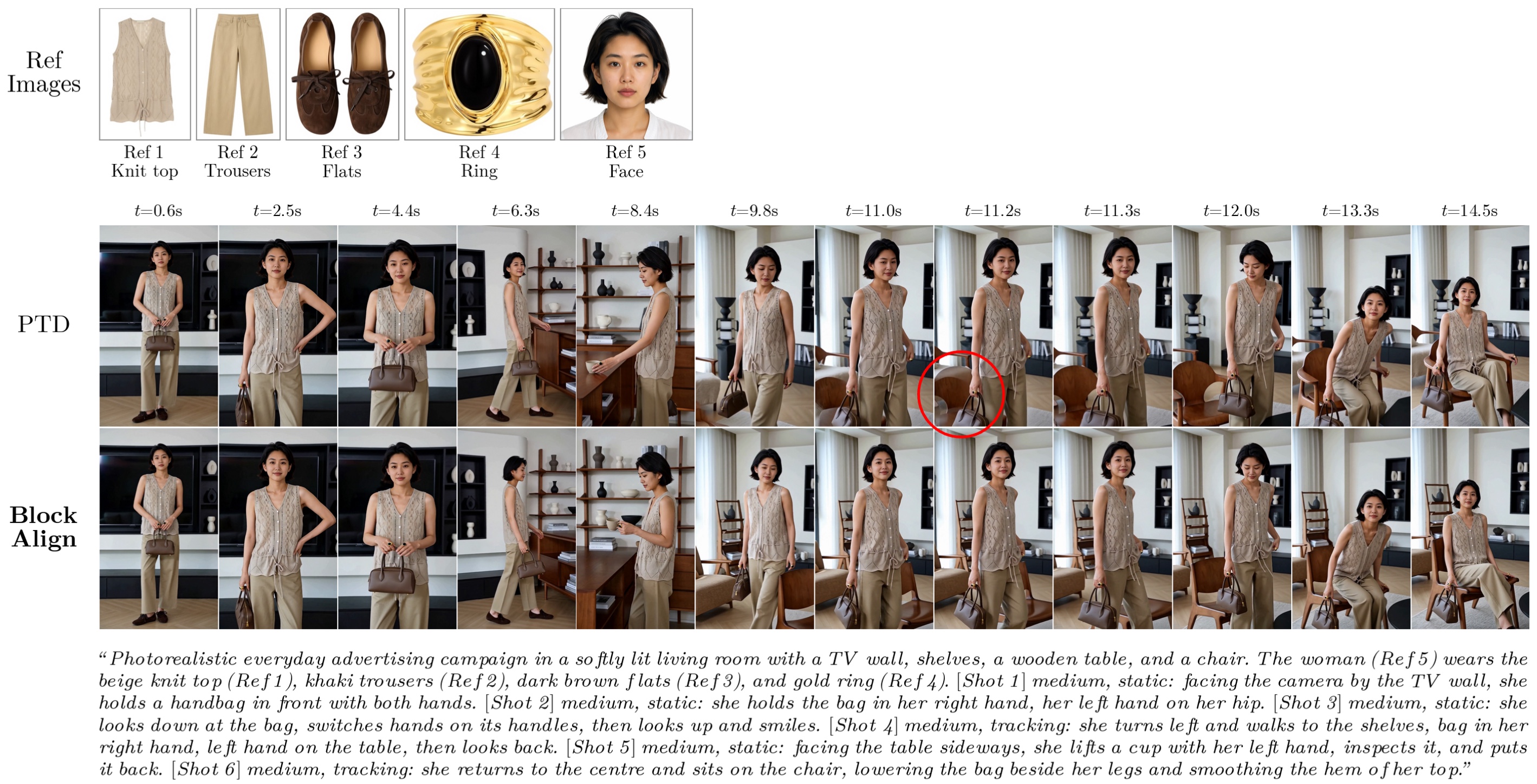}
    \caption{Qualitative comparison on MiniMax-H3 (PTD, Case 1). Both methods use identical inputs, random seeds, inference settings, and matched timestamps. The red circle highlights background blurring and spatial displacement of the wooden chair in the unaligned PTD baseline. BlockAlign preserves stable chair geometry and coherent scene structure alongside the referenced identity, knit top, trousers, flats, and ring. Best viewed zoomed in.}
    \label{fig:ptd_qual_case1}
\end{figure*}

\begin{figure*}[t]
    \centering
    \includegraphics[width=\textwidth]{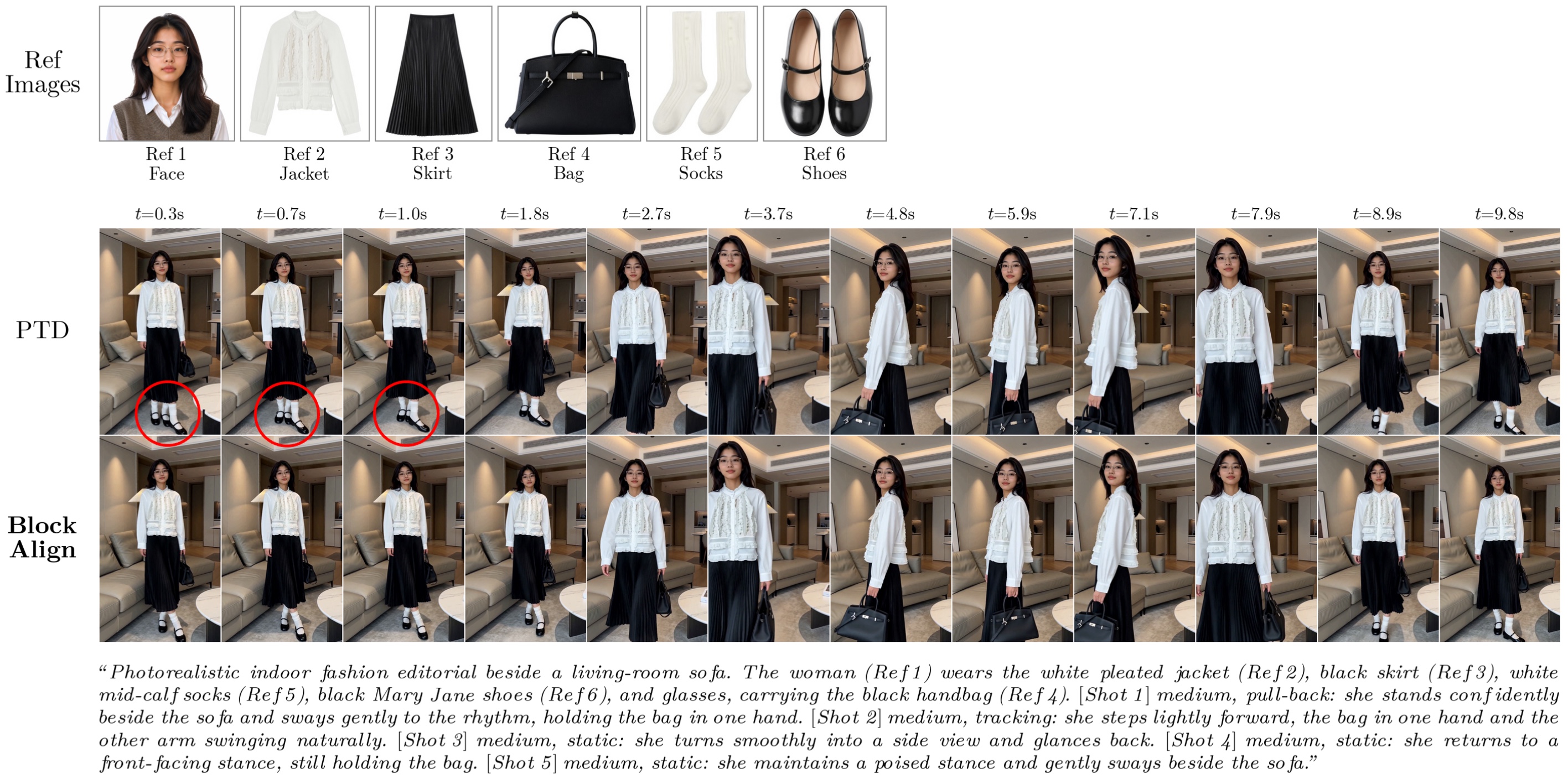}
    \caption{Qualitative comparison on MiniMax-H3 (PTD, Case 2). Red circles mark a severe anatomical deformity in the unaligned PTD baseline, which generates three feet/shoes. BlockAlign produces anatomically plausible lower-body structure and natural stepping motion while preserving the referenced face, jacket, skirt, handbag, socks, and Mary Jane shoes. Best viewed zoomed in.}
    \label{fig:ptd_qual_case2}
\end{figure*}

\begin{figure*}[t]
    \centering
    \includegraphics[width=\textwidth]{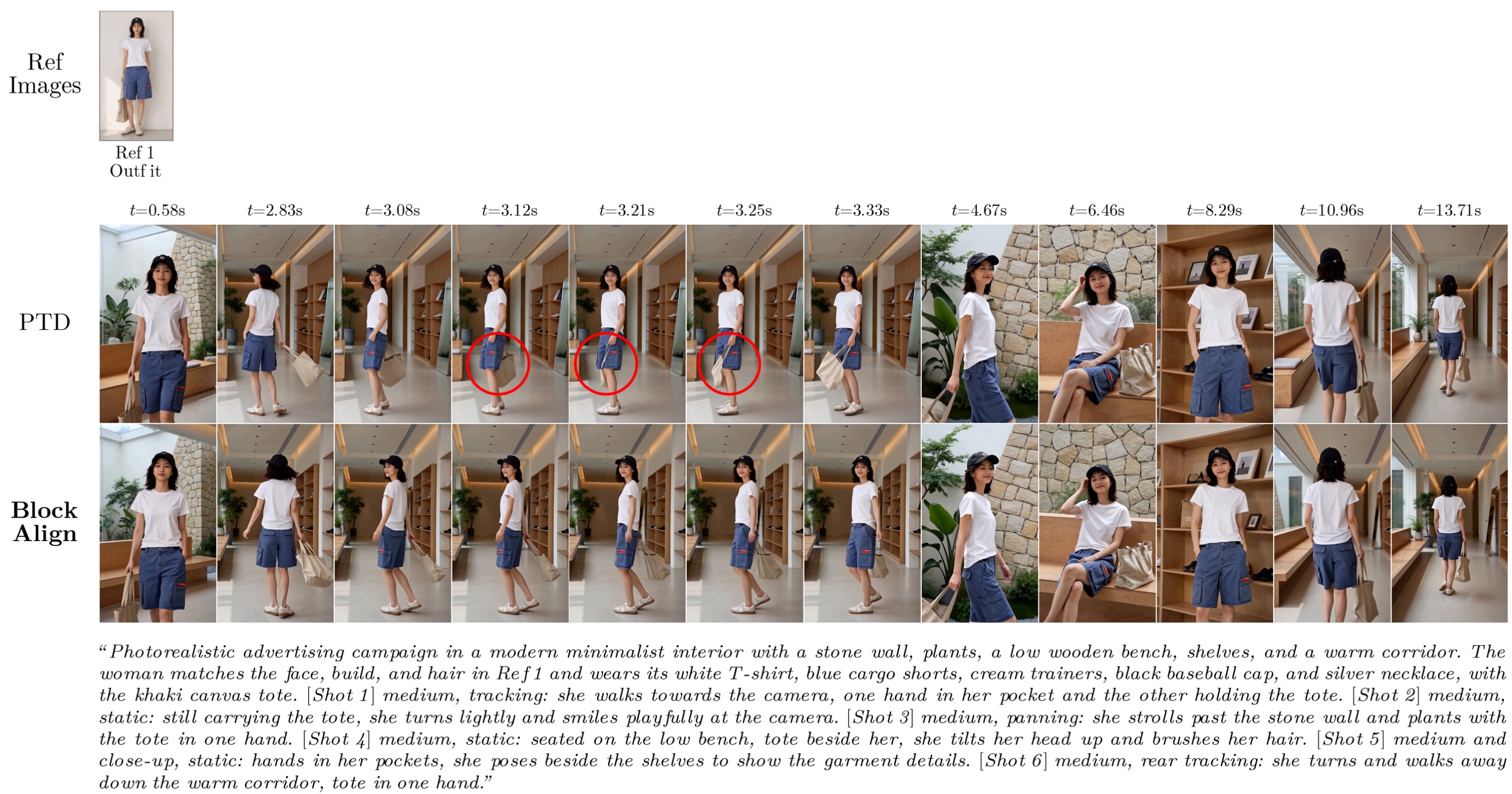} 
    \caption{Qualitative comparison on MiniMax-H3 (PTD, Case 3). Red circles highlight severe physical interpenetration between the khaki canvas tote and the blue cargo shorts in the unaligned PTD baseline. BlockAlign maintains physically valid object boundaries and natural hand--tote interactions while faithfully adhering to the referenced outfit and multi-shot prompt. Best viewed zoomed in.}
    \label{fig:ptd_qual_case3}
\end{figure*}

\paragraph{Qualitative results.}
Figures~\ref{fig:ptd_qual_case1}, \ref{fig:ptd_qual_case2}, and \ref{fig:ptd_qual_case3} compare the unaligned 10-hop PTD baseline and BlockAlign on MiniMax-H3 under identical reference images, text prompts, random seeds, and inference settings. Frames in each column are sampled at matched timestamps, and red circles highlight representative baseline failure modes:
\begin{itemize}
    \item \textbf{Background object blurring and spatial drift:} Figure~\ref{fig:ptd_qual_case1} evaluates a multi-shot living-room sequence conditioned on five reference images (beige knit top, khaki trousers, dark brown flats, gold ring, and face) involving walking, cup inspection, and sitting on a wooden chair. As the subject approaches the seating area ($t=11.0\text{--}12.0$\,s), the unaligned PTD baseline suffers from severe blurring and implausible spatial drift of the wooden chair behind her hand and handbag (highlighted at $t=11.2$\,s). In contrast, BlockAlign maintains sharp, temporally stable chair geometry and coherent spatial layout throughout the transition while faithfully preserving all five referenced items.
    \item \textbf{Extra-limb anatomical deformity:} Figure~\ref{fig:ptd_qual_case2} features an indoor fashion editorial conditioned on six references (face, white pleated jacket, black pleated skirt, black handbag, white mid-calf socks, and black Mary Jane shoes) across pull-back, tracking, and side-view shots. In the opening stance ($t=0.3\text{--}1.0$\,s), the unaligned PTD baseline exhibits a severe lower-body anatomical defect, synthesizing three feet and shoes simultaneously. BlockAlign completely eliminates this extra-limb hallucination, producing natural two-footed stances and smooth stepping motion while preserving the referenced identity, outfit, and accessories.
    \item \textbf{Severe hand--bag--garment interpenetration:} Figure~\ref{fig:ptd_qual_case3} shows a multi-shot sequence conditioned on a full-outfit reference image (white T-shirt, blue cargo shorts, cream trainers, black baseball cap, and khaki canvas tote) spanning walking, turning, and seated poses. During the turning motion ($t=3.12\text{--}3.25$\,s), the unaligned PTD model exhibits severe physical interpenetration, where the khaki canvas tote clips directly through the subject's blue cargo shorts and thigh. BlockAlign resolves this penetration artifact, maintaining physically plausible object boundaries and hand--tote interactions across all camera views.
\end{itemize}

\paragraph{Automated quantitative assessment.}
Following the evaluation protocol in Sec.~\ref{sec:quantitative_results} of the main paper, we report automated \texttt{gpt-6-astra}~\cite{openai2026gpt6astra} absolute scores (Table~\ref{tab:ptd_abs}) and bidirectional pairwise win rates (Table~\ref{tab:ptd_winrate}) across subject consistency (SC), prompt following (PF), visual quality (VQ), deformation (DF), and shot transitions (ST). As shown in Table~\ref{tab:ptd_abs}, BlockAlign elevates the overall mean absolute score of the 10-hop PTD generator from 3.82 to 3.84, with notable improvements in deformation reduction (DF: 3.70 to 3.74), prompt following (PF: 3.85 to 3.88), and shot transitions (ST: 3.85 to 3.88). In bidirectional head-to-head comparisons (Table~\ref{tab:ptd_winrate}), BlockAlign consistently surpasses the unaligned PTD baseline across all five dimensions, achieving a 52.8\% overall mean win rate (vs.\ 47.2\% for unaligned PTD) and pronounced gains in DF (55.0\%), VQ (54.5\%), and SC (53.8\%). Note that because both models share frozen base weights and near-ceiling absolute VQ scores ($3.98$), pairwise judgments on VQ exhibit higher presentation-order sensitivity ($38.7\%$ Agr.) than structural dimensions like ST ($73.3\%$) and PF ($56.0\%$), motivating our bidirectional averaging and human evaluation below. These results confirm that BlockAlign generalizes effectively to trajectory-distilled generators parameterized with hop-specific LoRA adapters.

\begin{table}[t]
    \centering
    \small
    \caption{
    Absolute assessment scores for the MiniMax-H3 (PTD) experiment (500 optimization steps; higher is better). Mean is the unweighted average over the five dimensions.
    }
    \label{tab:ptd_abs}
    \begin{tabular}{lrrrrrr}
        \toprule
        Method & SC & PF & VQ & DF & ST & Mean \\
        \midrule
        PTD & \textbf{3.73} & 3.85 & \textbf{3.98} & 3.70 & 3.85 & 3.82 \\
        BlockAlign & \textbf{3.73} & \textbf{3.88} & \textbf{3.98} & \textbf{3.74} & \textbf{3.88} & \textbf{3.84} \\
        \bottomrule
    \end{tabular}
\end{table}

\begin{table}[t]
    \centering
    \small
    \caption{
    Adjusted pairwise win rates (\%) for BlockAlign (500 steps) against PTD on MiniMax-H3; $50\%$ denotes parity. Agr.\ denotes bidirectional agreement between presentation orders.
    }
    \label{tab:ptd_winrate}
    \begin{tabular}{lrr}
        \toprule
        Dimension & Win Rate & Agr. \\
        \midrule
        Subject Consistency & 53.8 & 44.0 \\
        Prompt Following    & 50.5 & 56.0 \\
        Visual Quality      & 54.5 & 38.7 \\
        Deformation         & 55.0 & 42.7 \\
        Shot Transitions    & 50.2 & 73.3 \\
        \midrule
        Mean                & 52.8 & 50.9 \\
        \bottomrule
    \end{tabular}
\end{table}

\paragraph{Human evaluation.}
Following the blind pairwise human evaluation protocol in Sec.~\ref{sec:quantitative_results} of the main paper, five independent annotators evaluate the unaligned 10-step PTD baseline and BlockAlign across all $N=150$ multi-reference test conditions, jointly considering visual quality, multi-reference fidelity, prompt adherence, and structural plausibility to judge overall video usability. Human raters exhibit a clear and decisive preference for the BlockAlign post-trained PTD generator, which achieves $84/38/28$ wins/losses/ties (\textbf{56.0\%} wins vs.\ \textbf{25.3\%} losses, with \textbf{18.7\%} ties) over the unaligned PTD baseline—corresponding to a \textbf{65.3\%} win-plus-half-tie score and a \textbf{68.9\%} non-tie win rate (condition-level $95\%$ Wilson CI: $[60.2\%, 76.4\%]$, exact binomial $p < 10^{-4}$). Annotator feedback and post-evaluation inspection indicate that this pronounced margin—substantially wider than the $+0.02$ shift in frame-averaged VLM scores—stems from eliminating brief but critical anatomical deformities and physical interpenetration artifacts (e.g., extra limbs, malformed hands, and garment--object clipping) while preserving the strong multi-reference consistency and visual sharpness of the base PTD generator.

\section{Block-Aligned Adaptation of DiffusionART}
\label{app:block_art}

This section details how BlockAlign extends to rollout-based group feedback, instantiated with DiffusionART~\cite{diffusionart} (\textbf{BlockAlign-ART}). Unlike BlockAlign-DPO and BlockAlign-NFT, which construct preference objectives from linearly interpolated states (Eqs.~\eqref{eq:block_start_state}--\eqref{eq:block_end_state}), BlockAlign-ART operates directly on native PDD rollout states cached during candidate generation. Specifically, it constructs reward-guided block-endpoint targets from cached rollout states and fused block velocity predictions, and optimizes these targets via our span-normalized block error.

\subsection{Group-Level Reward Guidance}
\label{app:block_art_guidance}

For each condition $c$, we sample a group of $G$ trajectories across all $B$ native PDD blocks using the behavior PDD generator $\theta_{\text{old}}$. Let $x_1^{(g)}$ and $R_g$ denote the final generated video latent and scalar reward of rollout $g \in \{1, \dots, G\}$. For each block $b \in \{1, \dots, B\}$, we cache its block-start rollout state $z_b^{(g)}$ at boundary $s_b$ and its fused block-level velocity prediction:
\begin{equation}
    v_b^{(\text{old}, g)} = \bar{v}_{\theta_{\text{old}}, b}\!\left(z_b^{(g)}, c\right).
\end{equation}
All final latents, rewards, block-start states, and behavior predictions are detached and held fixed throughout the subsequent optimization round.

Let the group-level reward mean and endpoint latent mean be
\begin{equation}
    \bar{R}_c = \frac{1}{G}\sum_{g=1}^{G} R_g,
    \qquad
    \bar{x}_{1,c} = \frac{1}{G}\sum_{g=1}^{G} x_1^{(g)}.
\end{equation}
Following the group-feedback formulation of DiffusionART~\cite{diffusionart}, we estimate a reward-correlated endpoint direction via empirical covariance:
\begin{equation}
\begin{aligned}
    C_c &= \frac{1}{G}\sum_{g=1}^{G} (R_g - \bar{R}_c)\!\left(x_1^{(g)} - \bar{x}_{1,c}\right),\\
    G_c &= \frac{C_c}{\max\!\left(\operatorname{RMS}(C_c), \varepsilon\right)},\\
    \delta_c &= \eta G_c,
\end{aligned}
\label{eq:supp_art_guidance}
\end{equation}
where $\varepsilon > 0$ ensures numerical stability and $\eta > 0$ controls the guidance step size. Root-mean-square ($\operatorname{RMS}$) normalization renders the guidance scale invariant to variations in reward magnitude and latent dimensionality across rollout groups.

This covariance direction admits a natural interpretation via exponential reward tilting. Defining the tilted group probabilities and endpoint mean for temperature parameter $\alpha$ as
\begin{equation}
\begin{aligned}
    p_g(\alpha) &= \frac{\exp(\alpha R_g)}{\sum_{j=1}^{G}\exp(\alpha R_j)},\\
    \mu_c(\alpha) &= \sum_{g=1}^{G} p_g(\alpha) x_1^{(g)},
\end{aligned}
\end{equation}
and noting that $p_g(0) = 1/G$, the first-order derivative at $\alpha = 0$ satisfies
\begin{equation}
\begin{aligned}
    \left.\frac{\partial \mu_c(\alpha)}{\partial \alpha}\right|_{\alpha=0}
    &= \frac{1}{G}\sum_{g=1}^{G} (R_g - \bar{R}_c) x_1^{(g)}\\
    &= \frac{1}{G}\sum_{g=1}^{G} (R_g - \bar{R}_c)\!\left(x_1^{(g)} - \bar{x}_{1,c}\right)\\
    &= C_c.
\end{aligned}
\end{equation}
Thus, $C_c$ rigorously captures the first-order shift of the fixed-pool empirical endpoint mean under reward reweighting at $\alpha=0$, serving as a heuristic group-level endpoint guidance direction without requiring differentiable rewards.

\subsection{Block-Aligned Target Construction}
\label{app:block_art_target}

At cached block-start state $z_b^{(g)}$, the block endpoint predicted by the behavior policy $\theta_{\text{old}}$ over block span $h_b = e_b - s_b > 0$ is
\begin{equation}
    \hat{z}_{b+1}^{(\text{old}, g)} = z_b^{(g)} + h_b v_b^{(\text{old}, g)}.
\end{equation}
Expressing the group guidance $\delta_c$ in interval-average velocity space, we construct the detached target block endpoint as
\begin{equation}
    z_{b+1}^{\text{ART}, (g)} = z_b^{(g)} + h_b\!\left(v_b^{(\text{old}, g)} + \delta_c\right),
    \label{eq:supp_art_target}
\end{equation}
which shifts the local block transition by $z_{b+1}^{\text{ART}, (g)} - \hat{z}_{b+1}^{(\text{old}, g)} = h_b \delta_c$, preserving the scaling among block span $h_b$, interval-average velocity, and local endpoint displacement. Following DiffusionART~\cite{diffusionart}, broadcasting the endpoint-derived direction $\delta_c$ across blocks is an empirical approximation that avoids per-block posterior rollouts: because downstream block transitions are nonlinear, injecting $\delta_c$ at each block induces a first-order terminal shift $\Delta x_1 \approx \sum_{b=1}^{B} h_b J_{\text{suffix},b}\,\delta_c$ (where $J_{\text{suffix},b}$ denotes the Jacobian of the remaining blocks $b+1\dots B$) rather than $\delta_c$ identically. To prevent this heuristic cross-block credit assignment from destabilizing the trajectory, we regulate policy updates via the restoration pass and KL trust-region masking in Sec.~\ref{app:block_art_dynamics}.

Given the current PDD block transition $\mathcal{F}_{\theta, b}(z_b^{(g)}, c) = z_b^{(g)} + h_b \bar{v}_{\theta, b}(z_b^{(g)}, c)$, the BlockAlign-ART objective supervises the span-normalized endpoint residual:
\begin{small}
\begin{equation}
\begin{aligned}
    \mathcal{L}_{\text{BA-ART}}(\theta)
    &= \mathbb{E}_{c, g, b}\!\left[
    \frac{w_b^{(g)}}{d}
    \left\|
    \frac{\mathcal{F}_{\theta, b}(z_b^{(g)}, c) - z_{b+1}^{\text{ART}, (g)}}{h_b}
    \right\|_2^2
    \right]\\
    &= \mathbb{E}_{c, g, b}\!\left[
    \frac{w_b^{(g)}}{d}
    \left\|
    \bar{v}_{\theta, b}(z_b^{(g)}, c) - v_b^{(\text{old}, g)} - \delta_c
    \right\|_2^2
    \right],
\end{aligned}
\label{eq:supp_art_objective}
\end{equation}
\end{small}
where $d$ is the latent dimensionality and $w_b^{(g)} \ge 0$ is a detached weighting factor. Normalizing the endpoint discrepancy by $h_b$ resolves step-size scaling across nonuniform PDD blocks (though not exact multi-block credit assignment), reducing the objective to velocity-space matching on the fused operator $\bar{v}_{\theta,b}$.

\subsection{Guidance, Restoration, and Trust Region}
\label{app:block_art_dynamics}

\paragraph{Guidance and restoration dynamics.}
Defining the velocity prediction drift from the behavior policy as $\Delta_b^{(g)} = \bar{v}_{\theta, b}(z_b^{(g)}, c) - v_b^{(\text{old}, g)}$, the per-block loss is $\ell_b^{(g)} = \frac{w_b^{(g)}}{d}\|\Delta_b^{(g)} - \delta_c\|_2^2$, with negative output-space gradient:
\begin{equation}
    -\nabla_{\bar{v}_{\theta, b}}\ell_b^{(g)}
    = \frac{2w_b^{(g)}}{d}\!\left(\delta_c - \Delta_b^{(g)}\right).
    \label{eq:supp_art_restoration}
\end{equation}
Equation~\eqref{eq:supp_art_restoration} decomposes each update into a reward-guidance force $\delta_c$ and an intrinsic restoring force $-\Delta_b^{(g)}$. At the start of an optimization round ($\theta = \theta_{\text{old}}$), $\Delta_b^{(g)} = 0$, so updates follow the group reward guidance $\delta_c$. As sequential minibatch updates modify $\theta$ within the round, nonzero drift $\Delta_b^{(g)}$ emerges on subsequent cached states, and $-\Delta_b^{(g)}$ automatically counteracts excessive deviation from the behavior policy transitions.

\paragraph{Optional trust-region mask.}
As an additional safeguard against excessive policy drift, we optionally mask updates that would further amplify large deviations from $\theta_{\text{old}}$. Modeling the block transitions (omitting rollout index $g$ for brevity)  via auxiliary isotropic Gaussians with fixed variance $\tau_b^2 > 0$:
\begin{equation}
\begin{aligned}
    q_{\theta, b} &= \mathcal{N}\!\left(\mathcal{F}_{\theta, b}(z_b, c),\, \tau_b^2 I\right),\\
    q_{\text{old}, b} &= \mathcal{N}\!\left(z_b + h_b v_b^{(\text{old})},\, \tau_b^2 I\right),
\end{aligned}
\end{equation}
their per-coordinate KL divergence yields the transition-space drift metric:
\begin{equation}
\begin{aligned}
    D_b &= \frac{1}{d} D_{\text{KL}}\!\left(q_{\theta, b} \,\|\, q_{\text{old}, b}\right)\\
    &= \frac{\left\|\mathcal{F}_{\theta, b}(z_b, c) - (z_b + h_b v_b^{(\text{old})})\right\|_2^2}{2d\tau_b^2}
    &= \frac{h_b^2 \|\Delta_b\|_2^2}{2d\tau_b^2}.
\end{aligned}
\label{eq:supp_art_drift}
\end{equation}
For the unmasked descent direction $u_b = \delta_c - \Delta_b$, the directional derivative of the squared drift satisfies $\left.\frac{\partial}{\partial \rho}\|\Delta_b + \rho u_b\|_2^2\right|_{\rho=0} = 2\langle \Delta_b, u_b\rangle$. Thus, $\langle \Delta_b, \delta_c - \Delta_b\rangle > 0$ indicates an outward update that locally increases policy drift. We therefore define the detached trust-region weight as
\begin{equation}
    w_b = \omega_b\!\left[
    1 - \mathbf{1}_{D_b \ge \kappa}\,\mathbf{1}_{\langle \Delta_b, \delta_c - \Delta_b\rangle > 0}
    \right],
    \label{eq:supp_art_mask}
\end{equation}
where $\mathbf{1}_{(\cdot)}$ is the indicator function, $\kappa > 0$ is the drift threshold, and $\omega_b$ denotes the base block weight. This directional mask suppresses updates only when the current drift exceeds $\kappa$ \emph{and} the update points outward, while keeping inward corrective updates active. Note that $D_b$ serves purely as a transition-space drift proxy during training without altering the native PDD Euler sampler.

\begin{figure*}[t]
    \centering
    \includegraphics[
        width=\textwidth,
        height=0.48\textheight,
        keepaspectratio
    ]{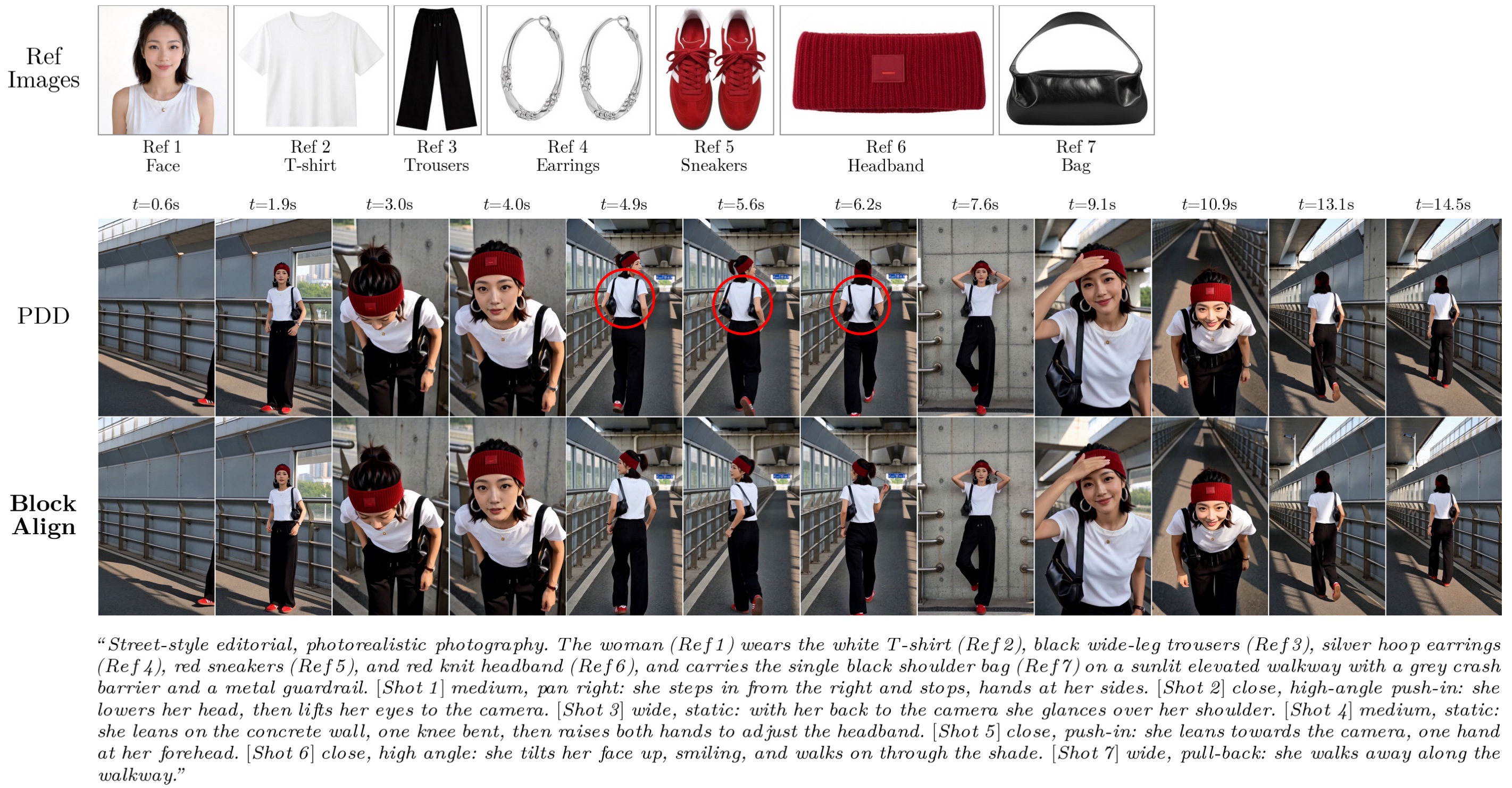}
    \vspace{4pt}
    \includegraphics[
        width=\textwidth,
        height=0.48\textheight,
        keepaspectratio
    ]{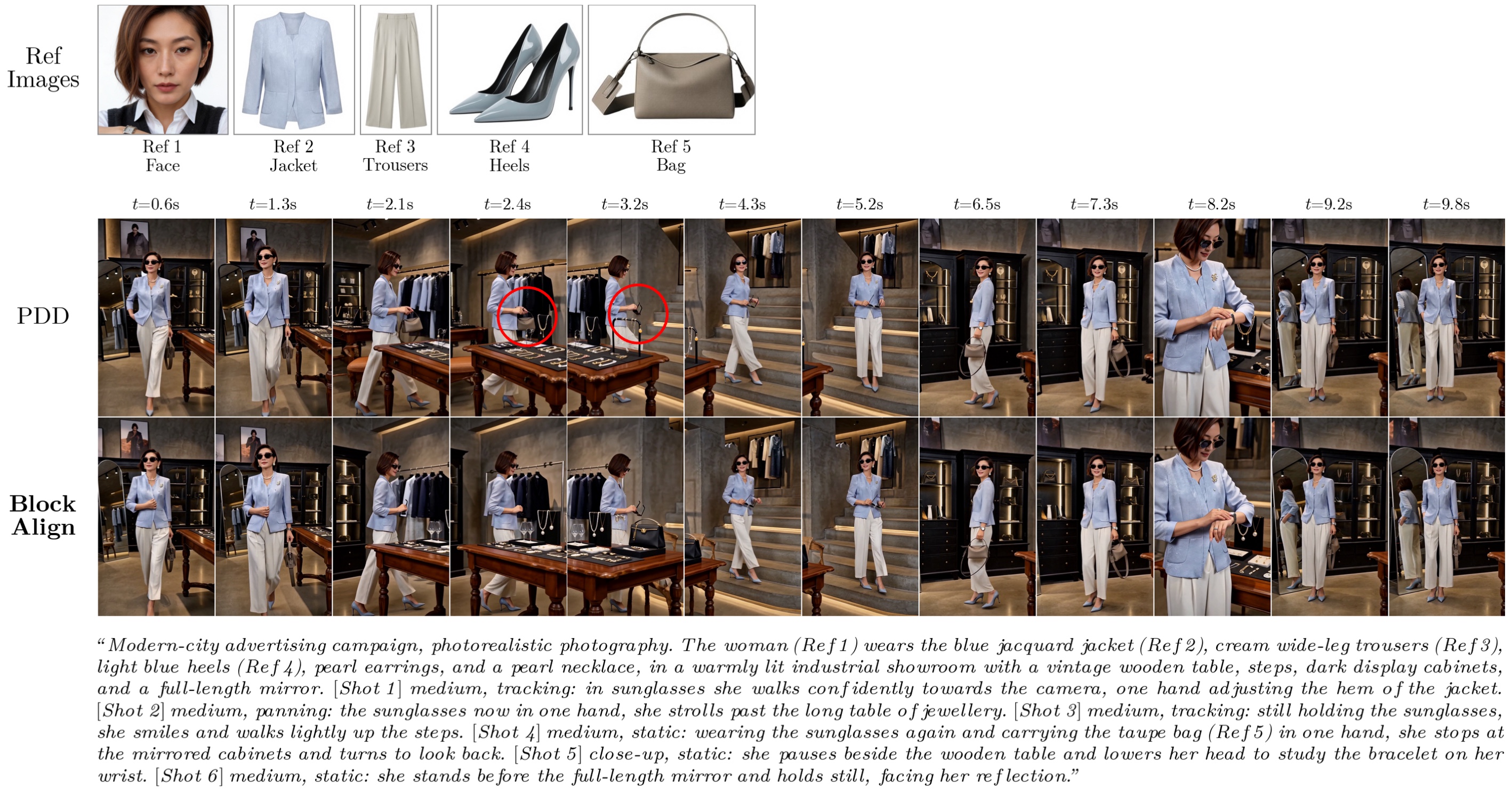}
    \caption{
    Qualitative evaluation of BlockAlign-ART on MiniMax-H3 after 400 optimization steps under identical conditioning inputs, random seeds, inference settings, and matched timestamps. \textbf{Top:} The prompt specifies a single black shoulder bag; PDD produces an additional bag-like structure around the subject's back and arm (red circles), whereas BlockAlign-ART consistently maintains a single referenced bag across viewpoints. \textbf{Bottom:} When the subject removes and holds her sunglasses, PDD generates a malformed handheld object conflated with the referenced bag, whereas BlockAlign-ART produces recognizable sunglasses and a plausible hand--object interaction. Best viewed zoomed in.
    }
    \label{fig:block_art_qualitative}
\end{figure*}

\subsection{Experimental Results}
\label{app:block_art_experiments}

\paragraph{Experimental setup.}
We conduct a proof-of-concept evaluation of BlockAlign-ART on the 10-step ($B=10$) PDD generator of the MiniMax-H3 backbone using the same hardware and LoRA configuration as in the main paper ($768\times 1280$ resolution, 5--15\,s duration at 24\,fps, LoRA~\cite{lora} on attention and MLP layers, Muon~\cite{liu2025muon} optimizer with learning rate $10^{-5}$, global batch size 12, and sequence-parallel degree 8 on 96 NVIDIA H200 GPUs).

For each condition, we generate a group of $G=8$ online rollout trajectories and assign each candidate a hierarchical integer reward $R_g \in \{0, 1, 2, 3, 4\}$: a video receives $0$ if it contains visible deformations or malformed content; a deformation-free video receives $1$ if it is inconsistent with the reference images, $2$ if it preserves the references but fails prompt following, $3$ if it satisfies structural validity, reference consistency, and prompt following, and $4$ if it additionally exhibits strong aesthetic quality. This hierarchical rubric mirrors our primary structural gating principle in Sec.~\ref{sec:supervision_modes} and drives the group-level guidance direction in Eq.~\eqref{eq:supp_art_guidance}. Because $G=8$ online rollouts and reward evaluations incur higher computational cost than paired DPO/NFT training, we train BlockAlign-ART for 400 optimization steps as a proof-of-concept evaluation rather than a budget-matched comparison with the 1,500-step main experiments. Note that the $0$--$4$ rollout rewards are used strictly for ART training and are distinct from the $1$--$5$ absolute evaluation scores below.

\paragraph{Qualitative results.}
Figure~\ref{fig:block_art_qualitative} compares the base PDD model and BlockAlign-ART after 400 steps under identical inputs, random seeds, inference settings, and matched timestamps. In the top case (specifying a single black shoulder bag), PDD hallucinates a duplicate bag-like structure around the subject's back and arm, whereas BlockAlign-ART eliminates this duplication and maintains a single bag across front, side, and rear views while preserving all referenced attributes. In the bottom case (removing and holding sunglasses), PDD generates a malformed handheld object conflated with the referenced bag, whereas BlockAlign-ART synthesizes recognizable sunglasses with a plausible hand--object interaction.

\begin{table}[t]
    \centering
    \small
    \setlength{\tabcolsep}{5pt}
    \caption{
    Absolute assessment scores for the rollout-based BlockAlign-ART experiment on MiniMax-H3 (400 optimization steps; higher is better). Mean is the unweighted average over the five dimensions.
    }
    \label{tab:art_auto_absolute}
    \begin{tabular}{lrrrrrr}
        \toprule
        Method & SC & PF & VQ & DF & ST & Mean \\
        \midrule
        PDD & 3.41 & 3.62 & 3.96 & 3.60 & 3.65 & 3.65 \\
        BlockAlign-ART & \textbf{3.44} & \textbf{3.69} & \textbf{3.98} & \textbf{3.66} & \textbf{3.68} & \textbf{3.69} \\
        \bottomrule
    \end{tabular}
\end{table}

\begin{table}[t]
    \centering
    \small
    \caption{
    Adjusted pairwise win rates (\%) for BlockAlign-ART (400 steps) against PDD on MiniMax-H3; $50\%$ denotes parity. Agr.\ denotes bidirectional agreement between presentation orders.
    }
    \label{tab:art_auto_pairwise}
    \begin{tabular}{lrr}
        \toprule
        Dimension & Win Rate & Agr. \\
        \midrule
        Subject Consistency & 53.7 & 46.7 \\
        Prompt Following    & 51.3 & 60.0 \\
        Visual Quality      & 52.7 & 43.3 \\
        Deformation         & 51.0 & 41.3 \\
        Shot Transitions    & 51.5 & 74.7 \\
        \midrule
        Mean                & 52.0 & 53.2 \\
        \bottomrule
    \end{tabular}
\end{table}

\paragraph{Quantitative results.}
Following the automated evaluation protocol in Sec.~\ref{sec:quantitative_results}, we use \texttt{gpt-6-astra}~\cite{openai2026gpt6astra} on the 150 test conditions to assess subject consistency (SC), prompt following (PF), visual quality (VQ), deformation (DF), and shot transitions (ST) via 1--5 absolute scoring (Table~\ref{tab:art_auto_absolute}) and bidirectional pairwise comparisons (Table~\ref{tab:art_auto_pairwise}).

As shown in Table~\ref{tab:art_auto_absolute}, BlockAlign-ART improves the mean absolute score from $3.65$ to $3.69$ within 400 optimization steps, achieving consistent gains across all five dimensions led by prompt following ($3.62 \to 3.69$) and deformation reduction ($3.60 \to 3.66$). In Table~\ref{tab:art_auto_pairwise}, BlockAlign-ART achieves an adjusted mean win rate of $52.0\%$ against PDD and exceeds the $50\%$ parity threshold across all five dimensions, led by subject consistency ($53.7\%$) and visual quality ($52.7\%$). While modest due to the 400-step budget, these consistent gains verify that group-relative rollout feedback can be effectively integrated via native PDD block transitions.

\begin{figure*}[t]
\centering
\includegraphics[width=\textwidth]{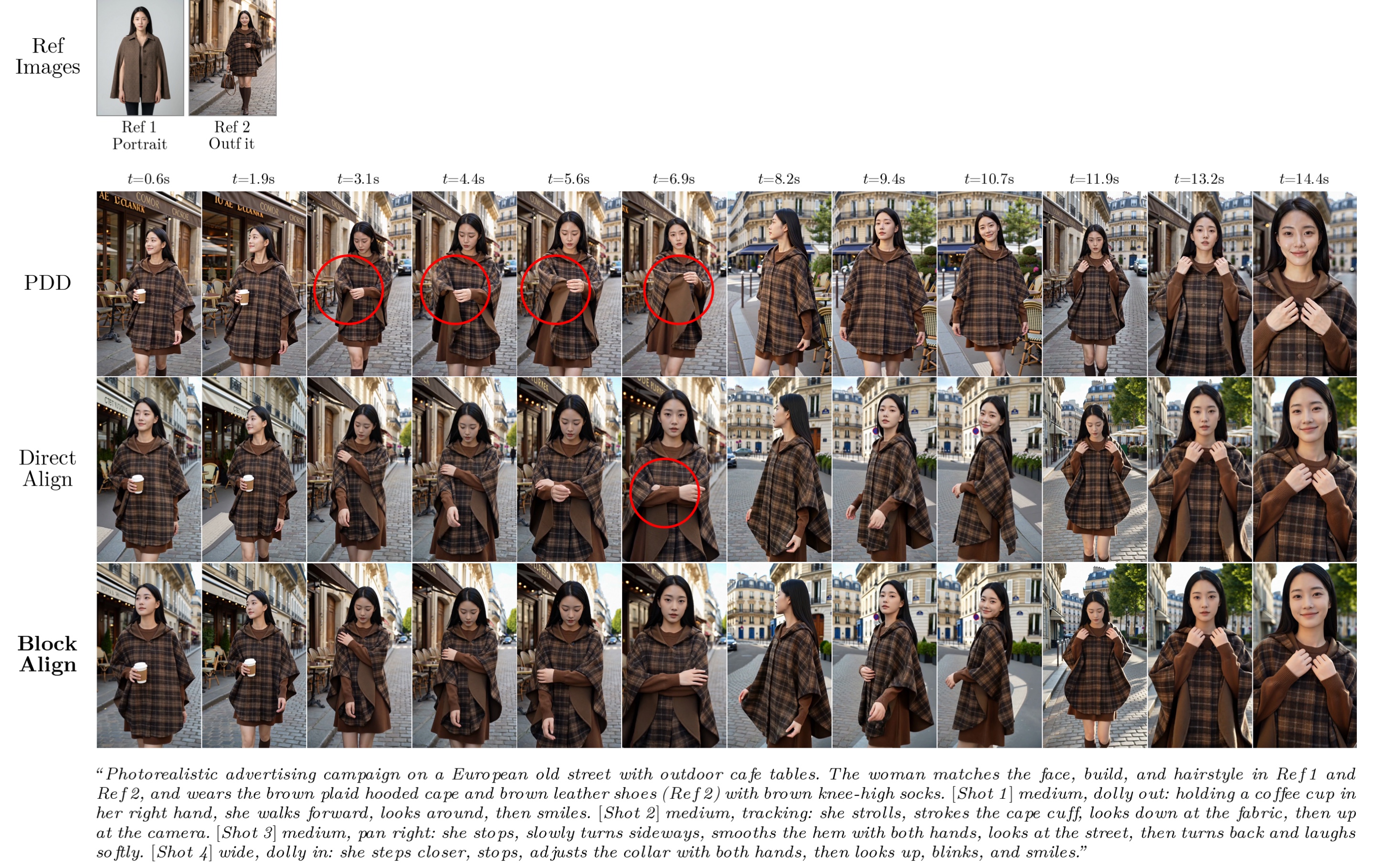}
\caption{Additional RefVid-V4 comparison. Red circles highlight implausible hand--arm configurations and garment boundaries in the baselines. BlockAlign better preserves the referenced identity and outfit while producing clearer and less patchy local textures. Best viewed zoomed in.}
\label{fig:qualitative_hh_case2}
\end{figure*}

\begin{figure*}[t]
\centering
\includegraphics[width=\textwidth]{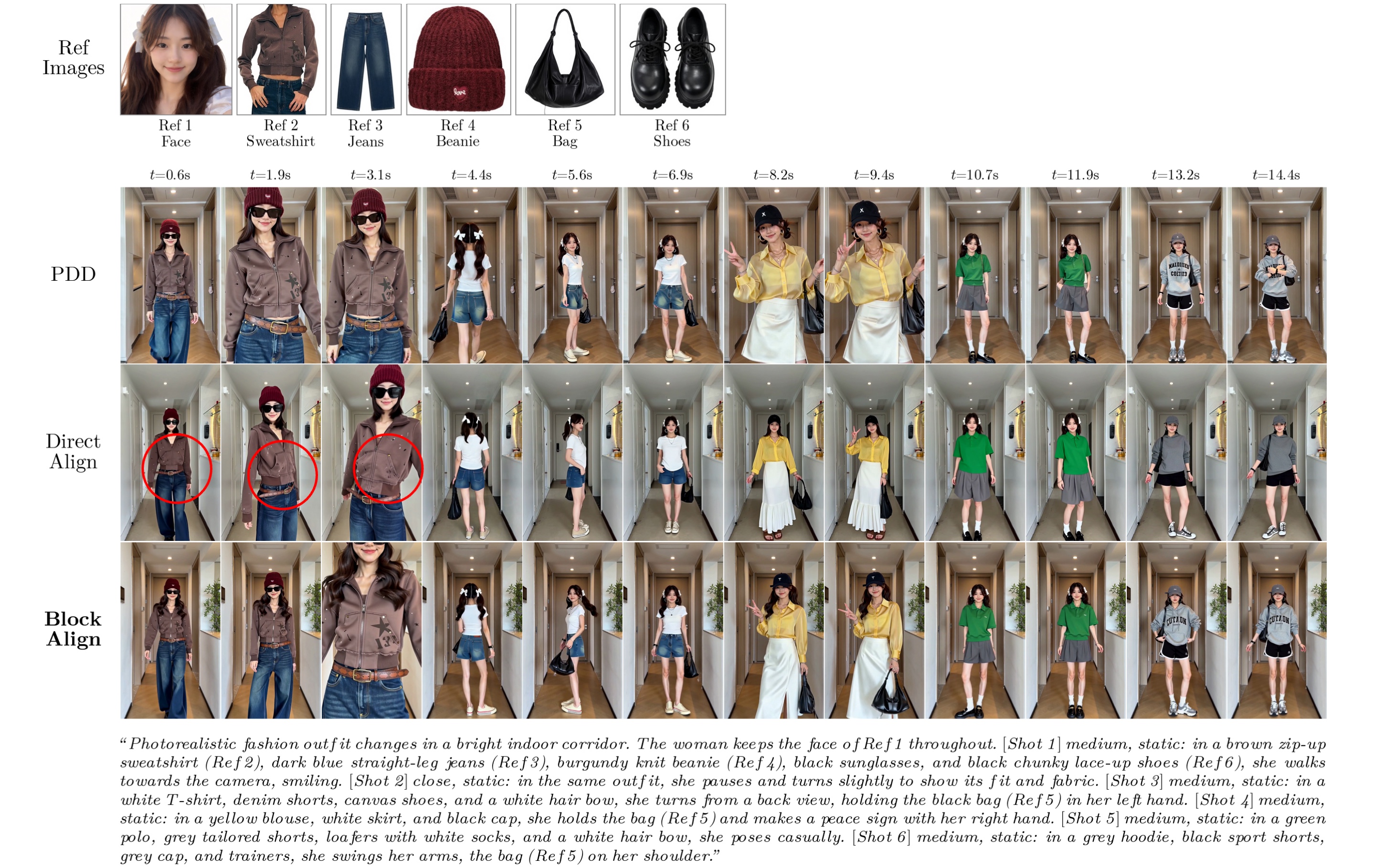}
\caption{Garment-reference consistency on RefVid-V4. Red circles mark deviations from the referenced sweatshirt in DirectAlign. BlockAlign better preserves its shape, closure, printed details, color, and associated accessories, with clearer and less patchy fabric textures. Best viewed zoomed in.}
\label{fig:qualitative_hh_case3}
\end{figure*}

\begin{figure*}[t]
\centering
\includegraphics[width=\textwidth]{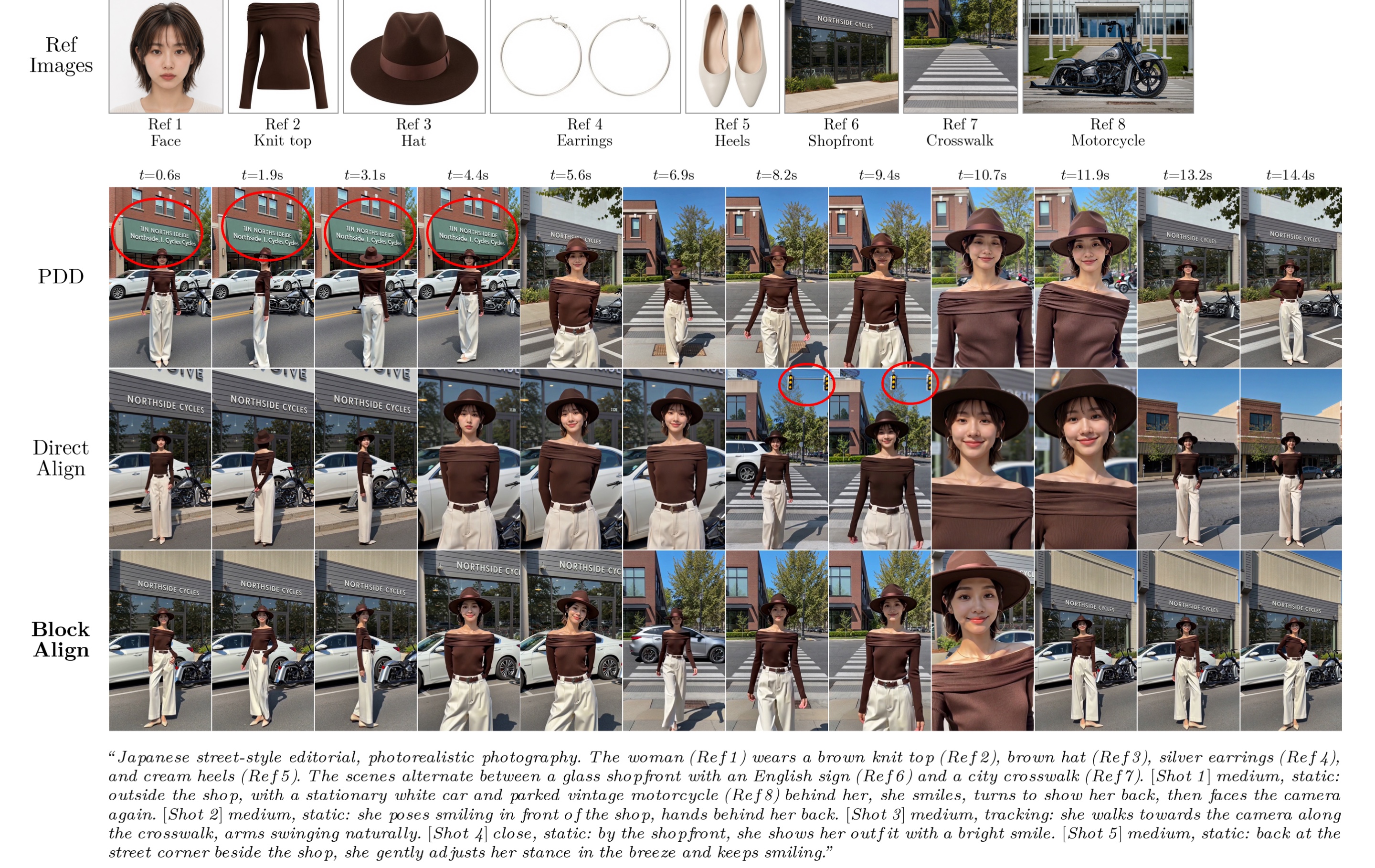}
\caption{Scene-reference consistency on RefVid-V4. Red circles highlight corrupted storefront text in PDD and scene-layout deviations in DirectAlign. BlockAlign produces a more coherent conditioned scene and cleaner local details. Best viewed zoomed in.}
\label{fig:qualitative_hh_case4}
\end{figure*}

\begin{figure*}[t]
\centering
\includegraphics[width=\textwidth]{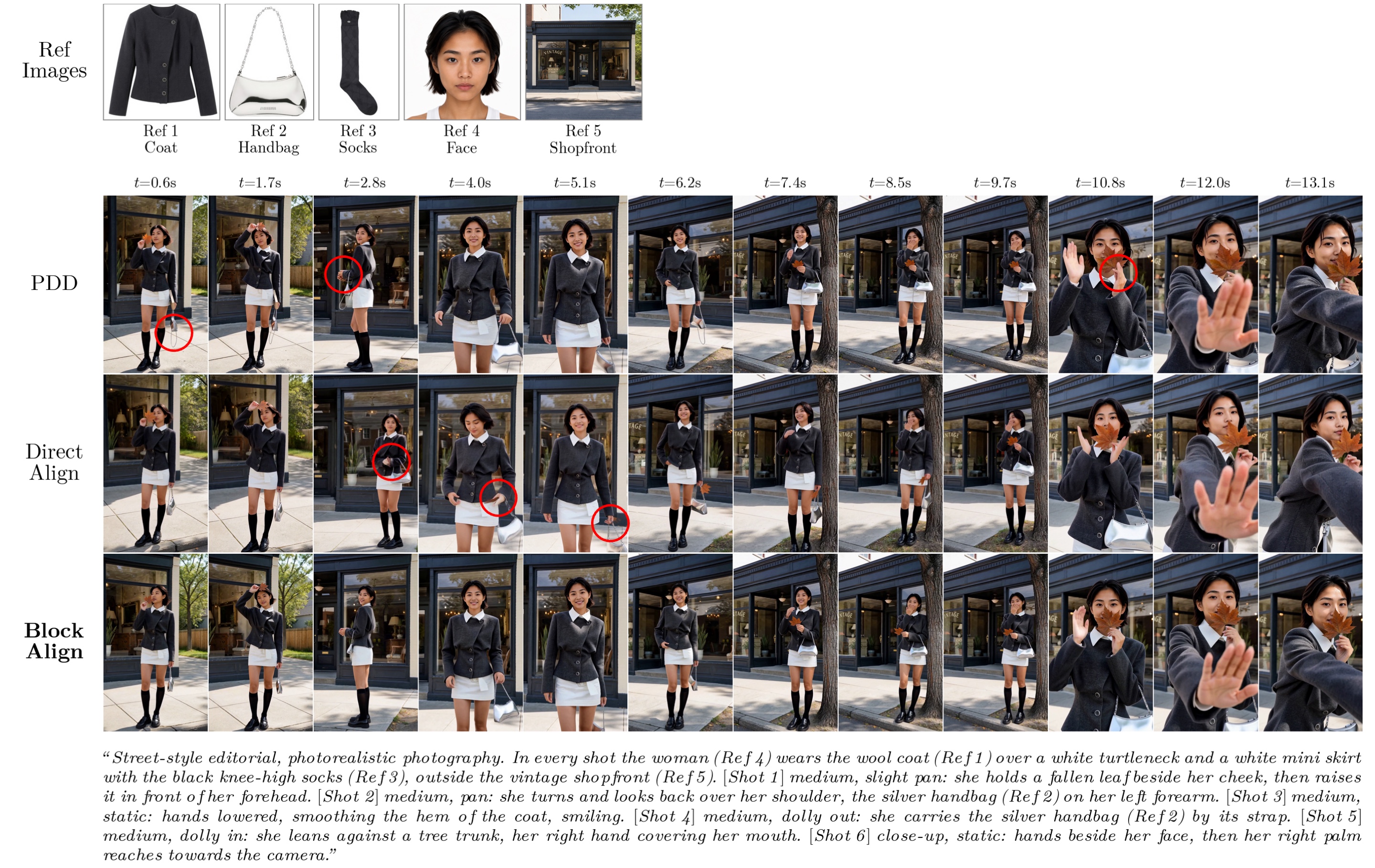}
\caption{Additional MiniMax-H3 comparison. Red circles highlight malformed hands and implausible interactions with the leaf or handbag. BlockAlign produces more coherent anatomy and interactions while preserving the referenced identity, garments, handbag, and shopfront. Best viewed zoomed in.}
\label{fig:qualitative_h3_case1}
\end{figure*}

\begin{figure*}[t]
\centering
\includegraphics[width=\textwidth]{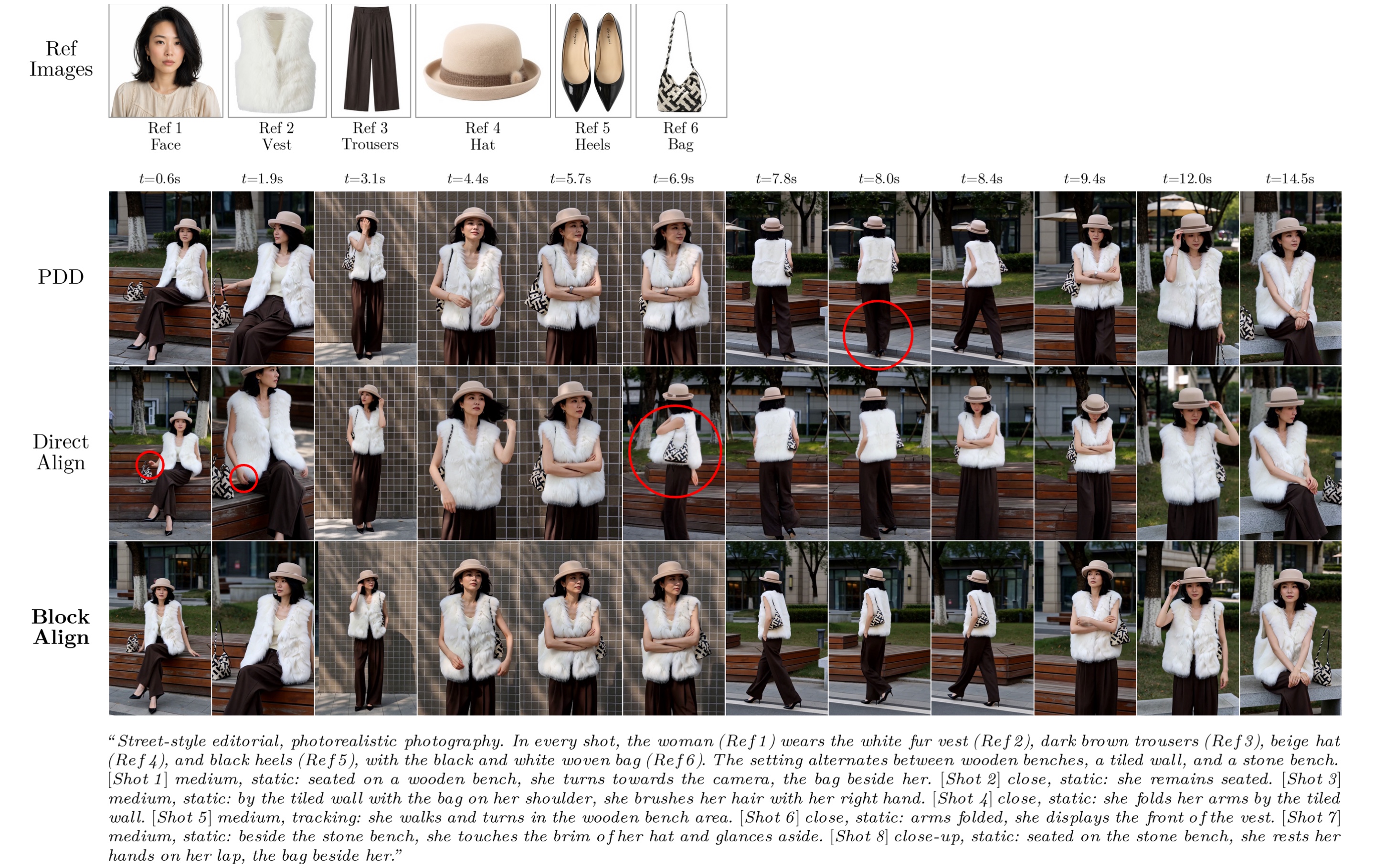}
\caption{Additional MiniMax-H3 comparison with seated and walking poses. Red circles indicate malformed hands, implausible arm--torso configurations, and inconsistent lower-body structure. BlockAlign produces more coherent anatomy while preserving the referenced garments and accessories. Best viewed zoomed in.}
\label{fig:qualitative_h3_case3}
\end{figure*}

\begin{figure*}[t]
\centering
\includegraphics[width=\textwidth]{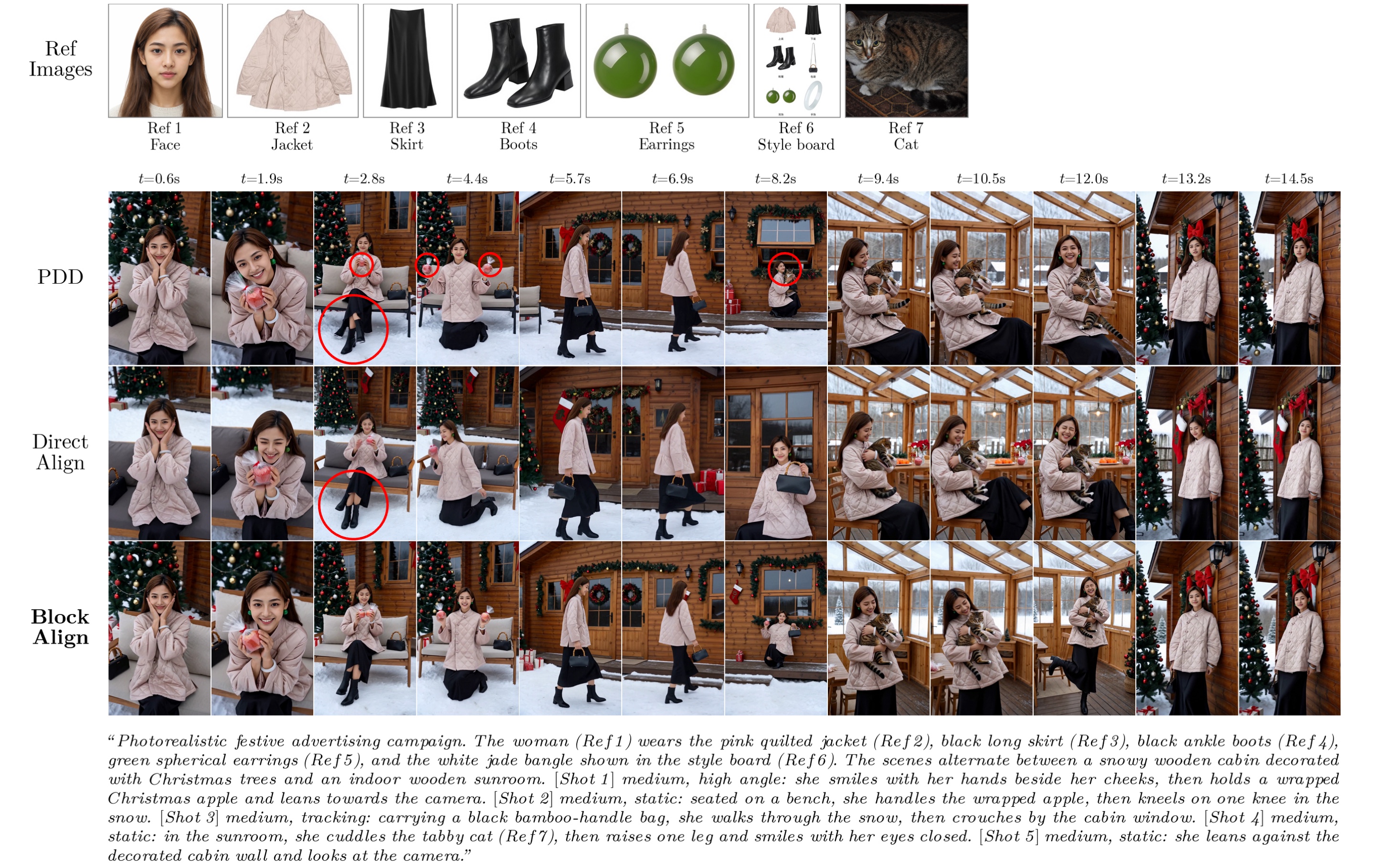}
\caption{Additional MiniMax-H3 comparison involving a wrapped object, kneeling motion, and a referenced cat. Red circles mark baseline errors in hand--object, lower-body, and subject--animal interactions. BlockAlign produces more plausible interactions while preserving the conditioned identity, clothing, accessories, and cat. Best viewed zoomed in.}
\label{fig:qualitative_h3_case4}
\end{figure*}

\section{Additional Qualitative Results}
\label{sec:additional_qualitative}

We provide additional qualitative comparisons on RefVid-V4 and MiniMax-H3. Across all cases, PDD, DirectAlign, and BlockAlign are evaluated under identical reference images, text prompts, random seeds, and inference settings, with frames in each column sampled at matched timestamps. Red circles highlight representative baseline errors in PDD or DirectAlign rather than exhaustive annotations.

\subsection{RefVid-V4}

Across Figures~\ref{fig:qualitative_hh_case2}--\ref{fig:qualitative_hh_case4}, BlockAlign consistently improves structural plausibility, multi-reference garment fidelity, and conditioned scene realization. Benefiting from our semi-online texture and composition supervision, its outputs also exhibit sharper local details, richer color saturation, and noticeably fewer patch-like texture artifacts across clothing, skin, and backgrounds.

\paragraph{Garment and body structure.}
In Figure~\ref{fig:qualitative_hh_case2}, both PDD and DirectAlign produce locally distorted hands, implausible forearm configurations, and blurred garment boundaries beneath the loose plaid cape. BlockAlign maintains coherent upper-body anatomy and hand geometry while faithfully preserving the referenced identity, plaid pattern, cape silhouette, and complete outfit.

\paragraph{Consistency with garment references.}
Figure~\ref{fig:qualitative_hh_case3} evaluates fine-grained fidelity to multiple garment and accessory references. DirectAlign deviates from the referenced sweatshirt across its silhouette, zipper closure, printed graphics, hem, and waist. In contrast, BlockAlign accurately reproduces the referenced sweatshirt, jeans, beanie, bag, and shoes across viewpoint and scale transitions while maintaining facial identity and clean fabric textures.

\paragraph{Scene-reference consistency.}
In Figure~\ref{fig:qualitative_hh_case4}, the base PDD model generates corrupted storefront signage text, while DirectAlign introduces noticeable scene-layout distortions around the crosswalk. BlockAlign coherently realizes the referenced storefront, legible signage, crosswalk geometry, motorcycle, clothing, and accessories.

\subsection{MiniMax-H3}

Across Figures~\ref{fig:qualitative_h3_case1}--\ref{fig:qualitative_h3_case4}, we evaluate MiniMax-H3 on challenging scenarios involving fine-grained hand--object manipulation, seated-to-walking transitions, lower-body kinematics, and subject--animal interactions.

\paragraph{Hand--object and accessory interactions.}
In Figure~\ref{fig:qualitative_h3_case1}, both PDD and DirectAlign exhibit malformed hand geometry and implausible physical interactions when the subject holds a leaf and carries a referenced handbag. BlockAlign synthesizes anatomically coherent hands and natural object manipulation while preserving the referenced identity, garments, handbag, and shopfront background.

\paragraph{Seated and walking pose transitions.}
Figure~\ref{fig:qualitative_h3_case3} examines multi-shot sequences covering seated and walking poses. The baselines suffer from distorted fingers, implausible arm--torso articulations, and inconsistent lower-body structure during motion. BlockAlign maintains temporally stable limb geometry and natural posture transitions while preserving all referenced garments and accessories.

\paragraph{Complex motion and subject--animal interaction.}
Figure~\ref{fig:qualitative_h3_case4} involves handling a wrapped object, kneeling motion, and interacting with a referenced cat. Both PDD and DirectAlign produce structural defects in hand--object contact, lower-body kneeling anatomy, and cat morphology. BlockAlign generates physically plausible interactions and coherent subject--animal geometry while preserving the conditioned identity, outfit, and pet appearance.

\paragraph{Summary.}
Overall, while DirectAlign occasionally alters individual frames, its unfused sub-head preference updates fail to resolve---and frequently exacerbate---PDD's structural, interaction, and reference-consistency errors. In contrast, BlockAlign delivers consistent improvements across consecutive timestamps: on RefVid-V4, it simultaneously enhances structural fidelity, clarity, color saturation, and texture coherence; on MiniMax-H3, it reliably eliminates anatomical and interaction defects.

\section{Public Video Generation Benchmarks}
\label{sec:public_benchmarks}

To comprehensively assess zero-shot cross-domain generalization beyond our primary e-commerce benchmark, we evaluate the 10-step PDD baseline, DirectAlign, and BlockAlign (on the MiniMax-H3 backbone) across two public video generation benchmarks: \textbf{OpenS2V-Eval}~\cite{yuan2025opens2v} ($360$ test cases across four domains) and \textbf{OmniVBench}~\cite{omnivbench2026} ($813$ test cases across seven task families). Both benchmarks consist predominantly of landscape-oriented videos, whereas our training data is dominated by portrait e-commerce clips; these evaluations thus also examine generalization under a combined domain and aspect-ratio
shift. Because neither benchmark enforces a mandatory fixed video duration in its prompt specifications, and our framework targets multi-shot, longer-horizon video generation, we standardize all three models to generate \textbf{8-second videos} (at $24\,\text{fps}$, $768p$ resolution) under identical conditioning inputs, 10-step PDD inference schedules, and random seeds. On OmniVBench, we additionally evaluate BlockAlign under a \textbf{5-second setting} (at $24\,\text{fps}$, $768p$ resolution) with identical prompts, references, and random seeds to match the 5-second horizon of the official leaderboard.

\begin{table*}[t]
\centering
\caption{Quantitative evaluation on the public OpenS2V-Eval benchmark~\cite{yuan2025opens2v} across all four official domains ($360$ videos per model; $1{,}080$ videos total). We report raw metric means grouped into five functional categories. Note that \textbf{Motion-FB} and \textbf{Motion-LK} measure raw optical-flow displacement magnitude ($\leftrightarrow$: amplitude indicator, not higher-is-better) and should be read alongside \textbf{Smooth.}~($\uparrow$) and \textbf{GME}~($\uparrow$). Symbols \texttt{---} indicate subsets without applicable human-face references or official \texttt{class\_label} metadata.}
\label{tab:opens2v}
\resizebox{\linewidth}{!}{%
\begin{tabular}{ll c ccc ccc c c}
\toprule
 & & \textbf{Visual Quality} & \multicolumn{3}{c}{\textbf{Motion (Quality \& Amplitude)}} & \multicolumn{3}{c}{\textbf{Subject Consistency}} & \textbf{Text Align.} & \textbf{Realism} \\
\cmidrule(lr){3-3} \cmidrule(lr){4-6} \cmidrule(lr){7-9} \cmidrule(lr){10-10} \cmidrule(lr){11-11}
\textbf{Domain} & \textbf{Method} & \textbf{Aes}~$\uparrow$ & \textbf{Smooth.}~$\uparrow$ & \textbf{Motion-FB}~$\leftrightarrow$ & \textbf{Motion-LK}~$\leftrightarrow$ & \textbf{FaceSim-CUR}~$\uparrow$ & \textbf{FaceSim-ARC}~$\uparrow$ & \textbf{Nexus}~$\uparrow$ & \textbf{GME}~$\uparrow$ & \textbf{Natural}~$\uparrow$ \\
\midrule
\multirow{3}{*}{\shortstack[l]{Open-Domain\\(180 cases)}}
 & PDD         & 5.5259 & \textbf{0.9913} & 0.2232 & 4.8244 & 0.5406 & 0.4962 & \textbf{0.0282} & 0.6525 & 4.5574 \\
 & DirectAlign & 5.4663 & 0.9886 & 0.2329 & 5.1657 & \textbf{0.5488} & 0.5028 & 0.0245 & 0.6515 & \textbf{4.5833} \\
 & BlockAlign  & \textbf{5.5591} & 0.9912 & 0.2003 & 4.4738 & 0.5455 & \textbf{0.5053} & 0.0237 & \textbf{0.6554} & 4.5111 \\
\midrule
\multirow{3}{*}{\shortstack[l]{Human-Domain\\(60 cases)}}
 & PDD         & 5.7233 & 0.9919 & 0.2728 & 5.4487 & 0.5581 & 0.5058 & 0.0053 & 0.6559 & 4.3444 \\
 & DirectAlign & 5.6798 & 0.9914 & 0.2812 & 5.6850 & \textbf{0.5692} & \textbf{0.5187} & 0.0048 & 0.6530 & \textbf{4.5167} \\
 & BlockAlign  & \textbf{5.7642} & \textbf{0.9930} & 0.2360 & 5.0215 & 0.5664 & 0.5135 & \textbf{0.0058} & \textbf{0.6589} & 4.3000 \\
\midrule
\multirow{3}{*}{\shortstack[l]{Single-Domain\\(90 cases)}}
 & PDD         & 5.5847 & 0.9903 & 0.3000 & 5.3680 & 0.5654 & 0.5186 & \textbf{0.0311} & 0.6364 & 4.5148 \\
 & DirectAlign & 5.5312 & 0.9878 & 0.2960 & 5.5276 & \textbf{0.5765} & \textbf{0.5292} & 0.0265 & 0.6357 & \textbf{4.6000} \\
 & BlockAlign  & \textbf{5.6187} & \textbf{0.9917} & 0.2672 & 4.8514 & 0.5685 & 0.5273 & 0.0229 & \textbf{0.6384} & 4.4370 \\
\midrule
\multirow{3}{*}{\shortstack[l]{Hard-Case-Dev\\(30 cases)}}
 & PDD         & 5.4220 & 0.9917 & 0.3331 & 6.9702 & --- & --- & --- & 0.6400 & \textbf{4.8222} \\
 & DirectAlign & 5.3835 & 0.9889 & 0.3191 & 6.0151 & --- & --- & --- & 0.6397 & 4.7556 \\
 & BlockAlign  & \textbf{5.4534} & \textbf{0.9955} & 0.3084 & 5.8047 & --- & --- & --- & \textbf{0.6414} & 4.7222 \\
\midrule
\multirow{3}{*}{\shortstack[l]{\textbf{Overall}\\(360 cases)}}
 & PDD         & 5.5648 & 0.9912 & 0.2598 & 5.2431 & 0.5527 & 0.5054 & \textbf{0.0259} & 0.6480 & 4.5333 \\
 & DirectAlign & 5.5112 & 0.9889 & 0.2639 & 5.4135 & \textbf{0.5626} & \textbf{0.5149} & 0.0223 & 0.6468 & \textbf{4.5907} \\
 & BlockAlign  & \textbf{5.5994} & \textbf{0.9920} & 0.2320 & 4.7704 & 0.5580 & 0.5139 & 0.0211 & \textbf{0.6506} & 4.4750 \\
\bottomrule
\end{tabular}%
}
\end{table*}

\subsection{OpenS2V-Eval Benchmark}
\label{sec:opens2v}

\textbf{Experimental setup and metric taxonomy.} On OpenS2V-Eval~\cite{yuan2025opens2v}, each model generates all $360$ benchmark videos across the four official subsets released in the public evaluation suite: the core \textit{Open-Domain} benchmark ($180$ cases) and the three official domain-diagnostic splits---\textit{Human-Domain} ($60$ cases), \textit{Single-Domain} ($90$ cases), and \textit{Hard-Case-Dev} ($30$ cases). Following the OpenS2V-Eval metric definitions---while reporting raw, unclipped metric means over all valid records (alongside auxiliary \textbf{Motion-LK} and \textbf{FaceSim-ARC}) rather than the official script's clipped $[0,1]$ rescaling and composite score, which omits \textit{Hard-Case-Dev} and \textit{Overall} totals and positively weights raw optical-flow magnitude---we organize the results into five functional categories in Table~\ref{tab:opens2v}:
\begin{itemize}[leftmargin=1.5em, itemsep=1pt, topsep=2pt]
    \item \textbf{Visual Quality}: \textit{AestheticScore} (\textbf{Aes}~$\uparrow$), evaluating frame-level composition, color fidelity, clarity, and visual appeal on the native aesthetic predictor scale.
    \item \textbf{Motion (Quality \& Amplitude)}: \textit{Motion Smoothness} (\textbf{Smooth.}~$\uparrow$), evaluated by OneAlign to measure temporal continuity and the absence of jitter or abrupt frame transitions; paired with optical-flow amplitude indicators \textbf{Motion-FB} (Farneback dense optical flow, primary amplitude metric) and \textbf{Motion-LK} (Lucas--Kanade sparse feature tracking, auxiliary amplitude metric). Crucially, Motion-FB and Motion-LK measure raw pixel displacement magnitude rather than motion quality (neither higher nor lower is intrinsically better) and must be interpreted jointly with Smoothness and text alignment (GME) to distinguish coherent action dynamics from unstable jitter or static frames.
    \item \textbf{Subject Consistency}: \textit{FaceSim-CUR}~$\uparrow$ (CurricularFace cosine similarity, primary face metric) and \textit{FaceSim-ARC}~$\uparrow$ (ArcFace similarity, auxiliary face metric), evaluated over the $210$ cases containing valid human-face references ($90$ Open, $60$ Human, $60$ Single, and $0$ Hard-Case-Dev); and \textit{NexusScore} (\textbf{Nexus}~$\uparrow$), which couples open-vocabulary subject localization with GME visual-text similarity over valid localized instances (\textit{Hard-Case-Dev} is omitted as official metadata does not provide \texttt{class\_label} annotations).
    \item \textbf{Text Alignment}: \textit{GmeScore} (\textbf{GME}~$\uparrow$), measuring semantic similarity between the text prompt and generated video frames via GME-Qwen2-VL.
    \item \textbf{Realism}: \textit{NaturalScore} (\textbf{Natural}~$\uparrow$, $1$--$5$ scale), averaging three independent GPT-4o evaluations over $16$ uniformly sampled frames per video.
\end{itemize}

\textbf{Results and analysis.} Table~\ref{tab:opens2v} highlights three key findings across the five functional categories:
\begin{enumerate}[leftmargin=1.5em, itemsep=2pt, topsep=2pt]
    \item \textbf{Consistent gains in visual quality, text alignment, and motion smoothness}: Across all four domains and the $360$-case overall aggregate, BlockAlign achieves the highest visual aesthetics (\textbf{Aes}: $5.5994$ overall vs.\ $5.5648$ for PDD and $5.5112$ for DirectAlign) and the strongest text--video semantic alignment (\textbf{GME}: $0.6506$ vs.\ $0.6480$ and $0.6468$). Simultaneously, BlockAlign attains the highest overall \textbf{Motion Smoothness} ($0.9920$ vs.\ $0.9912$ for PDD and $0.9889$ for DirectAlign), with pronounced gains on \textit{Human-Domain} ($0.9930$) and \textit{Hard-Case-Dev} ($0.9955$) while maintaining healthy optical-flow displacement (\textbf{Motion-FB}: $0.2320$, \textbf{Motion-LK}: $4.7704$). In contrast, DirectAlign degrades both \textbf{Aes} and \textbf{Smoothness} below the unaligned PDD baseline across every subset; its higher raw optical-flow magnitude ($0.2639$ / $5.4135$) paired with the lowest smoothness ($0.9889$) and lowest prompt alignment ($0.6468$) reflects high-frequency structural jitter and non-rigid deformation artifacts caused by unfused sub-head supervision rather than coherent motion.
    \item \textbf{Subject consistency across in-domain human scenes vs.\ out-of-domain objects}: Compared with the unaligned PDD baseline, BlockAlign consistently improves facial identity similarity across all applicable subsets on both \textbf{FaceSim-CUR} ($0.5580$ vs.\ $0.5527$ overall) and \textbf{FaceSim-ARC} ($0.5139$ vs.\ $0.5054$). Note that DirectAlign's marginally higher 2D face cosine similarity ($0.5626$) coincides with its lower prompt following (\textbf{GME}: $0.6468$), as failing to execute prompted head turns or viewpoint changes keeps faces closer to the frontal reference pose. Crucially, on \textit{Human-Domain}---which matches our human-centric training distribution---BlockAlign achieves the highest raw \textbf{NexusScore} ($0.0058$ vs.\ $0.0053$ for PDD and $0.0048$ for DirectAlign) and yields the most non-zero subject localizations across both \textit{Human-Domain} ($38/60$ valid cases vs.\ $34/60$ for PDD and DirectAlign) and the full $330$-case benchmark ($177$ vs.\ $172$ and $170$), confirming superior multi-attribute subject binding and fewer missed detections. On general non-human objects in \textit{Open-Domain} and \textit{Single-Domain}, our commercial aesthetic alignment enhances scene lighting and color saturation (\textbf{Aes}~$\uparrow$), which slightly shifts raw crop-level feature similarity relative to flat-lit object references (a $\sim\!0.001$ difference in non-zero raw means after the official $0.05$ outlier clipping: $0.0225$ vs.\ $0.0235$ for PDD and $0.0227$ for DirectAlign).
    \item \textbf{Commercial aesthetic refinement vs.\ 16-frame NaturalScore}: While all three methods achieve high realism scores within a narrow range ($4.48$--$4.59$ on a $1$--$5$ scale), BlockAlign's slightly lower GPT-4o \textbf{NaturalScore} ($4.4750$ vs.\ $4.5333$ for PDD and $4.5907$ for DirectAlign) stems from two factors. First, our post-training explicitly optimizes commercial e-commerce aesthetics---sharper textures, richer color saturation, and studio-like contrast---which significantly boosts \textbf{Aes} ($5.5994$) but can be penalized by GPT-4o's zero-shot ``unretouched natural footage'' rubric in favor of flatter, less saturated renderings (such as DirectAlign, which scores lowest in \textbf{Aes} at $5.5112$). Second, because NaturalScore evaluates only $16$ sparsely sampled static frames per video, it is largely insensitive to inter-frame jitter and temporal flickering, which are instead captured by continuous video metrics (\textbf{Smooth.}) and strongly penalized in full-video human evaluations on MiniMax-H3 (where BlockAlign outperforms PDD with a $59.7\%$ non-tie win rate and beats DirectAlign with $64.0\%$ vs.\ $17.3\%$ wins/losses, or $78.7\%$ non-tie win rate).
\end{enumerate}

\begin{table*}[t]
\centering
\caption{Task-family and overall quantitative evaluation on the public \textbf{OmniVBench} benchmark~\cite{omnivbench2026} ($813$ videos per model across seven task families). Each task family score averages Reference Fidelity (\textbf{RF}), Instruction Realization (\textbf{IR}), and Video Quality (\textbf{VQ}), and \textbf{Overall} denotes the macro-average across the seven families. Note that only \textbf{RF} and \textbf{IR} query a proprietary VLM judge, whereas \textbf{VQ} (\textbf{Technical}, \textbf{Aesthetic}, and \textbf{Physics}) is computed directly using fixed open-weight evaluation models and remains identical across VLM judges. Top block: 5-second generation results under the default \texttt{Gemini-3.1-Pro} judge for \textbf{RF}/\textbf{IR}, comparing our 10-step \textbf{BlockAlign} ($24\,\text{fps}$, $768p$) against official leaderboard entries. Middle block: controlled 8-second comparison ($24\,\text{fps}$, $768p$) under the \texttt{Gemini-3.1-Pro} judge for \textbf{RF}/\textbf{IR}. Bottom block: cross-judge verification using \texttt{GPT-4o} (the VLM judge used in OpenS2V-Eval) for \textbf{RF}/\textbf{IR} combined with the same fixed-model \textbf{VQ} scores. Best results within each 8-second comparison group (and top 5-second leaderboard entries) are highlighted in \textbf{bold}.}
\label{tab:omnivbench_main}
\resizebox{\linewidth}{!}{%
\begin{tabular}{ll ccccccc c}
\toprule
\textbf{Eval. Protocol (\textbf{RF}/\textbf{IR} Judge)} & \textbf{Model} & \textbf{Content} & \textbf{Motion} & \textbf{Style} & \textbf{Structure} & \textbf{Narrative} & \textbf{Multi-content} & \textbf{Cross-aspect} & \textbf{Overall}~$\uparrow$ \\
\midrule
\multirow{7}{*}{\shortstack[l]{5-second Generation\\(Official Leaderboard \& Ours,\\ \texttt{Gemini-3.1-Pro} for \textbf{RF}/\textbf{IR})}}
 & Seedance 2.5         & 78.88 & 66.10 & 67.13 & \textbf{73.35} & 73.97 & \textbf{77.77} & 71.53 & \textbf{72.68} \\
 & MiniMax-H3           & 78.86 & 65.32 & 66.24 & 72.94 & 73.90 & 77.22 & \textbf{72.36} & 72.41 \\
 & Seedance 2.0         & \textbf{79.00} & 63.05 & 63.73 & 68.36 & 73.76 & 76.63 & 71.83 & 70.91 \\
 & Happy Horse 1.0      & 75.90 & \textbf{68.58} & 65.62 & 69.46 & 72.34 & 74.62 & 69.71 & 70.89 \\
 & Gemini Omni          & 75.02 & 62.25 & \textbf{69.24} & 70.13 & \textbf{75.11} & 73.36 & 70.18 & 70.76 \\
 & Kling 3.0 Omni       & 75.99 & 64.20 & 55.90 & 71.12 & 68.59 & 75.98 & 68.33 & 68.59 \\
 & \textbf{BlockAlign (10-step, 5-second)} & 77.62 & 60.81 & 63.80 & 71.40 & 70.35 & 76.65 & 70.34 & 70.14 \\
\midrule
\multirow{3}{*}{\shortstack[l]{8-second Generation\\(Standardized 10-step PDD,\\ \texttt{Gemini-3.1-Pro} for \textbf{RF}/\textbf{IR})}}
 & PDD (10-step, 8-second)        & 76.81 & \textbf{60.52} & 64.23 & 71.51 & 71.59 & \textbf{77.63} & 70.93 & 70.46 \\
 & DirectAlign (10-step, 8-second)& \textbf{77.29} & 59.15 & \textbf{65.51} & 71.03 & \textbf{72.76} & 76.93 & \textbf{70.95} & \textbf{70.52} \\
 & \textbf{BlockAlign (10-step, 8-second)} & 77.16 & 59.93 & 63.18 & \textbf{72.32} & 72.20 & 76.85 & 70.65 & 70.33 \\
\midrule
\multirow{4}{*}{\shortstack[l]{Cross-Judge Verification\\(Standardized 10-step PDD,\\ \texttt{GPT-4o} for \textbf{RF}/\textbf{IR})}}
 & PDD (10-step, 8-second)        & 80.28 & 70.50 & 72.10 & 79.69 & \textbf{79.22} & 80.35 & 78.51 & 77.24 \\
 & DirectAlign (10-step, 8-second)& 80.12 & 70.86 & \textbf{72.77} & 79.56 & 79.10 & 80.24 & 78.18 & 77.26 \\
 & \textbf{BlockAlign (10-step, 8-second)} & \textbf{80.60} & \textbf{71.39} & 72.47 & \textbf{79.92} & 79.18 & \textbf{80.44} & \textbf{78.59} & \textbf{77.51} \\
 \cmidrule(lr){2-10}
 & \textbf{BlockAlign (10-step, 5-second)} & 80.52 & 71.79 & 72.05 & 79.87 & 79.90 & 80.50 & 79.03 & 77.66 \\
\bottomrule
\end{tabular}%
}
\end{table*}

\begin{table*}[t]
\centering
\caption{Fine-grained breakdown on \textbf{OmniVBench} across Reference Fidelity (\textbf{RF}), Instruction Realization (\textbf{IR}), Video Quality (\textbf{VQ}), and the three \textbf{VQ} sub-dimensions (\textbf{Technical}, \textbf{Aesthetic}, and \textbf{Physics}) for all seven task families and the 7-family \textbf{Macro} average. For \textbf{RF} and \textbf{IR}, we report both the default \texttt{Gemini-3.1-Pro} judge and cross-judge verification with \texttt{GPT-4o} (the VLM judge used in OpenS2V-Eval). Within our controlled 8-second comparison, DirectAlign degrades \textit{Technical} and \textit{Aesthetic} quality below the unaligned PDD baseline across every single task family ($7/7$) as well as macro \textbf{VQ} ($41.96$ vs.\ $42.12$), whereas BlockAlign (8-second) achieves the best \textit{Technical}, \textit{Aesthetic}, and \textbf{VQ} scores across all seven families ($7/7$, bolded within the 8-second group). We also report \textbf{BlockAlign (5-second)} at the bottom of each block.}
\label{tab:omnivbench_breakdown}
\resizebox{\linewidth}{!}{%
\begin{tabular}{ll ccccccc c}
\toprule
\textbf{Metric / Sub-metric} & \textbf{Method} & \textbf{Content} (112) & \textbf{Motion} (95) & \textbf{Style} (43) & \textbf{Structure} (108) & \textbf{Narrative} (120) & \textbf{Multi-content} (112) & \textbf{Cross-aspect} (223) & \textbf{Macro}~$\uparrow$ \\
\midrule
\multirow{4}{*}{\shortstack[l]{\textbf{Reference Fidelity (RF)}~$\uparrow$\\(\texttt{Gemini-3.1-Pro} Judge)}}
 & PDD (8-second)        & 91.15 & \textbf{46.54} & 61.05 & 78.40 & \textbf{87.90} & \textbf{93.75} & \textbf{79.88} & \textbf{76.95} \\
 & DirectAlign (8-second)& \textbf{92.41} & 41.84 & \textbf{65.70} & 75.89 & 87.57 & 92.95 & 79.79 & 76.59 \\
 & \textbf{BlockAlign (8-second)} & 91.52 & 43.64 & 59.30 & \textbf{79.90} & 87.29 & 92.22 & 79.82 & 76.24 \\
 \cmidrule(lr){2-10}
 & \textbf{BlockAlign (5-second)} & 94.01 & 47.81 & 63.37 & 78.78 & 81.93 & 92.06 & 79.19 & 76.74 \\
\midrule
\multirow{4}{*}{\shortstack[l]{\textbf{Reference Fidelity (RF)}~$\uparrow$\\(\texttt{GPT-4o} Judge)}}
 & PDD (8-second)        & \textbf{99.11} & 71.05 & 76.16 & 98.26 & \textbf{98.96} & \textbf{99.65} & \textbf{94.80} & 91.14 \\
 & DirectAlign (8-second)& 99.00 & 72.89 & \textbf{79.07} & 97.86 & 98.54 & 99.47 & 94.50 & \textbf{91.62} \\
 & \textbf{BlockAlign (8-second)} & \textbf{99.11} & \textbf{73.29} & 77.33 & \textbf{98.61} & 98.33 & 99.59 & 94.06 & 91.47 \\
 \cmidrule(lr){2-10}
 & \textbf{BlockAlign (5-second)} & 99.11 & 74.47 & 76.16 & 97.51 & 99.06 & 99.62 & 95.15 & 91.58 \\
\midrule
\multirow{4}{*}{\shortstack[l]{\textbf{Instruction Realization (IR)}~$\uparrow$\\(\texttt{Gemini-3.1-Pro} Judge)}}
 & PDD (8-second)        & 96.06 & 92.60 & 88.43 & 94.69 & 85.72 & \textbf{96.85} & 90.72 & 92.15 \\
 & DirectAlign (8-second)& \textbf{96.36} & \textbf{93.80} & \textbf{88.73} & \textbf{96.03} & \textbf{89.94} & 95.57 & \textbf{91.24} & \textbf{93.10} \\
 & \textbf{BlockAlign (8-second)} & 95.72 & 93.24 & 86.62 & 95.12 & 87.85 & 95.40 & 89.27 & 91.89 \\
 \cmidrule(lr){2-10}
 & \textbf{BlockAlign (5-second)} & 94.62 & 91.90 & 84.73 & 93.29 & 87.25 & 95.02 & 89.11 & 90.85 \\
\midrule
\multirow{4}{*}{\shortstack[l]{\textbf{Instruction Realization (IR)}~$\uparrow$\\(\texttt{GPT-4o} Judge)}}
 & PDD (8-second)        & 98.08 & \textbf{97.96} & \textbf{97.14} & \textbf{99.98} & 98.59 & \textbf{99.00} & \textbf{98.35} & \textbf{98.44} \\
 & DirectAlign (8-second)& 97.98 & 97.69 & 96.65 & 99.84 & 98.40 & \textbf{99.00} & 97.89 & 98.21 \\
 & \textbf{BlockAlign (8-second)} & \textbf{98.32} & 97.90 & 96.48 & 99.95 & \textbf{98.65} & 98.78 & 98.32 & 98.34 \\
 \cmidrule(lr){2-10}
 & \textbf{BlockAlign (5-second)} & 98.24 & 98.16 & 96.67 & 99.98 & 98.78 & 99.00 & 99.23 & 98.58 \\
\midrule
\multirow{4}{*}{\textbf{Video Quality (VQ)}~$\uparrow$}
 & PDD (8-second)        & 43.67 & 42.49 & 43.00 & 40.82 & 40.12 & 42.39 & 42.38 & 42.12 \\
 & DirectAlign (8-second)& 43.38 & 42.00 & 42.59 & 40.98 & 40.37 & 42.26 & 42.16 & 41.96 \\
 & \textbf{BlockAlign (8-second)} & \textbf{44.36} & \textbf{42.98} & \textbf{43.62} & \textbf{41.21} & \textbf{40.57} & \textbf{42.95} & \textbf{43.39} & \textbf{42.73} \\
 \cmidrule(lr){2-10}
 & \textbf{BlockAlign (5-second)} & 44.22 & 42.73 & 43.30 & 42.13 & 41.86 & 42.87 & 42.72 & 42.83 \\
\midrule
\multirow{4}{*}{\quad $\hookrightarrow$ \textit{VQ: Technical}~$\uparrow$}
 & PDD (8-second)        & 19.63 & 16.57 & 15.86 & 17.50 & 17.13 & 20.11 & 18.42 & 17.89 \\
 & DirectAlign (8-second)& 18.92 & 16.17 & 15.32 & 17.30 & 16.84 & 19.61 & 17.81 & 17.42 \\
 & \textbf{BlockAlign (8-second)} & \textbf{20.13} & \textbf{16.84} & \textbf{16.41} & \textbf{18.01} & \textbf{17.80} & \textbf{20.69} & \textbf{18.99} & \textbf{18.41} \\
 \cmidrule(lr){2-10}
 & \textbf{BlockAlign (5-second)} & 20.53 & 17.07 & 16.31 & 17.65 & 17.89 & 21.08 & 19.19 & 18.53 \\
\midrule
\multirow{4}{*}{\quad $\hookrightarrow$ \textit{VQ: Aesthetic}~$\uparrow$}
 & PDD (8-second)        & 46.97 & 48.03 & 50.09 & 45.64 & 47.00 & 46.23 & 47.03 & 47.28 \\
 & DirectAlign (8-second)& 46.63 & 47.55 & 49.45 & 45.32 & 46.76 & 45.82 & 46.65 & 46.88 \\
 & \textbf{BlockAlign (8-second)} & \textbf{47.82} & \textbf{48.87} & \textbf{50.85} & \textbf{46.50} & \textbf{47.82} & \textbf{47.02} & \textbf{47.87} & \textbf{48.11} \\
 \cmidrule(lr){2-10}
 & \textbf{BlockAlign (5-second)} & 48.39 & 49.09 & 50.99 & 46.26 & 47.81 & 47.38 & 48.01 & 48.28 \\
\midrule
\multirow{4}{*}{\quad $\hookrightarrow$ \textit{VQ: Physics}~$\uparrow$}
 & PDD (8-second)        & 64.41 & 62.87 & 63.04 & 59.33 & 56.23 & 60.82 & 61.69 & 61.20 \\
 & DirectAlign (8-second)& 64.59 & 62.29 & 63.00 & \textbf{60.31} & \textbf{57.50} & \textbf{61.34} & 62.01 & 61.58 \\
 & \textbf{BlockAlign (8-second)} & \textbf{65.14} & \textbf{63.24} & \textbf{63.61} & 59.11 & 56.08 & 61.14 & \textbf{63.31} & \textbf{61.66} \\
 \cmidrule(lr){2-10}
 & \textbf{BlockAlign (5-second)} & 63.73 & 62.03 & 62.62 & 62.49 & 59.87 & 60.13 & 60.95 & 61.69 \\
\bottomrule
\end{tabular}%
}
\end{table*}

\subsection{OmniVBench Benchmark}
\label{sec:omnivbench}

\textbf{Experimental setup and metric hierarchy.} We further evaluate PDD, DirectAlign, and BlockAlign on \textbf{OmniVBench}~\cite{omnivbench2026}, a comprehensive omni-reference video generation benchmark comprising $813$ evaluation prompts across seven task families: \textit{Content} ($112$ cases), \textit{Motion} ($95$ cases), \textit{Style} ($43$ cases), \textit{Structure} ($108$ cases), \textit{Narrative} ($120$ cases), \textit{Multi-content} ($112$ cases), and \textit{Cross-aspect} ($223$ cases). Each model generates all $813$ videos at our standardized 8-second duration ($24\,\text{fps}$, $768p$) using identical 10-step PDD inference schedules and random seeds; in addition, we evaluate BlockAlign under a 5-second setting ($24\,\text{fps}$, $768p$) with the exact same prompts, references, and seeds to match the 5-second horizon of the official leaderboard. Following the official OmniVBench evaluation pipeline, all metrics are normalized to a $0$--$100$ scale:
\begin{itemize}[leftmargin=1.5em, itemsep=1pt, topsep=2pt]
    \item \textbf{VLM-Judged Semantic Dimensions (\textbf{RF} \& \textbf{IR})}: Two axes query a proprietary VLM judge: \textit{Reference Fidelity} (\textbf{RF}~$\uparrow$), measuring identity, attribute, and visual preservation of all conditioned references; and \textit{Instruction Realization} (\textbf{IR}~$\uparrow$), measuring completion of prompted actions, camera movements, and multi-stage events. We report both the official default \texttt{Gemini-3.1-Pro} judge and cross-judge verification using \texttt{GPT-4o} (the VLM judge adopted in OpenS2V-Eval).
    \item \textbf{Model-Based Video Quality (\textbf{VQ})}: Unlike \textbf{RF} and \textbf{IR}, \textit{Video Quality} (\textbf{VQ}~$\uparrow$) does not call a proprietary VLM judge; it is computed directly from fixed open-weight evaluation models as the unweighted mean of three sub-metrics: \textbf{Technical} quality (official sigmoid normalization with center $0.1107$ and scale $0.07355$), \textbf{Aesthetic} appeal (raw score $\times 10$), and \textbf{Physics} plausibility (raw $1$--$5$ score $\times 20$).
    \item \textbf{Task Family \& Overall Aggregation}: Each task family score is the arithmetic mean of its family-level \textbf{RF}, \textbf{IR}, and \textbf{VQ}. The model-level \textbf{Overall} (Macro) score is the unweighted macro-average across all seven task families (equivalently, the mean of macro-averaged \textbf{RF}, \textbf{IR}, and \textbf{VQ}).
\end{itemize}

\textbf{Results and analysis.} Tables~\ref{tab:omnivbench_main} and~\ref{tab:omnivbench_breakdown} present the task-family and fine-grained results on OmniVBench, revealing four key findings:
\begin{enumerate}[leftmargin=1.5em, itemsep=2pt, topsep=2pt]
    \item \textbf{Strong multi-reference composition and structural control}: Under the default \texttt{Gemini-3.1-Pro} judge (Table~\ref{tab:omnivbench_main}), on \textit{Multi-content}---the task family most directly aligned with our core multi-reference subject--scene conditioning objective---BlockAlign ($76.65$ under 5-second and $76.85$ under 8-second generation) outperforms \textbf{Seedance~2.0} ($76.63$), \textbf{Kling~3.0 Omni} ($75.98$), \textbf{Happy Horse~1.0} ($74.62$), and \textbf{Gemini Omni} ($73.36$) on the official leaderboard, ranking second only to \textbf{Seedance~2.5} ($77.77$) and the base \textbf{MiniMax-H3} model ($77.22$). A similar hierarchy holds on \textit{Structure} ($71.40$ at 5-second; $72.32$ at 8-second, ranking first among 10-step models) and \textit{Content} ($77.62$ at 5-second; $77.16$ at 8-second), where BlockAlign surpasses Kling~3.0 Omni, Happy Horse~1.0, and Gemini Omni with 10-step inference.
        \item \textbf{DirectAlign universally degrades Technical and Aesthetic quality ($7/7$ families)}: In the controlled 8-second comparison (Table~\ref{tab:omnivbench_breakdown}), although DirectAlign slightly raises \texttt{Gemini-3.1-Pro} \textbf{IR} ($93.10$ vs.\ $92.15$ for PDD), supervising unfused sub-interval velocity heads across sub-steps violates PDD's fused block-average transition semantics and degrades \textit{Technical} quality and \textit{Aesthetic} appeal below the unaligned PDD baseline across every single task family ($7/7$)---reducing \textit{Technical} from $17.89$ to $17.42$, \textit{Aesthetic} from $47.28$ to $46.88$, and overall macro \textbf{VQ} from $42.12$ to $41.96$ (lower in $5/7$ families).
        \item \textbf{BlockAlign achieves across-the-board VQ superiority ($7/7$ families) across both horizons}: By restricting reward alignment to valid block-boundary transitions, BlockAlign (8-second) achieves the highest \textbf{VQ} ($42.73$, $+0.61$ over PDD and $+0.77$ over DirectAlign), \textbf{Technical} ($18.41$, $+0.52$ over PDD and $+0.99$ over DirectAlign), and \textbf{Aesthetic} ($48.11$, $+0.83$ over PDD and $+1.23$ over DirectAlign) in \textit{all seven task families ($7/7$)}, alongside the best macro \textbf{Physics} score ($61.66$), trading off a minor shift in \texttt{Gemini-3.1-Pro} \textbf{RF}/\textbf{IR} ($70.33$ vs.\ $70.46$ macro Overall for PDD, while leading on \textit{Structure} at $72.32$). Moreover, BlockAlign (5-second) maintains virtually identical visual quality gains ($42.83$ \textbf{VQ}, $18.53$ \textbf{Technical}, $48.28$ \textbf{Aesthetic}, and $61.69$ \textbf{Physics}), confirming duration-robust alignment across both 5-second and 8-second generation.
            \item \textbf{Cross-judge verification with \texttt{GPT-4o} and judge trade-off analysis}: Under the default \texttt{Gemini-3.1-Pro} judge, BlockAlign's slight $0.13$-point shift in 8-second macro \textbf{Overall} relative to PDD ($70.33$ vs.\ $70.46$, despite leading on \textit{Structure} at $72.32$ and improving \textit{Content} to $77.16$) is primarily driven by the small $43$-case \textit{Style} subset ($63.18$ vs.\ $64.23$), where our commercial photorealistic post-training trades off stylization scores on abstract/artistic prompts. When evaluating \textbf{RF} and \textbf{IR} with \texttt{GPT-4o} (Tables~\ref{tab:omnivbench_main} and~\ref{tab:omnivbench_breakdown}), DirectAlign degrades \textbf{IR} below PDD in six of seven families and below BlockAlign in five ($98.21$ macro vs.\ $98.44$ for PDD and $98.34$ for BlockAlign), whereas BlockAlign improves \textbf{RF} over PDD ($91.47$ vs.\ $91.14$) and achieves the best 8-second \textbf{Overall} score of \textbf{77.51} (ranking first in $5/7$ task families, including \textit{Style} over PDD), with 5-second generation further reaching \textbf{77.66} ($91.58$ \textbf{RF}, $98.58$ \textbf{IR}, and $42.83$ \textbf{VQ}).
\end{enumerate}

\begin{table*}[t]
\centering
\small
\caption{
Intra-condition generation diversity across five random seeds under identical reference images and text prompts. We report the mean and standard deviation of pairwise feature distances across test conditions (higher values indicate greater feature-space dispersion). Distances are computed independently within each encoder's feature space.
}
\label{tab:generation_diversity}
\begin{tabular}{clcccc}
\toprule
& & \multicolumn{2}{c}{V-JEPA 2}
  & \multicolumn{2}{c}{VideoMAE v2} \\
\cmidrule(lr){3-4}\cmidrule(lr){5-6}
Backbone & Method
& Cosine distance & $\ell_2$ distance
& Cosine distance & $\ell_2$ distance \\
\midrule
\multirow{2}{*}{RefVid-V4}
& PDD
& $0.0238 \pm 0.0067$
& $9.32 \pm 1.31$
& $0.0438 \pm 0.0158$
& $66.94 \pm 9.50$ \\
& BlockAlign
& $0.0258 \pm 0.0072$
& $9.74 \pm 1.37$
& $0.0453 \pm 0.0159$
& $68.90 \pm 8.89$ \\
\midrule
\multirow{2}{*}{MiniMax-H3}
& PDD
& $0.0308 \pm 0.0094$
& $10.65 \pm 1.54$
& $0.0432 \pm 0.0139$
& $69.06 \pm 9.22$ \\
& BlockAlign
& $0.0293 \pm 0.0087$
& $10.37 \pm 1.48$
& $0.0420 \pm 0.0140$
& $67.74 \pm 9.51$ \\
\bottomrule
\end{tabular}
\end{table*}

\section{Generation Diversity Analysis}
\label{sec:diversity}

Because preference post-training can potentially concentrate the learned distribution and induce mode collapse, we evaluate whether BlockAlign preserves intra-condition sample diversity across multiple random seeds. Extending the summary in Sec.~\ref{sec:quantitative_results} of the main paper, we assess generation diversity both quantitatively in spatiotemporal video-feature spaces and qualitatively across multi-seed rollouts.

\subsection{Feature-Space Diversity}

Following PDD~\cite{pdd}, we quantify reference-conditioned generation diversity using spatiotemporal feature embeddings extracted by V-JEPA 2~\cite{assran2025v} and VideoMAE v2~\cite{wang2023videomae}. For each test condition, we generate five videos using distinct random seeds while holding the text prompt, reference images, and inference settings fixed. We then compute pairwise cosine and Euclidean ($\ell_2$) distances across all $\binom{5}{2} = 10$ unordered video pairs per condition, averaging first within each condition and then across all test conditions. Both the base PDD model and BlockAlign are evaluated under identical conditions and random seed sets.

As shown in Table~\ref{tab:generation_diversity}, BlockAlign maintains feature-space dispersion closely comparable to the original PDD checkpoints across both backbones. On RefVid-V4, all four distance metrics increase slightly ($+2.9\%$ to $+8.4\%$), whereas on MiniMax-H3 they decrease marginally ($-1.9\%$ to $-4.9\%$), remaining well within one standard deviation of the base model. Notably, because spatiotemporal feature distances also respond to severe structural deformations or visual artifacts—which artificially inflate pairwise distances in unaligned models—the slight numerical decrease on MiniMax-H3 partly reflects the removal of erratic anatomical defects rather than a loss of semantic variation or motion amplitude. Rather than relying on feature distances alone, combining this preserved spatiotemporal dispersion with our quantitative gains in action completion (PF) and multi-shot dynamics (ST) (Tables~\ref{tab:auto_absolute}--\ref{tab:auto_pairwise}) and the matched-timestamp visual trajectories below confirms that BlockAlign resolves structural defects without degenerating into static frames or mode collapse.

\begin{figure*}[t]
    \centering
    \includegraphics[width=\textwidth]
    {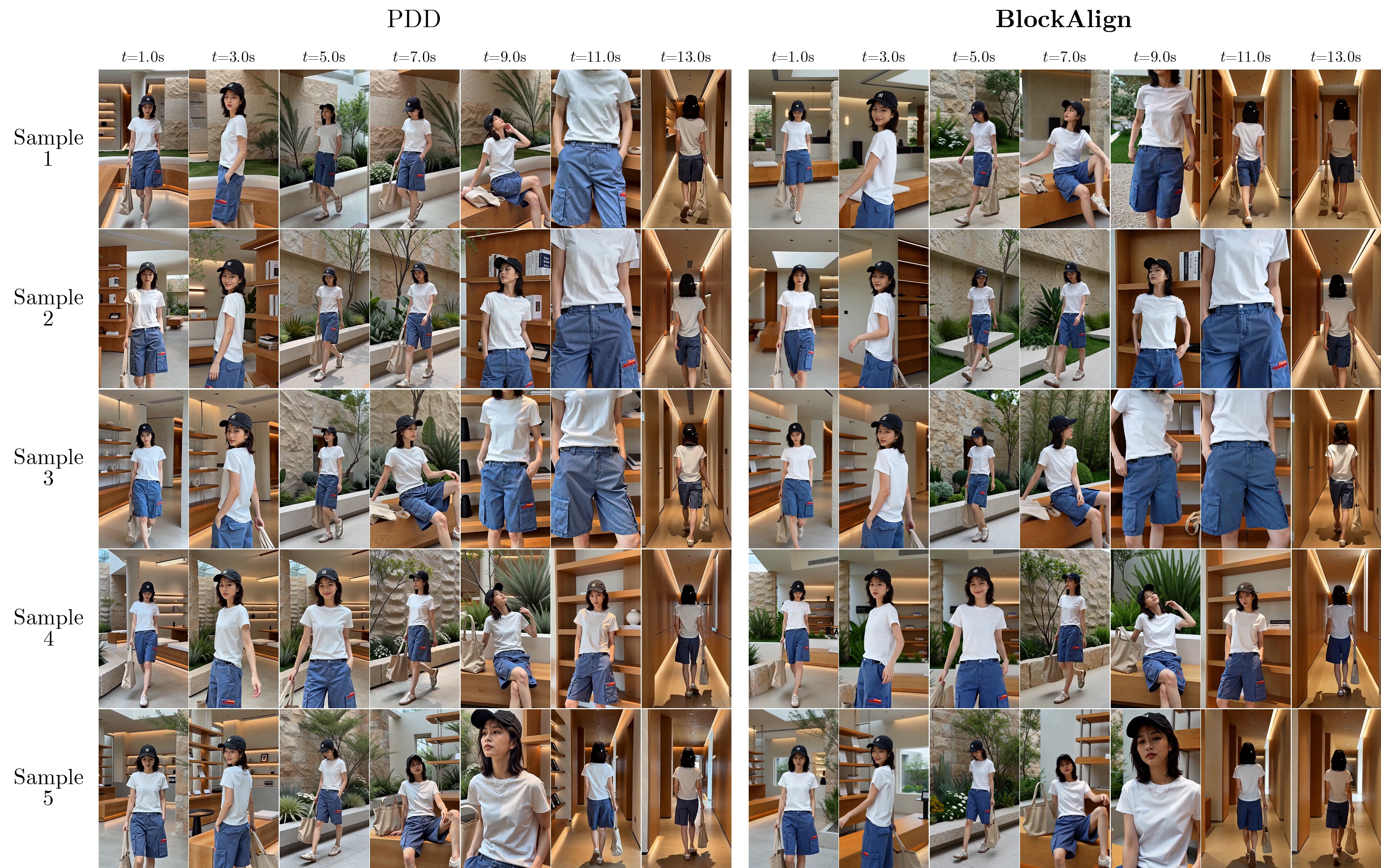}
    \caption{
    Qualitative diversity comparison on RefVid-V4 (Case 1). Each row shows a video generated from a distinct random seed, and columns show matched timestamps. Corresponding rows for PDD and BlockAlign share identical seeds, reference images, and prompts. Both models produce diverse poses, camera framings, subject placements, and indoor--outdoor scene realizations while preserving the referenced identity and outfit, with no sign of mode collapse after BlockAlign post-training. Best viewed zoomed in.
    }
    \label{fig:diversity_hh_case1}
\end{figure*}

\begin{figure*}[t]
    \centering
    \includegraphics[width=\textwidth]
    {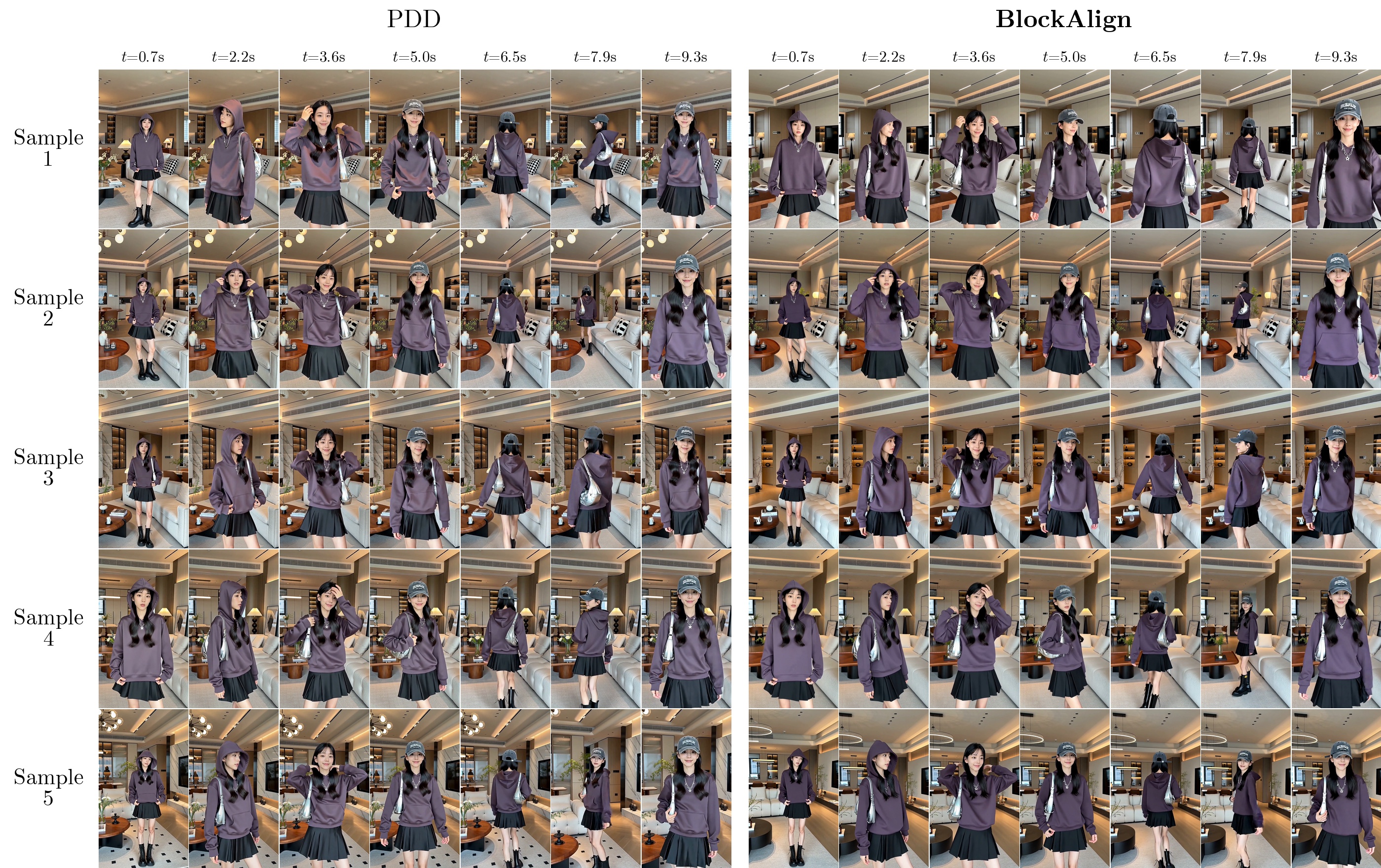}
    \caption{
    Qualitative diversity comparison on RefVid-V4 (Case 2). Across five shared random seeds, BlockAlign retains rich variation in pose, viewpoint, framing, motion trajectory, and spatial layout while faithfully adhering to the reference and prompt conditions. Best viewed zoomed in.
    }
    \label{fig:diversity_hh_case2}
\end{figure*}

\begin{figure*}[t]
    \centering
    \includegraphics[width=\textwidth]
    {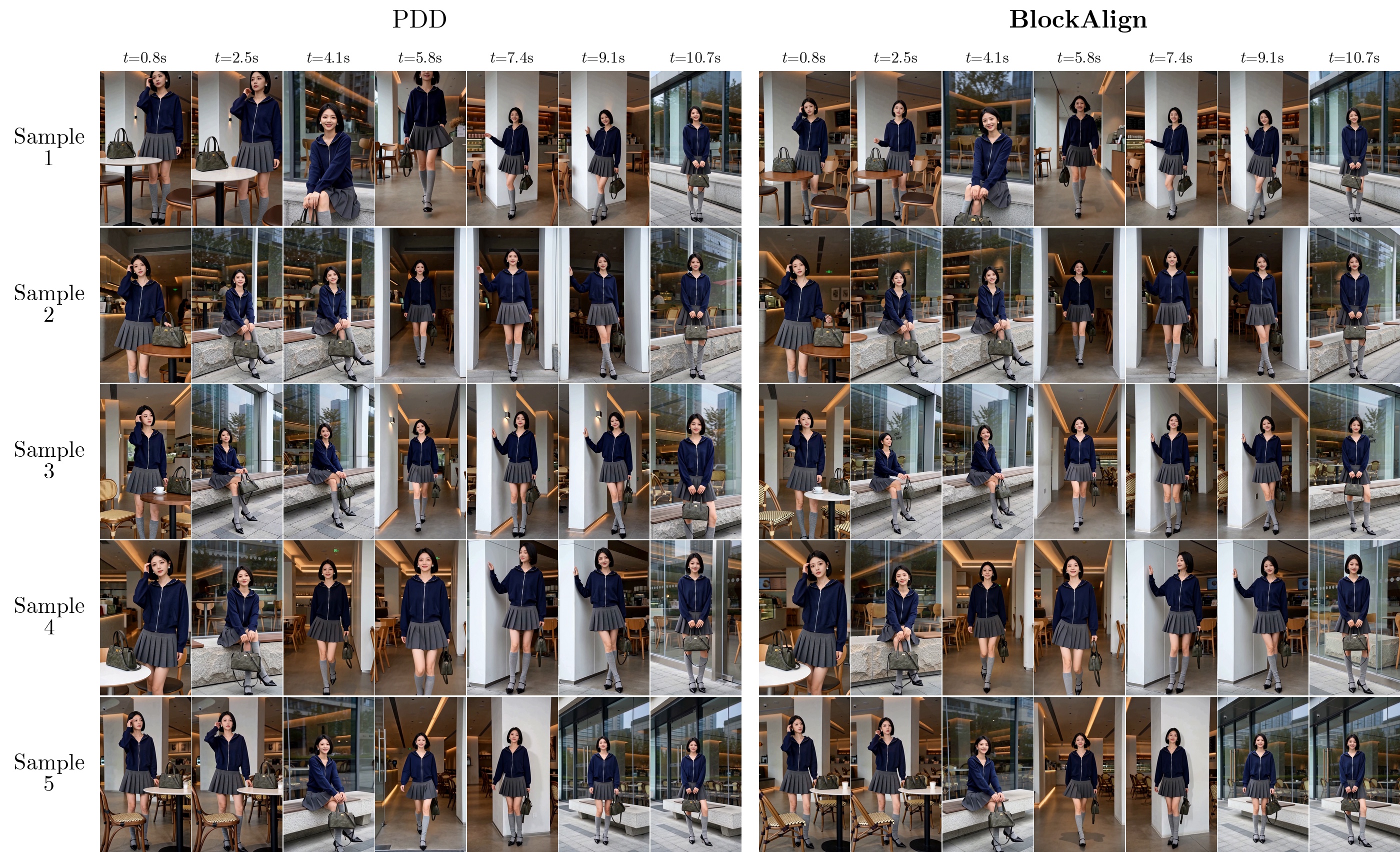}
    \caption{
    Qualitative diversity comparison on MiniMax-H3 (Case 1). Across five shared random seeds and matched timestamps, BlockAlign preserves clear variation in seated and standing postures, walking trajectories, camera distances, subject placement, and handbag orientation while maintaining reference fidelity. Best viewed zoomed in.
    }
    \label{fig:diversity_h3_case1}
\end{figure*}

\begin{figure*}[t]
    \centering
    \includegraphics[width=\textwidth]
    {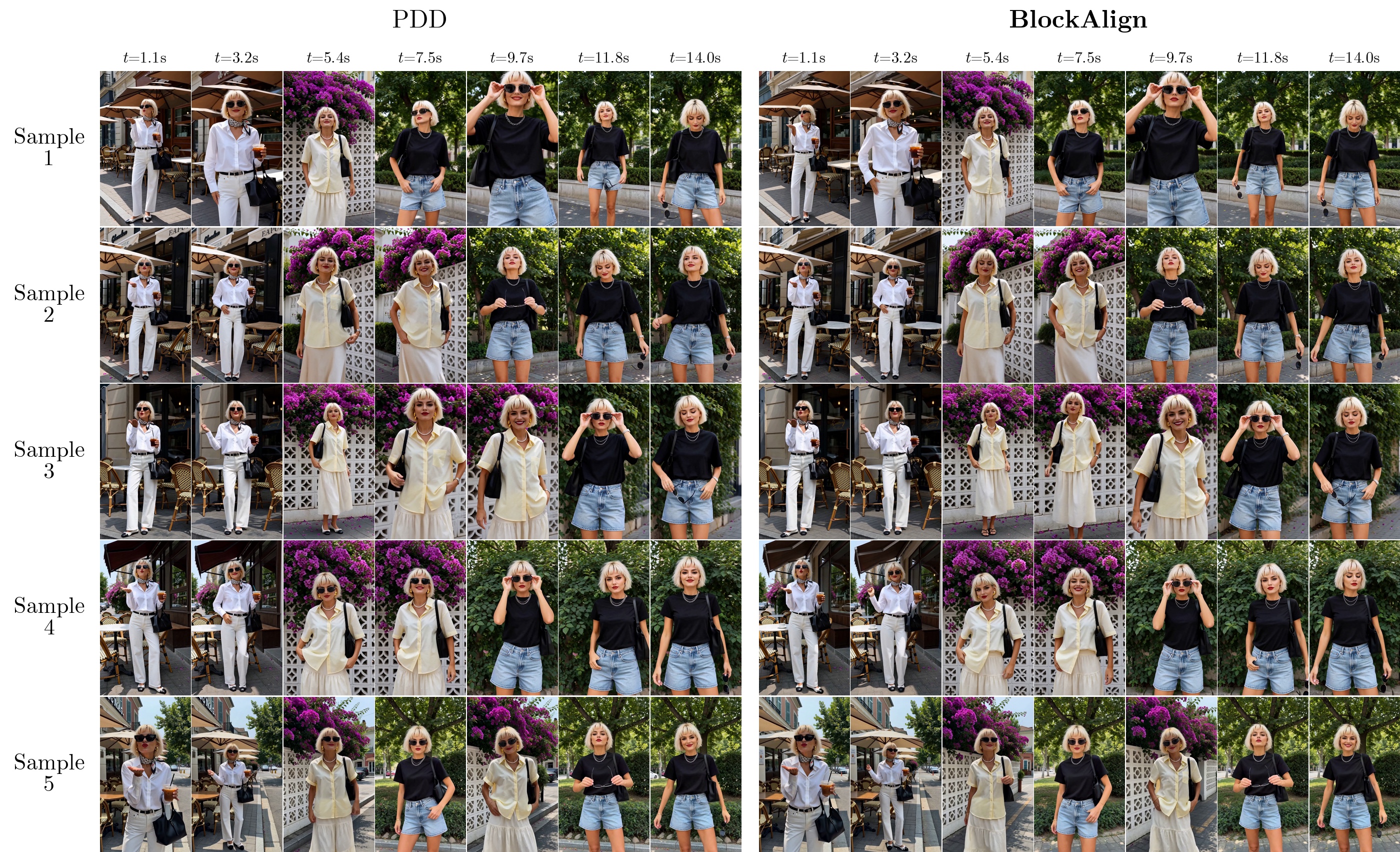}
    \caption{
    Qualitative diversity comparison on MiniMax-H3 (Case 2). Different random seeds yield distinct realizations of the multi-shot prompt—varying in pose, facial expression, gesture, framing, and subject placement—confirming that BlockAlign improves structural plausibility without inducing mode collapse. Best viewed zoomed in.
    }
    \label{fig:diversity_h3_case2}
\end{figure*}

\subsection{Qualitative Diversity Comparison}

Figures~\ref{fig:diversity_hh_case1}--\ref{fig:diversity_h3_case2} visualize five-seed rollouts across four representative test conditions (two per backbone). In each figure, rows correspond to five shared random seeds and columns show frames at matched timestamps, allowing direct comparison of both cross-seed diversity and temporal coherence.

On RefVid-V4 (Figures~\ref{fig:diversity_hh_case1} and~\ref{fig:diversity_hh_case2}), different random seeds yield substantially different body poses, camera distances, subject placements, and background realizations while preserving the referenced identity and garments. BlockAlign fully retains this cross-seed variation while improving local texture clarity and color fidelity, consistent with the slight increase in feature dispersion in Table~\ref{tab:generation_diversity}.

On MiniMax-H3 (Figures~\ref{fig:diversity_h3_case1} and~\ref{fig:diversity_h3_case2}), BlockAlign similarly produces distinct multi-shot realizations across seeds—varying in seated versus standing postures, walking trajectories, camera framing, facial expressions, and hand gestures—without converging to a repetitive template. Combined with Table~\ref{tab:generation_diversity}, these results confirm that BlockAlign improves visual quality, reference consistency, and anatomical plausibility without sacrificing generation diversity.

\section{Rollout and Training Efficiency}
\label{sec:supp_rollout_efficiency}

We provide additional details on the computational efficiency of candidate generation and end-to-end online post-training summarized in Sec.~\ref{sec:efficiency} of the main paper. While the per-rollout sampling speed originates from the few-step PDD generator, naive post-training (DirectAlign) degrades pairwise and human preference below the unaligned baseline. By resolving the post-training--inference mismatch, BlockAlign unlocks practical, high-quality online alignment directly on the deployed PDD model without costly multi-step teacher rollouts or re-distillation.

\paragraph{Forward-evaluation cost.}
For a standard multi-step Diffusion Transformer (DiT) using classifier-free guidance (CFG)~\cite{ho2022classifier}, an $N$-step sampler requires $2N$ backbone forward evaluations per rollout (accounting for both conditional and unconditional branches), totaling $60$--$100$ evaluations for typical schedules of $N = 30$--$50$. In contrast, a fused-head PDD generator with CFG distillation executes only $B$ block transitions per rollout ($B = 10$ in our setting), requiring just $B$ backbone evaluations. Even when compared against a multi-step baseline that already incorporates CFG distillation ($N$ evaluations), native PDD rollout generation reduces backbone evaluations by a factor of $N / B$.

\paragraph{MiniMax-H3 rollout and training efficiency.}
Our main training setup employs 96 NVIDIA H200 GPUs organized into 12 data-parallel (DP=12) replicas of 8-GPU sequence-parallel groups (SP=8). With a global batch size of 12 conditions per iteration, each 8-GPU SP group independently processes a local batch size of 1 condition. We therefore benchmark per-condition rollout and iteration latency on a single 8-GPU SP group on MiniMax-H3 (which already incorporates CFG distillation). Reducing rollout sampling from the default $N = 50$ teacher steps to $B = 10$ PDD blocks decreases the measured generation time for a single 768p ($768\times 1280$), 15-second video (at 24\,fps) on one 8-GPU SP group from approximately 5 minutes to 1 minute, yielding a $5\times$ rollout speedup.

To evaluate the impact on end-to-end online alignment, each 8-GPU SP group sequentially generates a rollout group of $G = 4$ candidates for its assigned condition, followed by hierarchical reward evaluation and synchronous gradient updates across the 12 DP groups (96 GPUs total). On 768p, 15-second videos, replacing 50-step rollouts with 10-step PDD rollouts reduces the measured wall-clock time per training iteration from approximately 24 minutes to 8 minutes. Across the full 1,500-step training schedule on 96 GPUs, total wall-clock training time drops from an extrapolated 600 hours ($57,600$ GPU hours for 50-step rollouts, extrapolated from measured per-iteration latency) to 200 measured hours ($19,200$ GPU hours for 10-step PDD rollouts), achieving a $3\times$ end-to-end online training speedup. Note that the end-to-end training speedup ($3\times$) is slightly lower than the pure rollout speedup ($5\times$) because VLM-based preference scoring and backpropagation updates incur fixed per-iteration overhead independent of the rollout step count.

\begin{table*}[t]
\centering
\caption{Exact 10-block ($B=10$) PDD denoising schedule, specifying the block-start and block-end noise levels $[\sigma_b, \sigma_{b+1}]$, corresponding flow-matching timestamps $[s_b, e_b]$ ($s_b = 1 - \sigma_b$), block span $h_b = e_b - s_b$, functional regime, and unnormalized scheduler weights $\omega_{\text{obj}, b}$ (early-to-mid / mid-to-late).}
\label{tab:block_schedule}
\begin{tabular}{cccccc}
\toprule
Block $b$ & Noise interval $[\sigma_b, \sigma_{b+1}]$ & Time interval $[s_b, e_b]$ & Span $h_b$ & Regime & $\omega_{\text{obj}, b}$ (E--M / M--L) \\
\midrule
1  & $[1.000000, 0.988235]$ & $[0.000000, 0.011765]$ & $0.011765$ & Early  & $2\ /\ 1$ \\
2  & $[0.988235, 0.972973]$ & $[0.011765, 0.027027]$ & $0.015262$ & Early  & $2\ /\ 1$ \\
3  & $[0.972973, 0.952381]$ & $[0.027027, 0.047619]$ & $0.020592$ & Early  & $2\ /\ 1$ \\
4  & $[0.952381, 0.923077]$ & $[0.047619, 0.076923]$ & $0.029304$ & Early  & $2\ /\ 1$ \\
\midrule
5  & $[0.923077, 0.878049]$ & $[0.076923, 0.121951]$ & $0.045028$ & Middle & $2\ /\ 2$ \\
6  & $[0.878049, 0.800000]$ & $[0.121951, 0.200000]$ & $0.078049$ & Middle & $2\ /\ 2$ \\
7  & $[0.800000, 0.631579]$ & $[0.200000, 0.368421]$ & $0.168421$ & Middle & $2\ /\ 2$ \\
\midrule
8  & $[0.631579, 0.521739]$ & $[0.368421, 0.478261]$ & $0.109840$ & Late   & $1\ /\ 2$ \\
9  & $[0.521739, 0.342857]$ & $[0.478261, 0.657143]$ & $0.178882$ & Late   & $1\ /\ 2$ \\
10 & $[0.342857, 0.000000]$ & $[0.657143, 1.000000]$ & $0.342857$ & Late   & $1\ /\ 2$ \\
\bottomrule
\end{tabular}%
\end{table*}

\section{Block Scheduling and Ablation}
\label{sec:block_scheduling_ablations}

\paragraph{Schedule specification.}
As described in Sec.~\ref{sec:block_scheduling} of the main paper, our 10-step ($B = 10$) PDD generator partitions the denoising trajectory into three functional regimes according to the block-start noise level $\sigma_b = 1 - s_b$ (with exact boundary values $\sigma_5 = 0.923077 \approx 0.9231$ and $\sigma_8 = 0.631579 \approx 0.6316$; see Table~\ref{tab:block_schedule} for the complete schedule): early blocks ($b \in \{1, 2, 3, 4\}$, $\sigma_b > 0.9231$), middle blocks ($b \in \{5, 6, 7\}$, $0.6316 < \sigma_b \le 0.9231$), and late blocks ($b \in \{8, 9, 10\}$, $\sigma_b \le 0.6316$). Under our objective-aware block scheduler $p_{\text{obj}}(b) = \omega_{\text{obj}, b} / \sum_{b'=1}^{B} \omega_{\text{obj}, b'}$, we adopt a $2:1$ relative weighting ratio between prioritized and non-prioritized blocks (setting $\omega_{\text{obj}, b} = 2$ for blocks within the target regime and $\omega_{\text{obj}, b} = 1$ otherwise) while keeping all updates strictly on native PDD boundaries:
\begin{itemize}
    \item \textbf{Early-to-mid emphasis} (for offline alignment $\mathcal{P}_{\text{off}}$ and online structural/reference alignment $\mathcal{P}_{\text{on}}$): We prioritize blocks $b \in \{1, \dots, 7\}$ ($\sigma_b > 0.6316$, i.e., $\sigma_b \ge \sigma_7 = 0.8000$) over late blocks $b \in \{8, 9, 10\}$ with a $2:1$ weight ratio ($\omega_{\text{obj}, b} = 2$ for $b \le 7$ and $\omega_{\text{obj}, b} = 1$ for $b \ge 8$) to target global motion, anatomical/physical validity, instruction following, and multi-reference identity binding.
    \item \textbf{Mid-to-late emphasis} (for semi-online composition $\mathcal{P}_{\text{comp}}$ and texture refinement $\mathcal{P}_{\text{tex}}$): Starting from the partial-noising boundary $b_0 = 5$ ($\sigma_{b_0} = \sigma_5 = 0.923077$), we prioritize blocks $b \in \{5, \dots, 10\}$ ($\sigma_b \le 0.9231$) over early blocks $b \in \{1, \dots, 4\}$ with a $2:1$ weight ratio ($\omega_{\text{obj}, b} = 2$ for $b \ge 5$ and $\omega_{\text{obj}, b} = 1$ for $b \le 4$) to refine spatial composition, high-frequency surface textures, color saturation, and fine text rendering without perturbing global trajectory structure.
\end{itemize}

\paragraph{Ablation against uniform block sampling.}
By the algebraic equivalence in Eq.~\eqref{eq:block_error}, BlockAlign with uniform block sampling $p_{\text{unif}}(b)=1/B$ corresponds to velocity-residual preference optimization adapted to query valid PDD boundaries $s_b$ and fused block-level predictors $\bar{v}_{\theta,b}$.
While our main 1,500-step experiments evaluate the full BlockAlign framework against DirectAlign and the unaligned PDD baseline, we conduct an additional controlled ablation on MiniMax-H3 to assess the effect of replacing $p_{\text{unif}}(b)$ with the objective-aware scheduler $p_{\text{obj}}(b)$ on the same boundary-aligned fused-head interface.
The two models are initialized from the same MiniMax-H3 PDD checkpoint and share the same training data, preference objectives, all other training hyperparameters, and random seeds, with a matched budget of 1,000 optimization steps.
This ablation isolates the scheduler choice within that budget; it does not provide a full decomposition of the framework or separately compare the uniform-sampling model against PDD and DirectAlign.

In a blind head-to-head human preference evaluation across all $N=150$ test conditions under identical conditioning inputs, random seeds, and inference settings, the model trained with $p_{\text{obj}}(b)$ achieves $72$ wins, $51$ losses, and $27$ ties against the model trained with $p_{\text{unif}}(b)$ ($48.0\%$ wins, $34.0\%$ losses, and $18.0\%$ ties).
This corresponds to a $57.0\%$ win-plus-half-tie preference score, compared with $43.0\%$ for uniform sampling.
Resampling the $150$ condition-level scores in $\{1,0.5,0\}$ with replacement yields a $95\%$ percentile bootstrap confidence interval of $[49.7\%,64.0\%]$ for the objective-aware scheduler's score.
Among the $123$ decisive conditions, its non-tie win rate is $58.5\%$ ($95\%$ Wilson CI: $[49.7\%,66.9\%]$; two-sided exact binomial test against a win probability of $0.5$: $p=0.0709$).
The point estimates favor objective-aware scheduling, suggesting a potential additional alignment benefit within the fixed compute budget; however, both confidence intervals include parity, and the non-tie test does not reach significance at the conventional $0.05$ level.

\end{document}